\documentclass[10pt,twocolumn,letterpaper]{article}

\usepackage{times}
\usepackage{epsfig}
\usepackage{graphicx}
\usepackage{amsmath}
\usepackage{amssymb}
\usepackage{siunitx} 
\usepackage{newtxtext}
\usepackage{floatpag}
\usepackage{floatflt}

\usepackage[pagebackref=true,breaklinks=true,letterpaper=true,colorlinks,bookmarks=false]{hyperref}

\newcommand{\Mesh}{{\mathbf{M}}}

\newcommand{\Pointcloud}{{\mathbf{P}}}
\newcommand{\DistanceFunction}{{\mathcal{D}}}
\newcommand{\NormalPenalty}{{\mathcal{T}}}

\renewcommand{\phi}{\varphi}

\newcommand{\IN}{{\mathbb{N}}}
\newcommand{\IR}{{\mathbb{R}}}
\newcommand{\IS}{{\mathbb{S}}}

\newcommand{\cI}{{\mathcal{I}}}
\newcommand{\cT}{{\mathcal{T}}}

\newcommand{\ds}{\,\mathrm{ds}}
\newcommand{\du}{\,\mathrm{du}}
\newcommand{\dx}{\,\mathrm{dx}}

\newcommand{\abs}[1]{{\left|#1\right|}}
\newcommand{\norm}[1]{{\left\|#1\right\|}}
\newcommand{\vecprod}[2]{{\left\langle#1,#2\right\rangle}}

\DeclareMathOperator{\dist}{dist}

\date{\vspace{-5ex}}
\begin{document}

\title{Compression for Smooth Shape Analysis}

\author{V.~Estellers \\ TUM \\ virginia.estellers@tum.de
\and
F.R.~Schmidt \\ TUM, BCAI \\ frank.schmidt12@de.bosch.com
\and
D.~Cremers \\ TUM \\ cremers@tum.de
}

\maketitle


\begin{abstract}
   Most 3D shape analysis methods use triangular meshes to discretize both the shape and functions on it as piecewise linear functions. With this representation, shape analysis requires fine meshes to represent smooth shapes and geometric operators like normals, curvatures, or Laplace-Beltrami eigenfunctions at large computational and memory costs.

  We avoid this bottleneck with a compression technique that represents a smooth shape as subdivision surfaces and exploits the subdivision scheme to parametrize smooth functions on that shape with a few control parameters. This compression does not affect the accuracy of the Laplace-Beltrami operator and its eigenfunctions and allow us to compute shape descriptors and shape matchings at an accuracy comparable to triangular meshes but a fraction of the computational cost.

  Our framework can also compress surfaces represented by point clouds to do shape analysis of 3D scanning data.
\end{abstract}
\begin{figure*}
  \centering
    \begin{tabular}{cccc}
\setlength\extrarowheight{-2pt}
	\includegraphics[width = 0.23\textwidth]{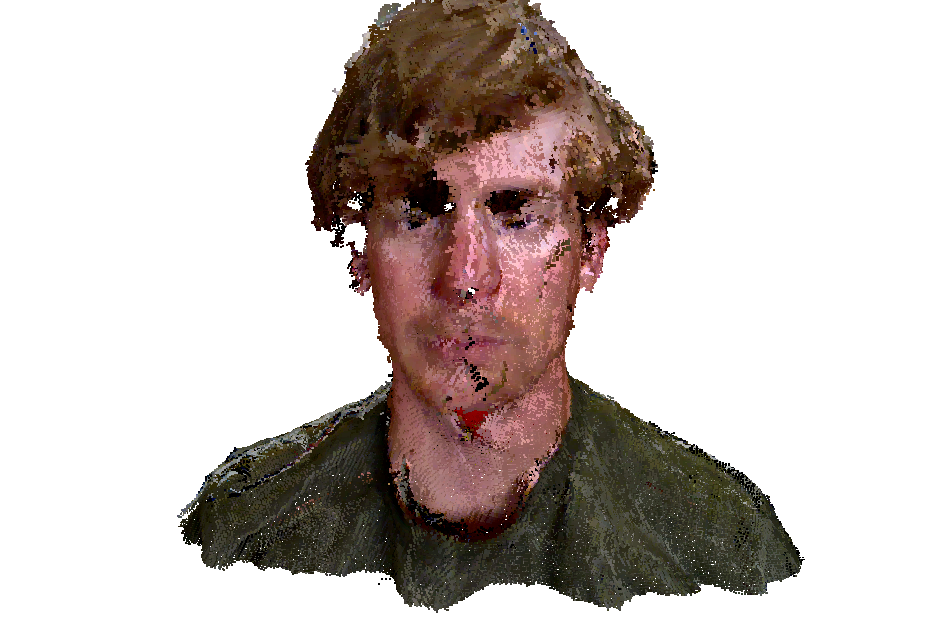} &
      \includegraphics[width = 0.23\textwidth]{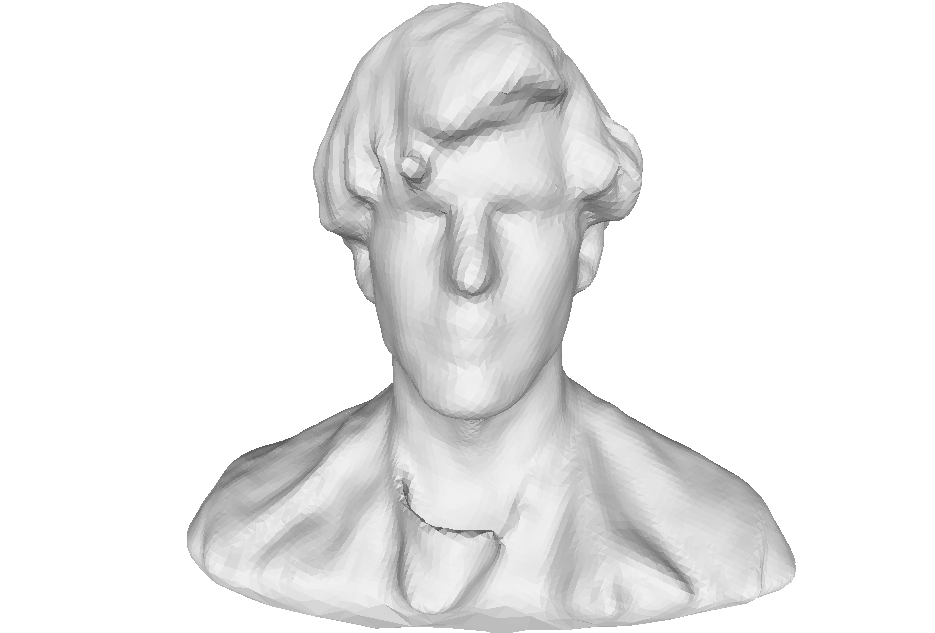}&
            \includegraphics[width = 0.23\textwidth]{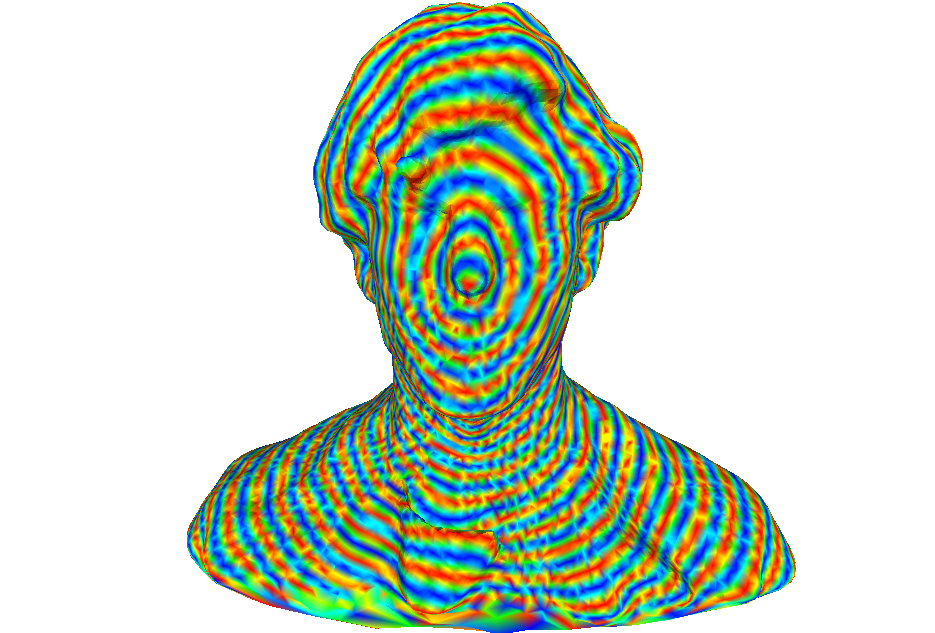}& 
      \includegraphics[width = 0.23\textwidth]{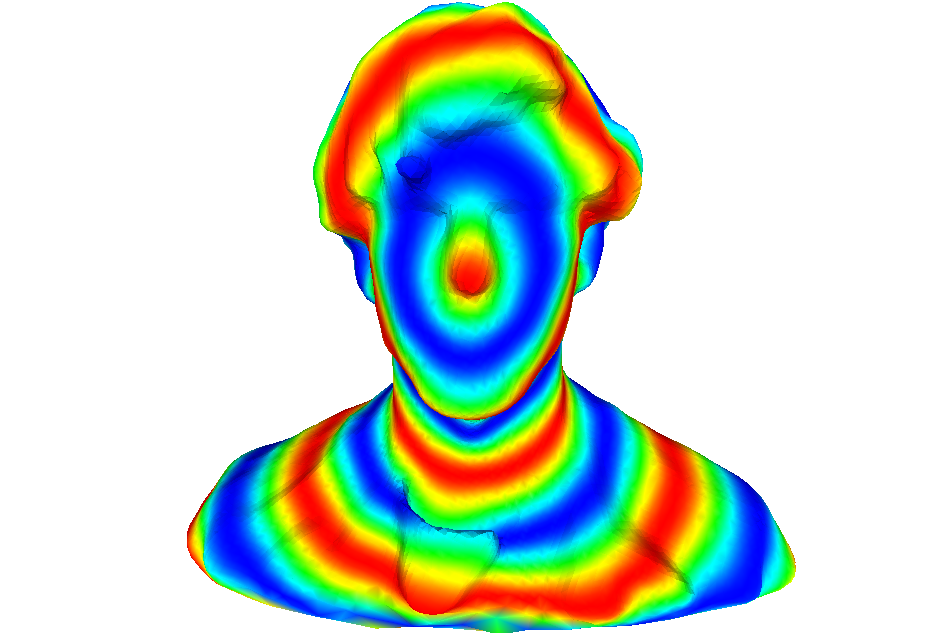} \\
      input pointcloud &  triangular mesh from \cite{Kazhdan2013} & 30 geodesic level lines & 5 geodesic 5 level lines \\
  501007 points &    86246 vertices & $\cos( 30 \pi g )$ & $\cos( 5 \pi g )$ \\
      \includegraphics[width = 0.23\textwidth]{head_pointcloud.png}  &      
      \includegraphics[width = 0.23\textwidth]{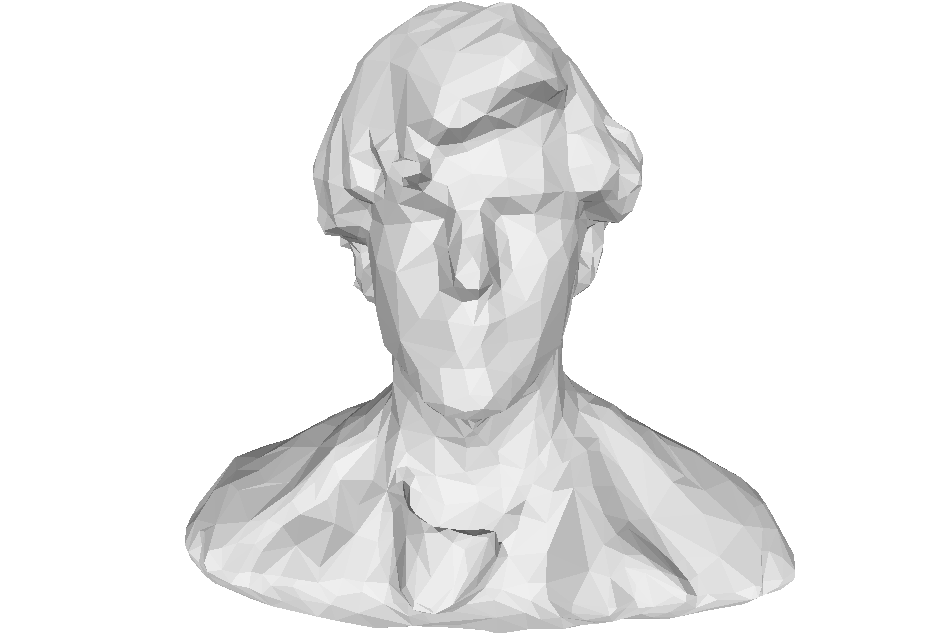}&
      \includegraphics[width=0.20\textwidth]{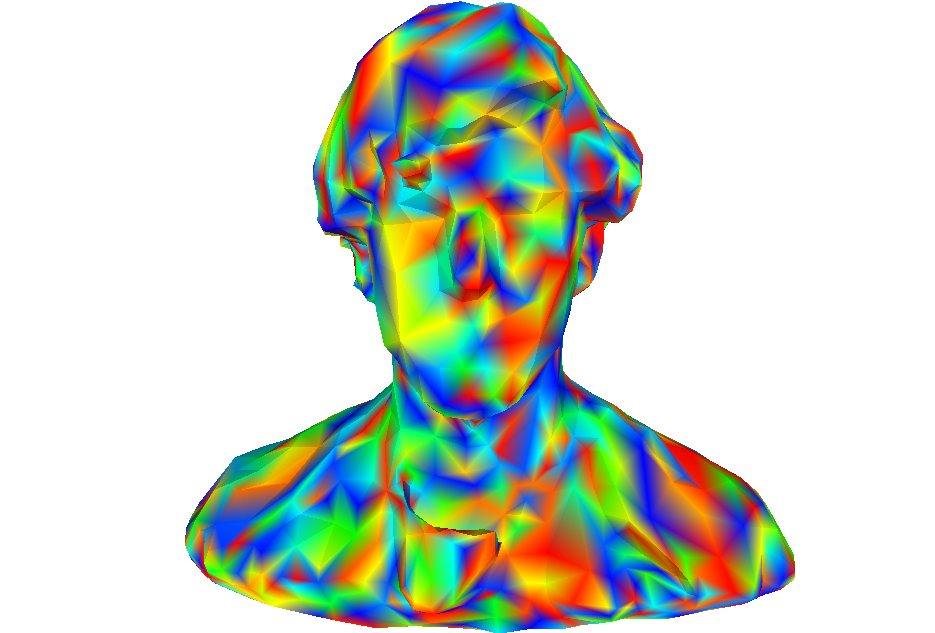} &
      \includegraphics[width=0.20\textwidth]{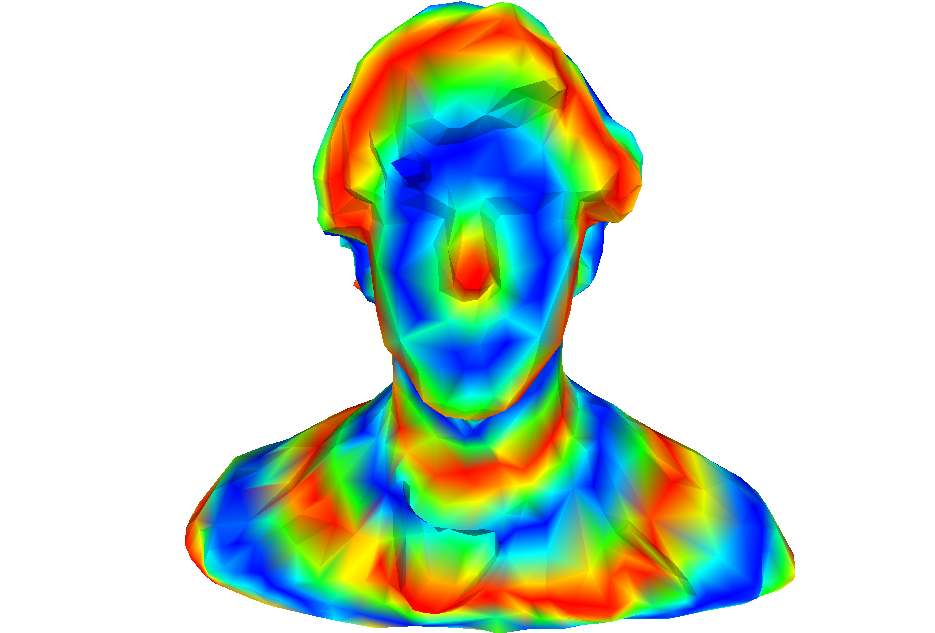} \\
      input pointcloud & edge-collapsed mesh \cite{Kazhdan2013} &  30 geodesic level lines & 5 geodesic 5 level lines \\
   501007 points &   1500 vertices &  $\cos( 30 \pi g )$ & $\cos( 5 \pi g )$ \\
      \includegraphics[width = 0.23\textwidth]{head_pointcloud.png} &
      \includegraphics[width=0.20\textwidth]{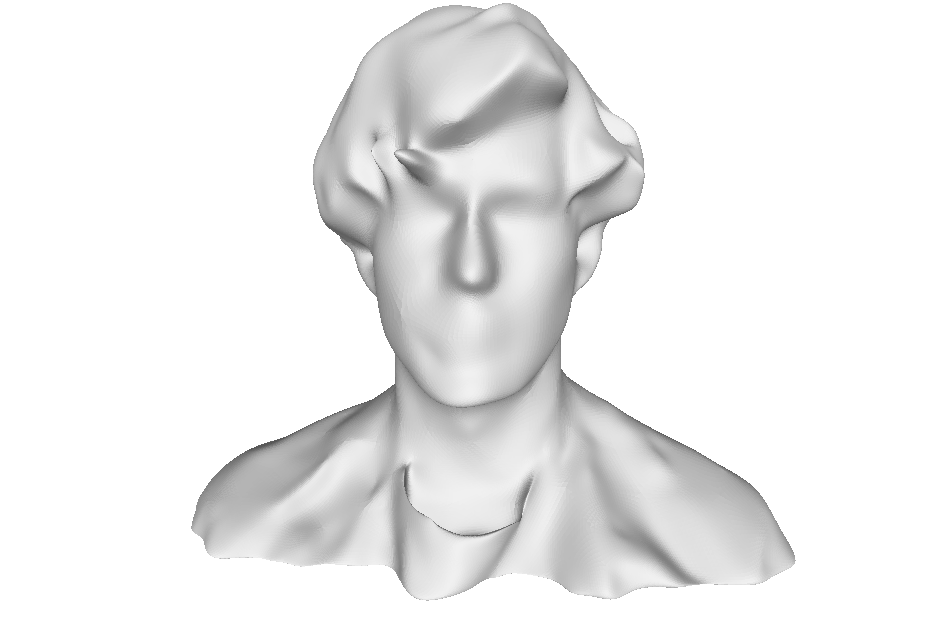} & 
 	  \includegraphics[width=0.20\textwidth]{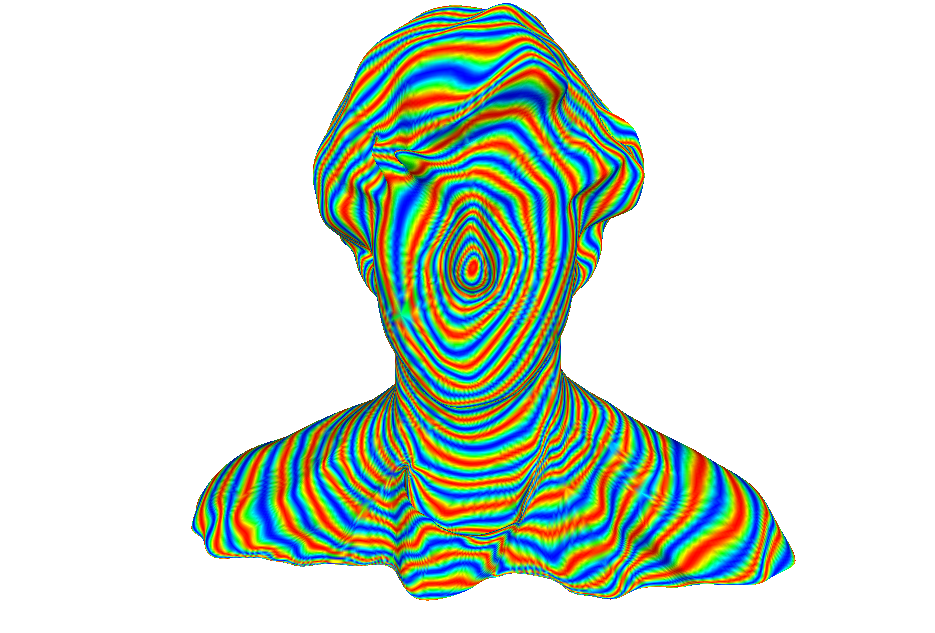} &
 	  \includegraphics[width=0.20\textwidth]{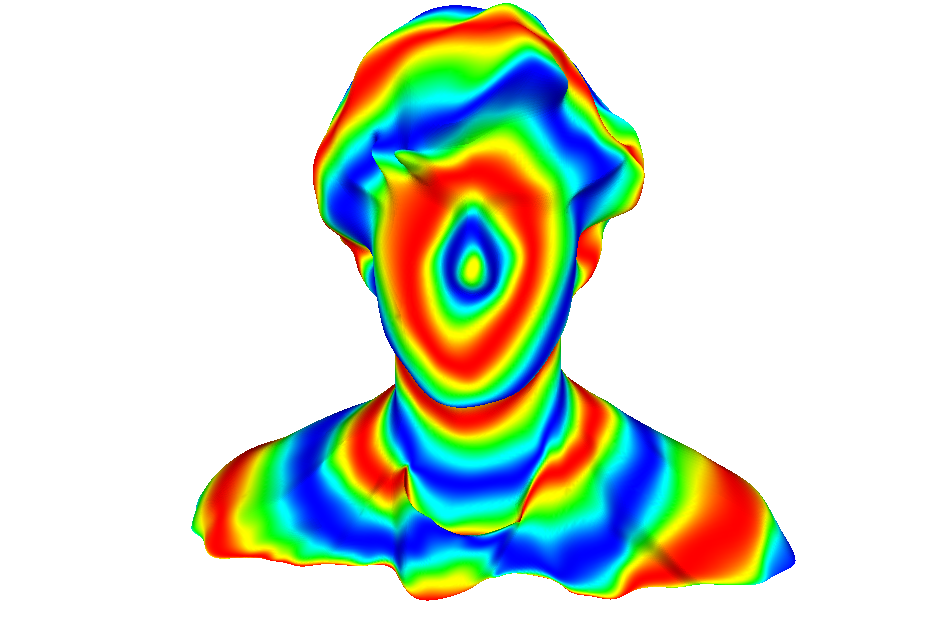}\\ 
input pointcloud & subdivision surface & 30 geodesic level lines & 5 geodesic 5 level lines \\
 501007 points & 1500 control vertices  & $\cos( 30 \pi g )$ & $\cos( 5 \pi g )$ \\
          \end{tabular}     
  \caption{Computation of approximate geodesic $g$ with different surface representations by the Heat method \cite{Crane2013} . Column 1: input Kinect pointcloud. Column 2: representation of the surface with a high-resolution mesh obtained with Poisson reconstruction~\cite{Kazhdan2013} (row 1), a low-resolution triangular mesh obtained by quadratic edge collapse~\cite{Garland1997} of the high-resolution mesh (row 2), and with our subdivision surface (row 3). Columns 3 and 4: level lines of the geodesic visualized as $\cos( \varpi g)$. Our subdivision surface is comparable to the surface obtained by Poisson reconstruction and faithfully represent geodesics at low and high resolution with 2\% of vertices. Compressing the Poisson mesh by edge collapse looses all the small scale details of the surface and its geodesic, as highlighted by high-frequency level lines of the geodesic.}
  \label{fig:geodesics0}         
\end{figure*}  
\section{Introduction}

Many shape analysis tasks describe shapes as smooth manifolds and analyze them with respect to their geometry in terms of normals, curvatures, and geodesics that require accurate estimates of fist- and second-order derivatives over the shape to compute tangent spaces, the Riemannian metrics, or Laplace-Beltrami operators~\cite{reuter06,Levy2006,Windheuser2014,Litman2014}. 
When the surface is represented as a mesh or point cloud, these differential operators can only be computed approximately at large memory and computational costs. 

Once these differential operators have been used to compute normals or metrics, most of the high resolution information is discarded by operating only with the leading eigenfunctions of the Laplace-Beltrami operator necessary to estimate geodesics~\cite{Crane2013}, compute shape descriptors~\cite{aubry-et-al-11,reuter06,Levy2006,Litman2014} or shape matches~\cite{Ovsjanikov2012} with tractable problem sizes. This creates a paradox, as shapes are first discretized with fine meshes or large pointclouds to estimate high-dimensional shape operators that are mostly discarded to analyze the shape. This paradox appears because we only need large meshes or dense pointclouds to discretize accurate differential operators over the shape, not for shape analysis.

We propose a paradigm shift that avoids this paradox by representing a shape with a subdivision surface~\cite{Notes2000} that parametrizes the surface with a small set of smooth base functions. Differentiability is then intrinsic to the shape representation and does not require large bases to compute principal curvatures, wave-kernel signatures, or Laplace-Beltrami eigenfunctions. Subdivision surfaces are a generalization of splines to surfaces of arbitrary topology that 
provide set of base functions to compactly parametrize the surface as well as smooth functions over it. As a result, subdivision surfaces provide a compressed shape representation well suited to shape analysis.

Our first contribution (Section~\ref{sec:model}) is a method to estimate a subdivision surface from the triangular meshes common in shape analysis\cite{bronstein08}. This kind of manifold meshes, however, are far from the meshes or pointclouds obtained from range measurements or computer-vision models. As these low-level representations proliferate, they create a challenge for shape analysis to adapt to the characteristics of raw data (irregular surface sampling, noise, and outliers) that cause havoc in the estimation of differential operators. Up until now, raw measurements have been largely ignored in shape analysis by estimating a mesh representation from a pointcloud and performing analysis on this representation. This avoids dealing with noise and outliers but ignores the direct measurements and makes shape analysis dependent on the model used to estimate the surface.\footnote{Least-squares models \cite{Kazhdan2013,Calakli2011}, for instance, oversmooth surface edges and require less eigenfunctions to match than the $\ell_1$ models~\cite{Ummenhofer2015,Estellers2016} that emphasize corners.}.
For this reasons, our second contribution is a robust method that fits a subdivision surface to a noisy pointcloud and estimates the geometric operators required to do shape analysis on it. Our model is robust to both noise and outliers in the pointcloud and to erroneous correspondences between the input points and the subdivision surface. This is a critical aspect of fitting a parametric surface to raw data without developing the non-manifold artifacts that prevent shape analysis.

Our third contribution is a fast optimization algorithm (Section~\ref{sec:optimization}) that combines a sequential quadratic approximation with a good initialization strategy. 
This is important because fitting a subdivision surface is not a convex problem and vanilla initializations~\cite{Jaimez2017} oversmooth the surface to avoid small-scale local minima and prevent self-intersections.

Our fourth contribution (Section~\ref{sec:shape_compression}) describes how to accurately compute many shape operators with a subdivision surface. We focus on Laplace-Beltrami operator, wave kernel signatures~\cite{aubry-et-al-11}, approximate geodesics \cite{Crane2013} and prove its efficacy in the full-pipeline task of shape matching~\cite{Ovsjanikov2012}. Our experiments show state-of-the-art performance at a fraction of the memory requirements of shape representations with triangular meshes. 


\section{Smooth Shapes}\label{sec:shape_compression}
In this paper, a \emph{3D shape} is a smooth\footnote{
  We call a $C^1$ object smooth if it is $C^2$ except for a finite
  subset. This differs from the mathematical notion, which identifies
  smoothness with $C^\infty$.}, compact oriented surface $S\subset\IR^3$ without boundaries.

For computational purposes, shapes are usually discretized by a
mesh $\Mesh=(V,F)$ in terms of two matrices: the $i$-th row of $V\in\IR^{n\times3}$ stores the location of the $i$-th vertex of the mesh, while each row of $F\in\IN^{m\times k}$ stores the indices of the vertices of a face of $\Mesh$. We focus on quad meshes with $k=4$.

In shape analysis, quad meshes usually discretize a smooth surfaces as a piecewise $C^1$ surface that is not smooth. Subdivision surfaces, on the other hand, only use meshes to discretize the domain of a smooth surface parametrization.  To formally introduce this parametrization, we first revisit the topological space associated to a quad mesh and show how it can be used to define a smooth subdivision parametrization.

\subsection{Meshes -- Piecewise Bilinear Surfaces}\label{sec:quad}
The regular quad mesh parametrization of a surface $S_\Mesh$ consists of quadrilaterals ``glued'' together along their edges. Each quadrilateral in $S_\Mesh$ is modeled by a bilinear patch that fits the fours vertices of a facet in $\Mesh$ and is glued to its neighboring facets. This topological gluing extends the independent parametrization of each quadrilateral to the whole mesh and is formalized by the topological space, $\cT_\Mesh$, that defines the domain of the parametrization~\cite{Juettler-et-al-16}.

Formally, given a quad mesh $\Mesh=(V,F)$, the topological space $\cT_\Mesh$ is defined by a set of quads that is indexed by the mesh facets $\cI_F=\{1,\ldots,m\}$ and glued together by the equivalence relation $\sim$, that is $\cT_\Mesh:=\square\times\cI_F/\sim$. The \emph{reference quad} $\square=[0,1]\times[0,1]$ parametrizes each quadrilateral patch while the equivalence relation $\sim$ connects neighboring faces along their common edge.
%

A parametrization of the surface is then obtained by defining a piecewise bilinear function 
$$B_\Mesh^i\colon\cT_\Mesh\to\IR$$ for each vertex in $x_i \in \cT_\Mesh$ satisfying $B_\Mesh^i(x_j)=\delta_{ij}$. The resulting parametrization 
$$\Phi_\Mesh\colon\cT_\Mesh\to S_\Mesh\subset\IR^3$$ defines the piecewise $C^1$ surface $S_\Mesh:=\Phi_\Mesh(\cT_\Mesh)$ by bilinear interpolation 
$$\Phi(u) = \sum_{i=1}^n B_\Mesh^i(u)\cdot v_i.$$
With such a piecewise $C^1$ paremetrization, Riemannian quantities like the first or second fundamental form can only be approximated and need large meshes with small quadrilaterals for accurate approximations. This limits their use in real-time applications and calls for other shape representations.%


\subsection{Smooth Subdivision Surfaces}\label{sec:sds}

A subdivision surface represents a smooth surface $S$ by a coarse control mesh $\Mesh^0=(V^0,F^0)$ and a subdivision scheme. The subdivision scheme transforms a mesh $\Mesh^k=(V^k,F^k)$ into a finer
mesh $\Mesh^{k+1}$ such that if we start with $\Mesh^0:=(V,F)$, we obtain the smooth surface as the limit of iterating subdivision ad infinitum:$$S = \lim_{k\to\infty} S_{\Mesh^k}.$$

\begin{figure*}
  \centering
    \begin{tabular}{cccc}
      \includegraphics[trim=0cm 0cm 0cm 0cm, clip=true, width = 0.18\textwidth]{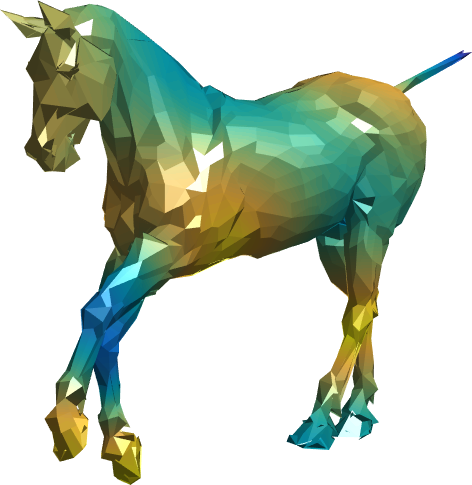}&
      \includegraphics[trim=0cm 0cm 0cm 0cm, clip=true, width = 0.18\textwidth]{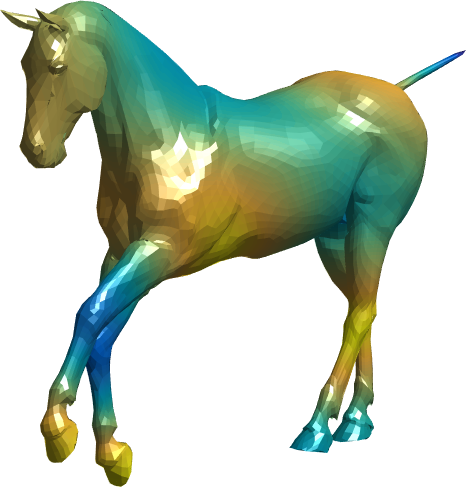}&
      \includegraphics[trim=0cm 0cm 0cm 0cm, clip=true, width = 0.18\textwidth]{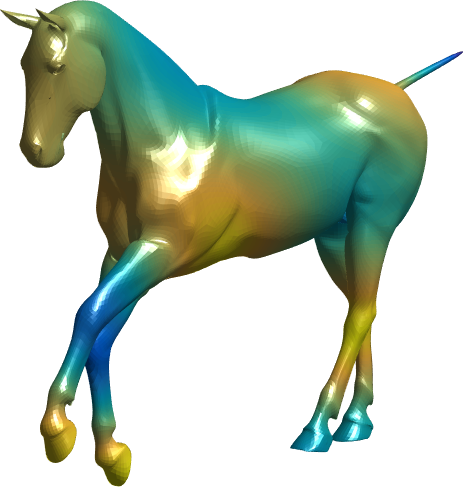}&
      \includegraphics[trim=0cm 0cm 0cm 0cm, clip=true, width = 0.18\textwidth]{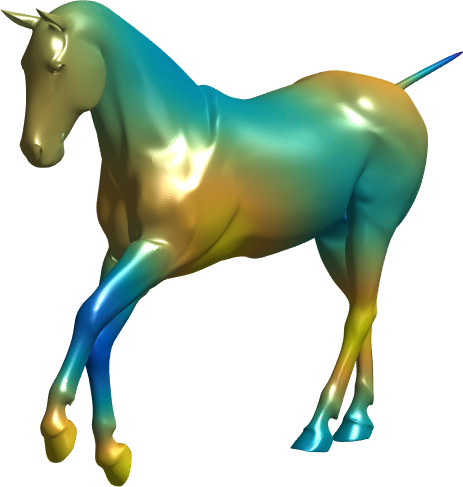} \\
      $\Mesh^0$ &
      $\Mesh^1$ &
      $\Mesh^2$ &
      $\Mesh^3$ \\
      1701 Control Points & %
      6798 Vertices &%
      27186 Vertices &%
      108738 Vertices                       
    \end{tabular}
  \caption{Different subdivision levels $\Mesh^k$ of Catmull-Clark subdivision scheme applied to a control mesh and a smooth function defined over the mesh (its 18-th $\Delta$ eigenfunction). $\Mesh^3$ is already a good approximation of the smooth surface $S=\Mesh^\infty$.}
  \label{fig:sds}
\end{figure*}

In this paper we follow the Catmull-Clark subdivision scheme $\Mesh^k\mapsto \Mesh^{k+1}$  consisting of two steps: First, the topology of the new mesh $\Mesh^{k+1}$ is defined by subdividing each quad in $\cT_\Mesh^{k}$ into its four quadrants that define four adjacent facets in $\cT_\Mesh^{k+1}$. Second, the location of each vertex of the new mesh $\Mesh^{k+1}$ is computed as a linear combination of the positions of the nearby vertices. Figure \ref{fig:sds} illustrates this subdivision process.

This produces a compressed surface representation because the topology of $\Mesh^{k+1}$ is completely determined by the topology of $\Mesh^{k}$. Similarly, the space of bilinear functions over $\cT_{\Mesh^{k+1}}$ created by subdivision is determined by the space of bilinear functions over $\cT_{\Mesh^{k}}$. This means that we can parametrize the $i$-th function created by subdivision $B_{\Mesh^{k+1}}^i$ over the domain $\cT_\Mesh^{k}$, and by induction over $\cT_\Mesh$, instead than $\cT_\Mesh^{k+1}$. We denote these functions $B_\Mesh^{k,i}~\colon~\cT_\Mesh~\to~\IR$ and use them to parametrize the surface $S_{\Mesh^{k}}=\Phi_\Mesh^k(\cT_\Mesh)$ as a function of the vertices of $\Mesh^{k}$.

In the second step, the position of the vertices of $\Mesh^{k+1}$ are
computed as linear combination of the positions of the vertices of
$\Mesh^{k}$. The weights of this linear combination are organized into
a subdivision matrix $\hat{A}_k\in\IR^{n_k\times~n_{k-1}}$ that
computes the position of the vertices of $\Mesh^k$ as 
$$V_k=\underbrace{\hat{A}_k\ldots\hat{A}_1}_{A_k \in \mathbb{R}^{n_k \times n}}~V=A_k~V.$$
Using this linear dependency, we can parametrize a surface approximation $S_{\Mesh^k}$ resulting from $k$ subdivisions directly over $\cT_\Mesh$ as follows:
\begin{align*}
  \Phi_\Mesh^k(u) = \sum_{i=1}^{n_k}B_\Mesh^{k,i}(u) \left(A_{k}
    V\right)_i = %
  \sum_{j=1}^{n}\sum_{i=1}^{n_k}B_\Mesh^{k,i}(u) A_{k,ij}  v_j
\end{align*}
By iterating this process, the subdivision surface of the limit $$S = \lim_{k\to\infty} S_{\Mesh^k}$$ can be formulated as a limit of functions defined over the topological space $\cT_\Mesh$. Evaluating the limit surface at the location of the vertex thus amounts to computing the base functions $$
    \Phi_j:=\lim_{k\to\infty}\sum_{i=1}^{n_k}B_\Mesh^{k,i}\cdot
    A_{k,ij}.$$
    
The existence of this limit and the smoothness of the limit functions is guaranteed by the subdivision scheme. All the \emph{limit base functions} $\Phi_j$ are compactly supported because the subdivision rules act locally and converge to spline basis functions. This equivalence provides analytic derivatives and fast evaluation techniques \cite{Stam1998} for smooth functions or tangent vectors defined over the surface and makes subdivision surfaces well-suited for shape analysis.

We focuses on Catmull-Clark subdivision~\cite{catmull-clark-78} for simplicity. This scheme ensures the existence and $C^2$ smoothness of the basis functions everywhere except at extraordinary vertices (vertices with valence different than four). This scheme is designed for control grids with quadrilateral connectivity, but our compression generalizes to any subdivision schemes that parametrizes $C^2$ surfaces.

\section{Robust Fitting Model}\label{sec:model}

We formulate the problem of fitting a subdivision surface $S$ to a
  set of surface samples $\Pointcloud= \{p_1,\ldots, p_N \}$ and a set
  of normals $\mathbf{T}=\{t_1,\ldots,t_N\}$ as the  minimization problem:
\begin{align}\label{eq:optimization}
  \min_{S} \ E(S) :=& \underbrace{\dist(S,\Pointcloud)}_\text{Point Fit} + %
  \underbrace{\alpha\cdot\NormalPenalty(S, \mathbf{T})}_\text{Tangent
    Fit}+%
  \underbrace{\beta\cdot R(S)}_\text{Regularization},
\end{align}
where $\alpha,\beta\geq0$ are model parameters. %
  The different terms take different information into account:
  $\dist(S,\Pointcloud)$ fits the surface $S$ to the points
  $\Pointcloud$, $\NormalPenalty(S,\mathbf{T})$ matches the tangent plane of $S$ to the normal information $\mathbf{T}$, and $R(S)$ regularizes the discrete representation of $S$ to avoid degenerate faces in $\Mesh$. If no normal information is provided, $\alpha = 0$. %

\subsubsection*{Point Fit}
Let
  $\DistanceFunction_{S}(p)=\min_{s\in S}\norm{s-p}$ denote the Euclidean
  distance between a point $p\in\IR^3$ and the compact surface $S$, we
  define the \emph{point fit energy} as
\begin{align}
  \dist(S,\Pointcloud) := \sum_{j=1}^N\DistanceFunction_{S}(p_j)^q = \sum_{j=1}^N\min_{u_j\in\cT_\Mesh}\norm{\Phi(u_j)-p_j}^q \nonumber.
\end{align}
Using the $q$-th power of the distance with $q\in(1,2)$, instead of the common squared distance, makes our model robust to outliers and scanning artifacts in the input samples.
The minimization in $u_1,\ldots,u_N$ results from the parametric representation of the surface $S=\Phi(\cT_\Mesh)$ and introduces a large number of additional variables over which the objective function is not convex. We avoid the optimization of the correspondence parameters $u_1,\ldots,u_N$ by using a second-order approximation of the squared distance function to a surface in Section~\ref{sec:optimization}.

\subsubsection*{Tangent Fit}
To obtain a surface that also matches the normals, we include the \emph{tangent energy}
  \begin{align}
    \NormalPenalty(S, \mathbf{T}) = \sum_{j=1}^{N} %
    \abs{\vecprod{t_j}{\partial_1 \Phi(u_j)}}^q + %
    \abs{\vecprod{t_j}{\partial_2 \Phi(u_j)}}^q,
  \end{align} %
where $t_j$ is the input surface normal at point $p_j$ and
$\partial_i\Phi(u_j)$ the $i$-th base vector of the tangent space $T_{\Phi(u_j)}S$. Each term
in $\NormalPenalty$ aligns the tangent space of the subdivision
surface in the direction orthogonal to the sample normal. This penalty
is designed to be independent of the orientation of the normals to account for noisy point clouds where estimating consistent normal orientations is prone to fail. The $\ell_q$-norm makes this term robust to noise and outliers in the normals and the correspondence parameters $u_1,\ldots, u_N$.

\subsubsection*{Regularization}
The regularizer $R(S)$ penalizes the squared distance between the
vertices of the mesh $M$ incident to the same quad. This keeps the
size and shape of the quads regular and avoids skewed elements that
cause instabilities in finite-element computations. The regularizer is thus a simple quadratic penalty that can be described by a sparse matrix $R$ that has a row for each edge in the mesh with entries $\pm 1$ at the columns of the vertices incident to the edge. We choose this regularizer for its simplicity.

\subsection{Initialization and Surface Topology}\label{sec:initialization}

A major challenge of fitting any explicit parametrization to a set of
points is the lack of convex formulations for the optimization
problem. Non-convexity introduces the difficulty of
finding a suitable initialization that leads to a \textit{good} local minimum. It is an important step of the algorithm that conditions the topology of the surface.
  
By parameterizing $S$ with a subdivision surface, our optimization problem is formulated in terms of the control mesh $\Mesh=(V,F)$. To avoid solving a combinatorial problem, we fix the topology of the control mesh, and thus the topology of the surface, and only optimize for the location of the vertices $V$. We design the control mesh to satisfy two properties: it determines a surface with the same topology as the input data, and it has a small
number of vertices to represent the surface compactly. Depending on the input data, we apply different methods to obtain a good initialization.

If the point samples $\Pointcloud$ are the vertices of a mesh $\overline\Mesh$, the topology of the surface is already encoded by the mesh. Without loss of generality, we assume that $\overline\Mesh$ is a triangular mesh without boundaries and use quadratic edge-collapse~\cite{Garland1997} to reduce the number of vertices in the mesh and preserve the surface topology. We then transform this triangular mesh into a quad mesh by solving a perfect matching problem that pairs triangles of the collapsed mesh to create quads~\cite{Remacle2012a}. For each edge $e$, let $\alpha_1(e),\ldots,\alpha_4(e)\in\IS^1$ be the angles of the quad that results from removing $e$ and $\eta(e)\in\IS^1$ the angle between the normals of the triangles incident to $e$. We set the cost of removing an edge $e$ to
  \begin{align*}
  \small
    c(e) = \begin{cases} \frac14\sum_{i=1}^4 d\left(\alpha_i(e),\frac\pi2\right)^2 +
    \tan(\eta(e))^2 & \text{if
        $\abs{\eta(e)}<\frac\pi2$}\\
      \infty & \text{if $\abs{\eta(e)}\geq\frac\pi2$}.
    \end{cases}   
  \end{align*}
to assign infinity costs to edges that connect strongly bent triangles ($\eta(e)>\frac\pi2$) or edges that would create quads with straight or reflex angles. The cost favors rectangular quads to obtain a subdivision surface where finite-element computations are accurate and numerically stable. We solve the perfect matching and find the quad mesh with minimum cost with the \emph{Blossom V} method~\cite{Kolmogorov-MPC09}. 

When the input is a point cloud, we use an implicit representation to estimate the topology of the surface from the point samples and only then extract a control mesh. Estimating first the topology of the surface with an
implicit representation avoids explicitly handling topology changes during the compression process. We use the popular Poisson reconstruction~\cite{Kazhdan2013} for simplicity but other techniques like voxel hashing work well. From this implicit representation, we extract a closed triangular mesh with marching cubes~\cite{Lorensen1987} and create a compact quad mesh with the collapse and blossom techniques described for meshes.



\section{Efficient Optimization}\label{sec:optimization}

Our optimization algorithm exploits the properties of subdivision surfaces, namely, the compact support
of the basis functions and the ability to analytically evaluate its geometry, with a sequential quadratic program that is more efficient than the gradient-based algorithms proposed by~\cite{Cashman2013,Ili2006,Jaimez2017}. We solve
\eqref{eq:optimization} as a sequence of convex problems
  \begin{align}
    v^{m+1} \leftarrow \min_v \, v^\top\left(Q^m v - b\right)
  \end{align} %
that approximate the original energy around the current surface estimate by a least-squares problem. The approximation is derived from the surface geometry, while the sparsity of $Q^m$ results from compactness of the subdivision basis.

We follow a \emph{Majorize-Minimize} (MM) principle to minimize $E(S)$ by iterating two steps until convergence. The first step finds an upper envelop of the objective function $E(S|S_0)\geq E(S)$ that coincides with $E$ at $S_0$. We derive an upper envelop of the data and tangent fits from the inequality
\begin{align}
  \abs{d}^q \leq \frac{q}{2} \abs{d_0}^{q-2}d^2 +
  (1-\frac{q}{2}) \abs{d_0}^q && \forall d_0 \neq 0, ~q\in[1,2]\nonumber.
\end{align}%

The second step of the MM algorithm minimizes the upper envelop the upper envelop and drives the value of the original function downwards by solving $$S^{m+1}\leftarrow\min_{S}E(S|S^m).$$ In our case, the envelop 
\begin{footnotesize}
\begin{align}
E(S|S^m) = \sum_{j=1}^{n}
  w^m_j \DistanceFunction_{S}(p_j)^2 + \sum_{i=1}^2
  \alpha^m_{ij}\vecprod{t_j}{\partial_i\Phi^m(u_j)}^2 + \beta R(S)  \nonumber
\end{align}
\end{footnotesize}
defines a weighted least-squares problem where the weights $w^m_j$ and $\alpha^m_{ij}$ are
determined by the MM principle
\begin{align}
  w^m_j &= \frac{q}2\DistanceFunction_{S^m}(p_j)^{q-2} & %
  \alpha^m_{i j} &= \frac{\alpha q}{2} %
  \abs{\vecprod{t_j}{\partial_i\Phi^m(u_j)}}_{\epsilon}^{q-2}. \nonumber
\end{align} %
%
%
To simplify the minimization step of MM, we approximate $\DistanceFunction_{S^m}(\cdot)^2$ with a quadratic function that parametrizes the squared distance to a surface in terms of its geometry as follows:
\begin{align}
  \DistanceFunction^2_{S}(x) \approx (x-p)^\top\left[ \frac{d\cdot
      \tau_1\tau_1^\top}{d+\rho_1} + \frac{d\cdot
      \tau_2\tau_2^\top}{d+\rho_2} + \nu\nu^\top \right](x-p),
  \nonumber
\end{align}
where $p$ is a point on the surface $S$ close to $x$ (but not necessarily the projection of $x$ onto $S$), $d:=\norm{x-p}$ is the signed distance from $x$ to this point, and $\rho_i, \tau_i,  \nu$ are the principal curvature radii, the principal curvature directions and the normal to the surface at $p$.
This approximation is due to \cite{Pottmann2003a} and coincides with the second order Taylor approximation within the radius of curvature of the surface at $p$, where it can be relaxed into a positive definite quadratic
form by taking the absolute values of $d,~\rho_1,~\rho_2$.

Incorporating this approximation into the upper envelop corresponds to using a quasi-Newton algorithm for optimizing jointly the control vertices of the subdivision surface and the parametric coordinates of the surface point closest to $x$~\cite{Wang2006,Liu2008}. The approximation let us thus work with approximate correspondences between surface samples
and input points, instead of explicitly finding the correspondences by solving a larger optimization problem. It also makes our model robust to erroneous correspondences by taking into account their approximate nature in the objective function. The approximation, however, can affect the majorizing property and puts our algorithm into the
category of sequential quadratic programs instead of MM algorithms.
The combination of the MM upper envelope with the quadratic approximation of the squared distance to $S$ reduces to weighted least-squares problems in the vertices $V$ (c.f. Appendix) that we solve efficiently with conjugate gradient solver warm-started with the solution of the previous MM minimization step.

After optimizing $V$ at iteration $m$, we update the parameter values $U$ where we approximate the squared distance to the surface by sampling the surface $S^m=\Phi^m(\cT_\Mesh)$ uniformly and creating a kd-tree from these samples to find the parameter of the sample closest to each input point.
This results in an efficient algorithm that avoids the slow convergence rates of coordinate descent algorithms and the large optimization problems of the Levenberg-Marquard solvers advocated by \cite{Cashman2013,Ili2006,Jaimez2017}.

\section{Smooth Shape Analysis}\label{sec:smooth}
We have showed how to obtain smooth shapes from non-smooth input
data like point clouds or meshes. We now show how to do \emph{smooth shape analysis} with a subdivision surface $S$.  To this purpose, we focus on the computation of the Laplace-Beltrami operator and its eigenfunctions and use them to match shapes, compute shape descriptors, and approximate geodesics.

\subsection{Laplace-Beltrami-Operator}\label{sec:LBO}
Given a smooth shape $S$, the Laplace-Beltrami-Operator $\Delta$ maps a twice continuously differentiable function $g\in~C^2(S)$ to a continuous function $$\Delta(g)\colon S\to\IR,$$ If $g$ satisfies Neumann boundary conditions, then Stokes' theorem $$\int_S~\Delta~g~\cdot~h~\dx~=~\int_S~\vecprod{\nabla~g}{\nabla~h}~\dx$$ defines a weak formulation of $\Delta$ for Sobolev functions $g\in~H^1(S)$. The Galerkin method uses this identity to define a discretization of $\Delta$ in the ansatz space $H^1_n(S)$ spanned by Sobolev functions $\Phi_1,\ldots,\Phi_n\in H^1(S)$. 

Using $H^1_n(S)$ as test space, $g$, $h$ and $\Delta g$ are represented by vectors $\alpha,\beta,\gamma\in\IR^n$ as follows
\begin{align*}
  g =& \sum_{i=1}^n\alpha_i\Phi_i& %
  h =& \sum_{i=1}^n\beta_i\Phi_i& %
  \Delta g =& \sum_{i=1}^n\gamma_i\Phi_i&%
\end{align*}
and the weak formulation of $\Delta$ in $H^1_n(S)$ reads
\begin{align}
\vecprod{\gamma}{\beta}_{D_0} = \vecprod{\alpha}{\beta}_{D_1} ~\forall\beta \in \IR^n.\label{eq:weakLB}
\end{align}
The Laplace-Beltrami operator in $H^1_n(S)$ thus depends on the mass and stiffness matrices $D_0,D_1\in\IR^{n\times n}$ whose entries
\begin{align*}\left(D_0\right)_{ij} = \int_S \Phi_i\cdot\Phi_j\dx ~~
  \left(D_1\right)_{ij} = \int_S
  \vecprod{\nabla\Phi_i}{\nabla\Phi_j}\dx.
\end{align*}
compute the scalar product of functions and tangent vectors in the surface $S$.

From \eqref{eq:weakLB}, the Laplace-Beltrami operator in $H^1_n(S)$ can be parametrized by the matrix $D_2:=D_0^{-1}D_1$. This matrix has a non-negative spectrum and its eigenvector discretizes the eigenfunctions of the operator $\Delta$ in $H^1_n(S)$. Indeed, if $v\in\IR^n$ is an eigenvector of $D_2$ with eigenvalue $\lambda$, then $g_v=\sum_{i=1}^n v_i\Phi_i$ is an eigenfunction of $\Delta$ with eigenvalue $\lambda$ in $H^1_n(S)$.

\begin{figure*}
  \centering
    \begin{tabular}{ccc c}
      \includegraphics[width = 0.24\textwidth]{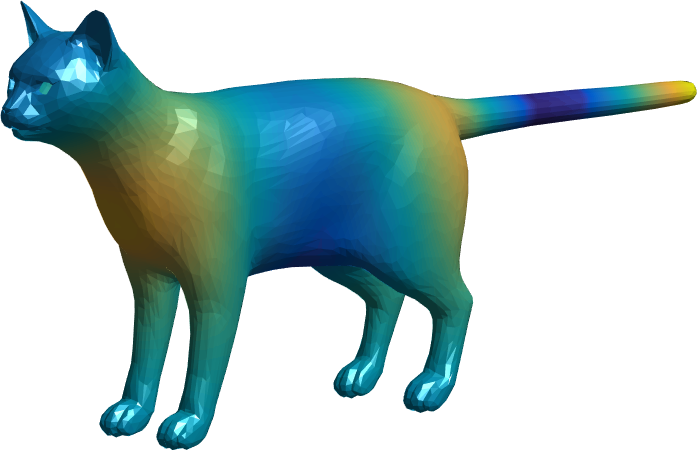}&
      \includegraphics[width = 0.24\textwidth]{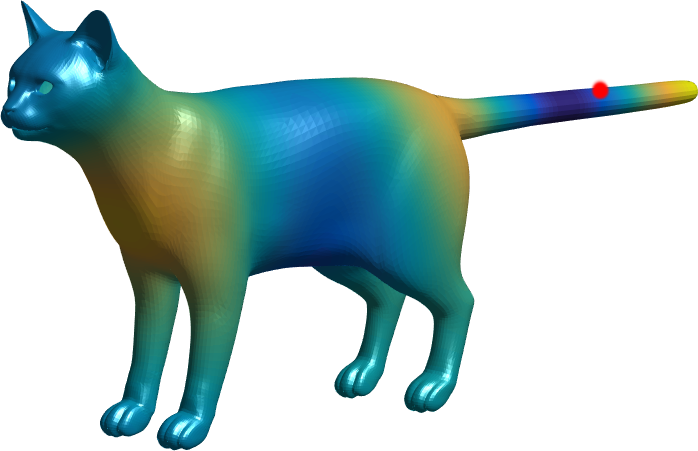}&
      \includegraphics[width = 0.24\textwidth]{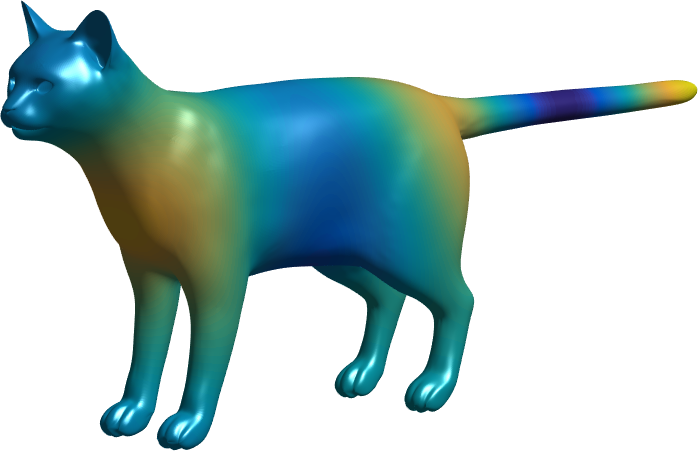}&
       \includegraphics[width = 0.2\textwidth]{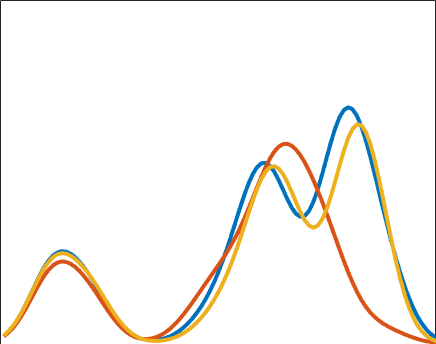}\\
      Mesh 3,400 vertices & Mesh 27,984 vertices & 
      Subdiv. Surf. 1700 vertices & Wave-kernel signatures
    \end{tabular}
  \caption{$12^\text{th}$ $\Delta$-eigenfunction and wave-kernel signature of the a surface point in red computed with fine (blue line) and coarse triangular meshes (red line) and a subdivision surface (yellow). The subdivision surface obtains a wake signature comparable to the fine mesh while the signature of the coarse mesh lacks detail.}
  \label{fig:eigen}
\end{figure*}

To particularize this construction to functions defined on a subdivision surface $S$, we consider the mapping $$X(u,f) = \sum^{n}_{i=1} \Phi_i(u,f) v_i$$ that parametrizes the surface as a linear combination of $n$ limit basis functions $\Phi_1,\ldots,\Phi_n$ and $n$ mesh vertices $v_1,\ldots,v_n$. Similarly, we can parametrize Sobolev funcctions over the surface $g\in H^1(S)$ by $n$ scalar values $g_1,\ldots,g_n\in\IR$ with a mapping
\begin{small}
\begin{align}
  g\colon S\to&\IR\\
  \sum_{i=1}^{n}\Phi_i(u,f)\cdot v_i \mapsto& %
  \sum_{i=1}^{n}\Phi_i(u,f)\cdot g_i
\end{align}
\end{small}
that depends on each basis function $\Phi_i$ twice, once to describe the function $g$ and once to describe the function domain $S$. As a result, the gradient of a surface function is
\begin{small}
\begin{align}
  \nabla g(X(u,f))=JX(u,f) (G^{-1}(u,f)
    \sum_{i=1}^{n}\nabla\Phi_i(u,f)\cdot g_i)\nonumber,
\end{align}
\end{small}
where $JX$ is the Jacobian of the surface mapping $X$ with respect to $u$.

When we use this parametrization to compute the mass and stiffness matrix, we need to evaluate the first fundamental form of $S$ at a point $x\in S$ explicitly. Let $f\in\cI_F$ be an arbitrary face index, then the first fundamental form at surface point $X(u,f)$ is 
\begin{align}
G(u,f)=JX(u,f)^\top JX(u,f).
\end{align}
We compute the surface integrals in the mass and stiffness matrices by pulling-back the integrand to the reference quad as follows:
\begin{align*}
  \int_S g(s)\ds = \sum_{f\in\cI_F}\int_\square g\circ X(u,f)\sqrt{\det
    G(u,f)}\du
\end{align*}
We can compute the integral over the quad with arbitrary accuracy with high-order quadrature approximations. Experimentally, we obtain reliable mass and stiffness matrices with simple $3\times 3$ Gaussian quadrature.

Note that when the subdivision surface $S$ is derived from a control mesh with $n$ vertices, $D_0, D_1, D_2\in\IR^{n\times n}$. An eigenfunction $g_v$ of $\Delta$ is then obtained by solving an $n\times n$ generalized eigen problem and by parameterizing $g_v=\sum_{i=1}^n v_i\Phi_i$. In our figures, we use the refined $\Mesh^k$ mesh and the refined eigenfunction $g_v^k$ for visualization. That is, $g_v^k$ is not derived from the huge mass and stiffness matrices of $\Mesh^k$ but from the projection of $g_v\in C^1(S)$ onto the mesh $\Mesh^k$.

\subsection{Geodesic Distances}
The classic approach to computing the geodesic distance $\chi$ to a surface point $S(u_0)$ is to solve the eikonal equation
\begin{align} 
\Vert \nabla \chi \Vert  = 1 \label{eq:eikonal}
\end{align}
subject to boundary conditions $\chi(u_0) = 0$. This results in a nonlinear hyperbolic PDE that is computationally expensive to solve. To alleviate the computational burden, \cite{Crane2013} proposes a method to approximately solve \eqref{eq:eikonal} in terms of the Heat equation. This method first computes a function $f\in H^1(S)$ whose gradient is parallel to $\nabla \chi$ by integrating the Heat flow $\dot{f} = \Delta f$. It then approximates $\nabla \chi$ by the vector field $X = -\frac{\nabla f}{\Vert \nabla f \Vert}$ by observing that $\nabla \chi$ is parallel to $\nabla f$ and has unit norm as a result of \eqref{eq:eikonal}. The method then finds an approximate $\chi$ by minimizing $\int_{S} \Vert \nabla \chi - X \Vert^2$ through its Euler-Lagrange equation $ \Delta \chi = \nabla\cdot X$. As all the computations are formulated in terms of the Heat kernel $\Delta$, the method can be directly use to compute approximate geodesics in our subdivision surface framework. We refer to \cite{Crane2013} for the details of the method.

\section{Related Work} \label{sec:related_lierature}

Subdivision surfaces are a standard representation of shapes in computer graphics and animation, but they have largely been neglected in vision due to the challenges of working with noisy measurements that favor convex formulations and implicit representations. 

Computer graphics techniques fit a subdivision surface by minimize the sum of squared distances between the subdivision surface and the vertices of a clean and regular mesh~\cite{Hoppe1994,Ma2002}. Research has focused into efficient ways of minimizing this non-convex energy by gradient-based techniques \cite{Zheng2012}, quasi-Newton methods \cite{Marinov2005,Wang2006,Liu2008}, and sequential convex programing \cite{Marinov2005,Lavoue2005}. A very effective approach iteratively approximates the squared distance to the surface with a quadratic penalty \cite{Pottmann2003a} and minimizes the resulting convex energy. Our method method adapts this strategy to the robust energies necessary to fit a surface to noisy data and outperform the squared distance penalty of \cite{Marinov2005} in three points: 1) We use robust data terms to account for noise and outliers in the input data. A robust energy also reduces the effects of errors in the correspondences (between input samples and the surface point closest them) and the quadratic distance approximation. This is improves the accuracy of the surface independently of the noise present in the data, as our experiments show. 2) We use a second-order approximation to the squared distance function, instead of the first order of \cite{Marinov2005}, to take into account curvature information and fit better the small-scale structures of the surface. 3) We penalize deviations of the tangent space to improve the appearance of sharp creases. 

Computer Vision, on the other hand, has focused on developing energy models to fit subdivision surfaces to noisy data~\cite{Taylor2016a,Cashman2013,Ili2006,Jaimez2017}, but has ignored the impact of outliers and the optimization on the accuracy of the surface. \cite{Ili2006} fits a sub vision surface to range data by solving a weighted least-squares problem that assumes optimal correspondences between input point and domain parameters. This assumption is removed in \cite{Taylor2016a,Cashman2013,Jaimez2017} by explicitly optimizing the correspondences with least-squares models that preprocess the data to eliminate outliers and can only handle gross outliers. These methods also assume the topology of the surface to be know and initialize it with a manually-designed control mesh \cite{Taylor2016a,Cashman2013} or a sphere~\cite{Jaimez2017}. The sphere initialization is flexible but oversmooths the surface because to prevents self intersections as surface evolves. In terms of optimization, all these techniques use a Levenberg-Marquardt (LM) algorithm that estimates jointly the vertices of the control mesh and the correspondences between the input points and surface samples. This increases the size of the problem and slows the optimization because gradient-based techniques, like LM, only update the surface correspondences to adjacent faces of the control mesh. Our majorize-minimize (MM) algorithm places no limits on these updates and can jump to different valleys of the energy landscape.

Our application of subdivision surfaces to shape analysis builds on the work of \cite{Juettler-et-al-16,Burkhart2010,Wawrzinek2016,Goes2016} that investigate the accuracy of computing mass and stiffness matrices on a subdivision surface, but do not consider the estimation of wave-kernel signatures, geodesics, or shape matches in the subdivision framework.



\section{Experiments} \label{sec:experiments}

\subsection*{Experiments on Surface Compression}

\begin{figure*}
  \centering
    \begin{tabular}{cccccc}
      \includegraphics[trim=17cm 0cm 17cm 0cm, clip=true, width = 0.09\textwidth]{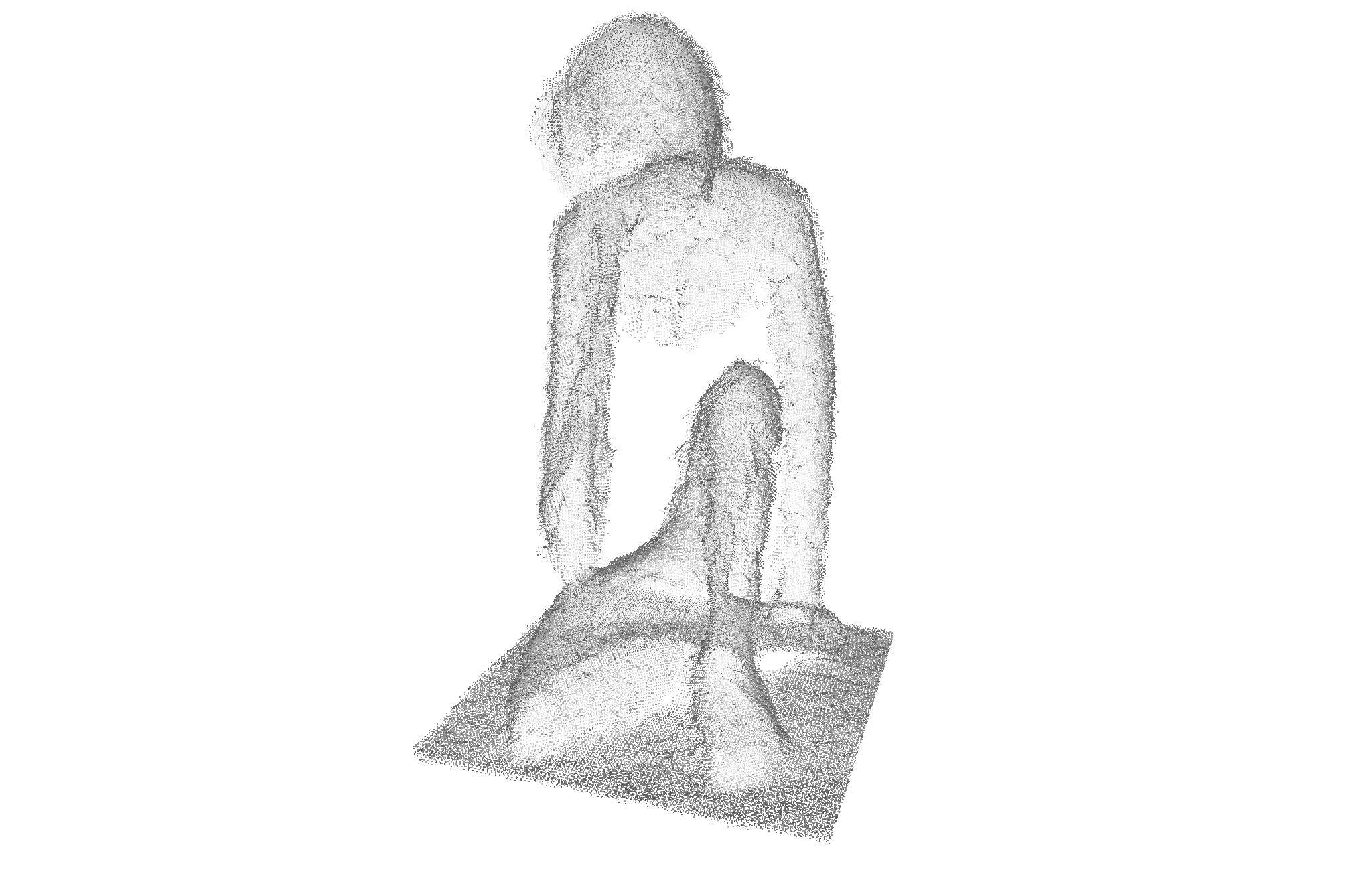}   &
      \includegraphics[trim=17cm 0cm 17cm 0cm, clip=true, width = 0.09\textwidth]{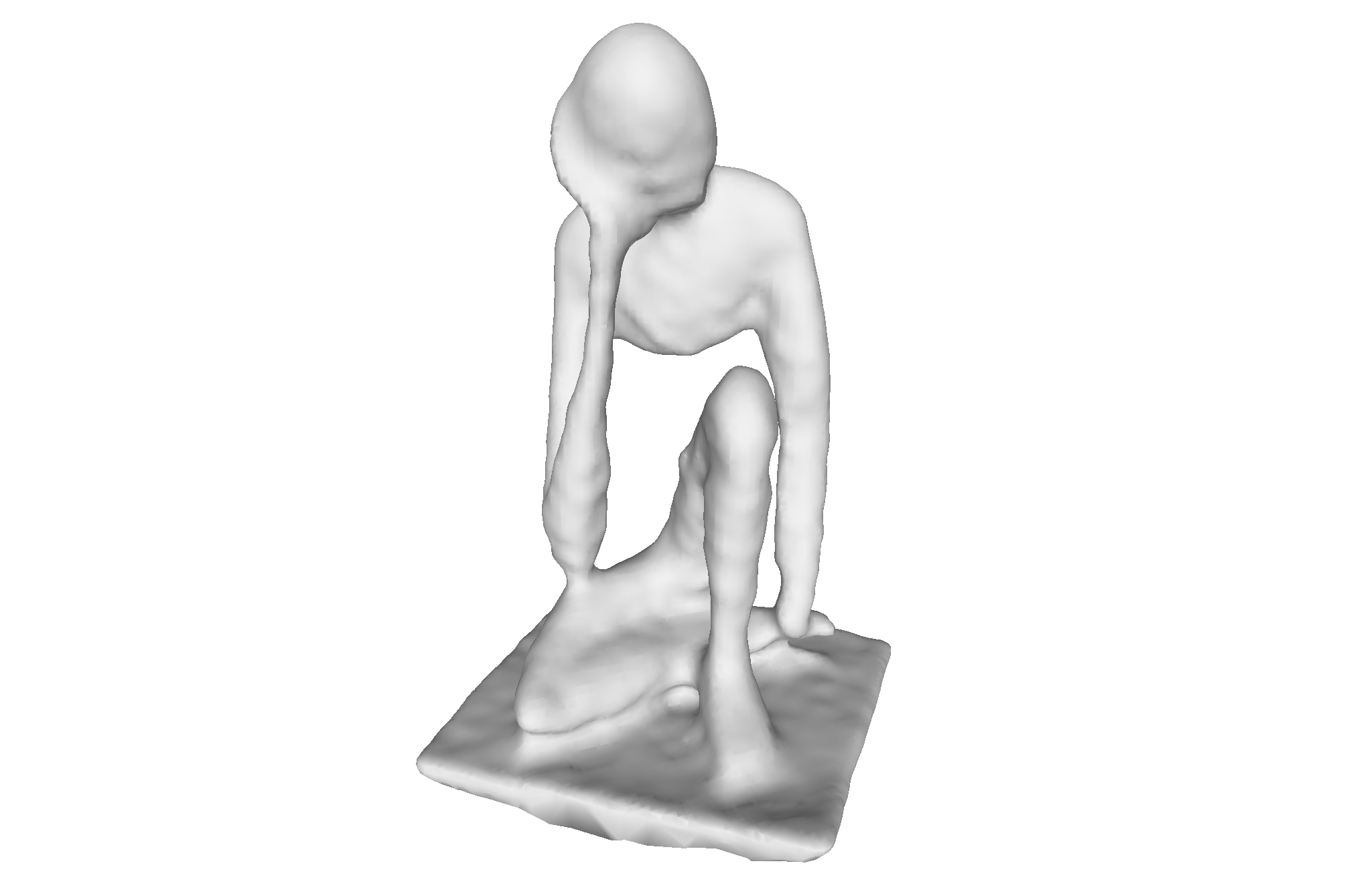}   &
      \includegraphics[trim=17cm 0cm 17cm 0cm, clip=true, width = 0.09\textwidth]{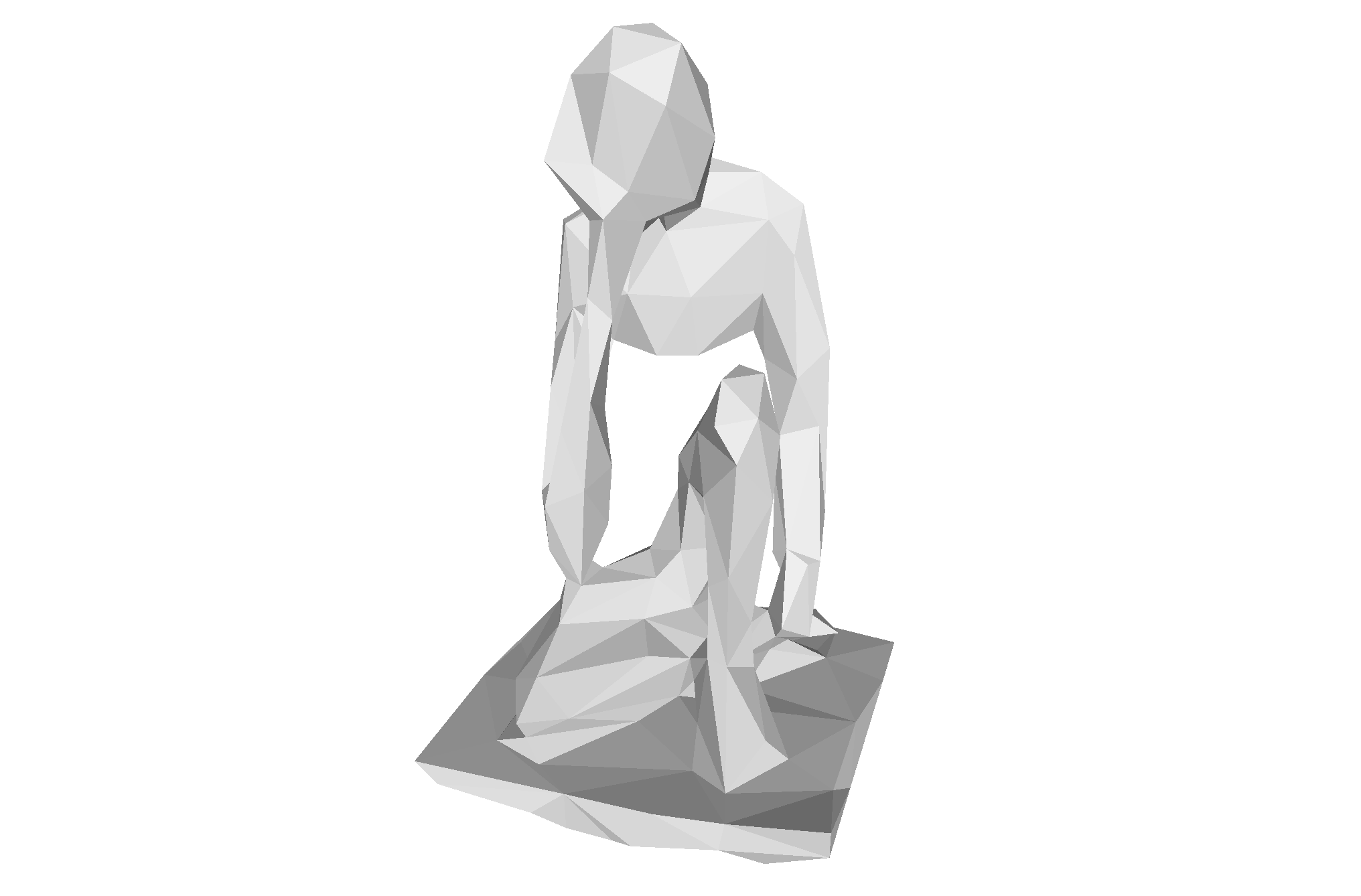}   &
      \includegraphics[trim=17cm 0cm 17cm 0cm, clip=true, width = 0.09\textwidth]{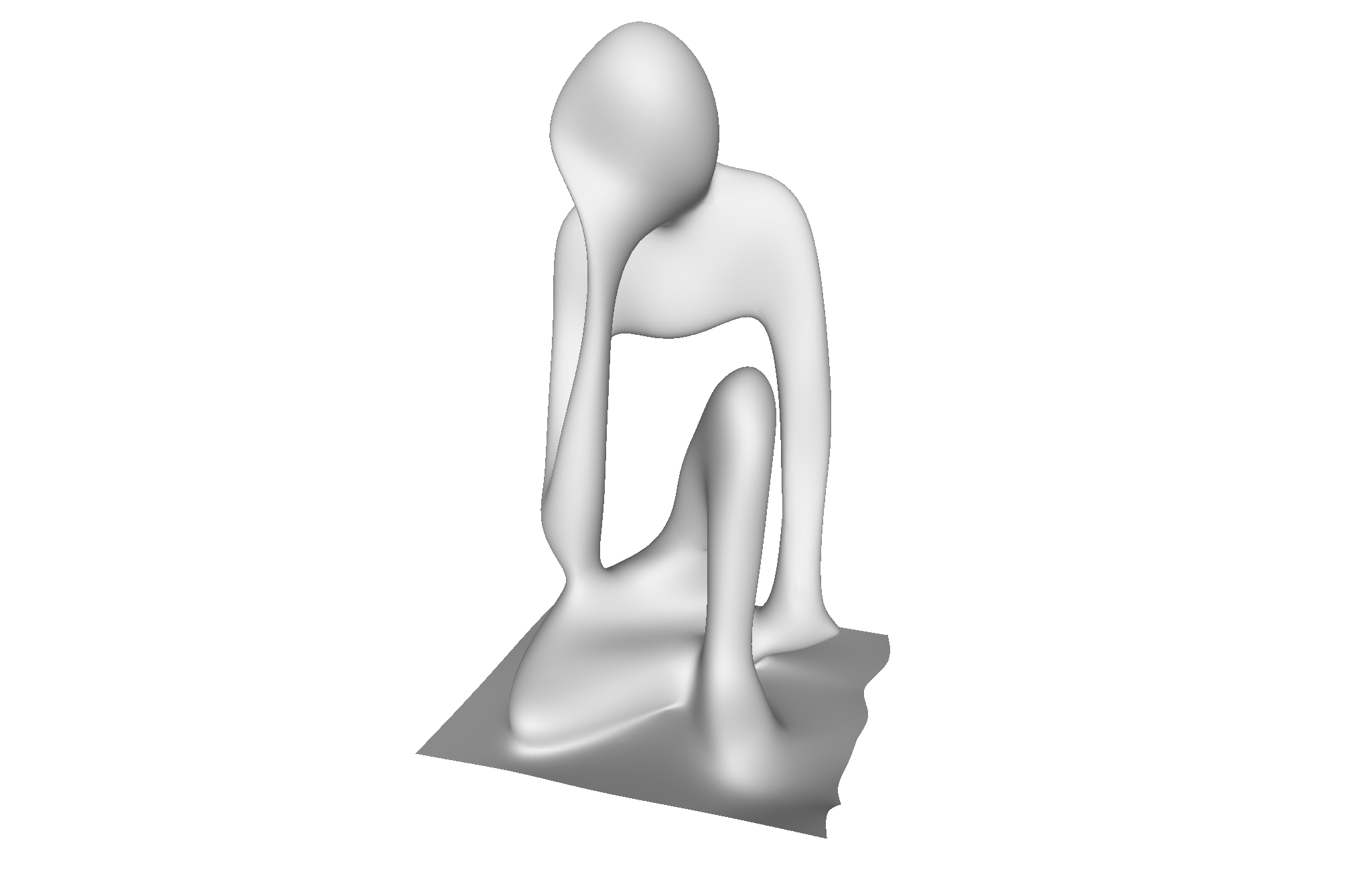} &
      \includegraphics[trim=17cm 0cm 17cm 0cm, clip=true, width = 0.09\textwidth]{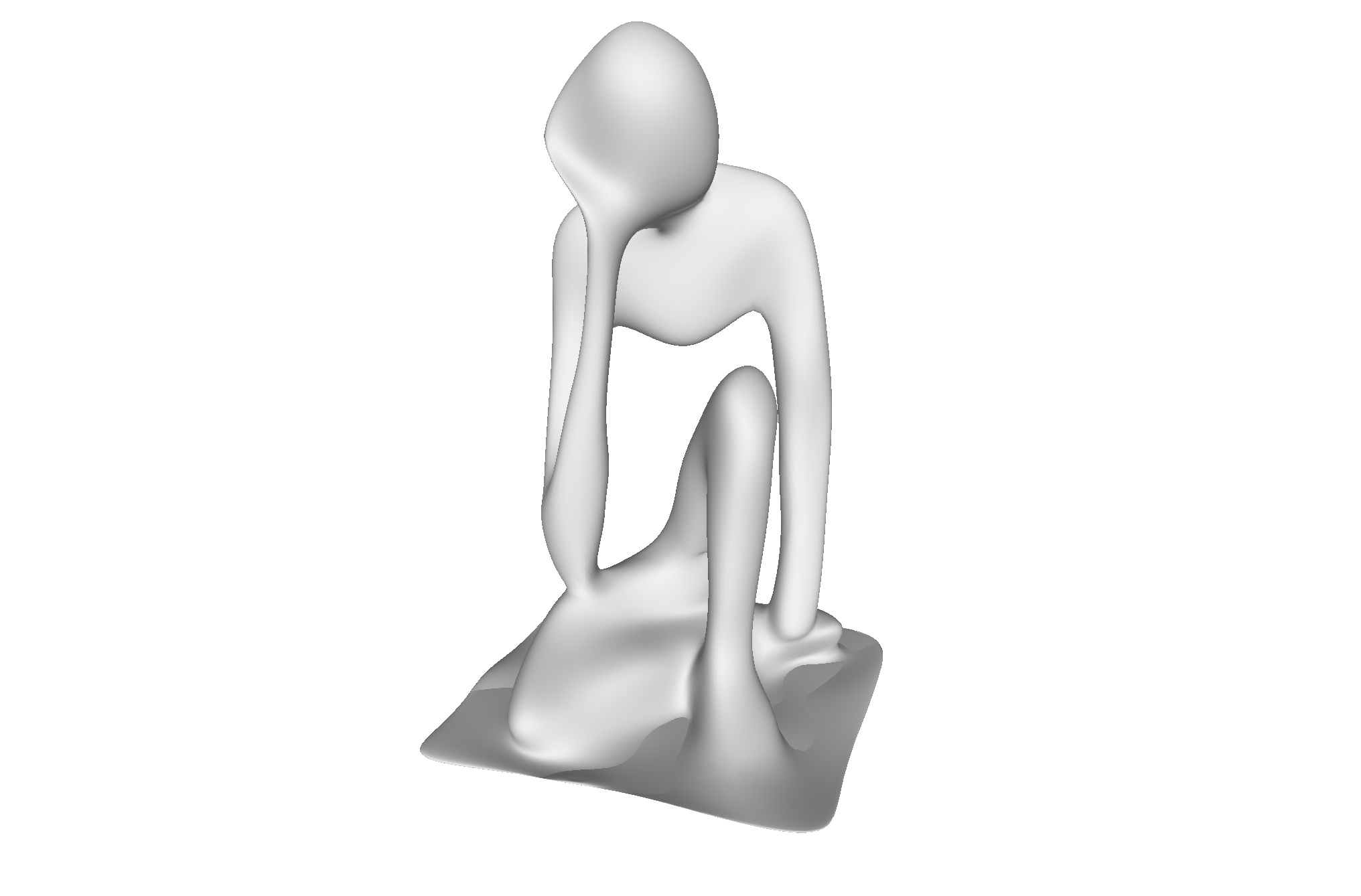} &
     \includegraphics[trim=2cm 0cm 1.75cm 0cm, clip=true, width = 0.09\textwidth]{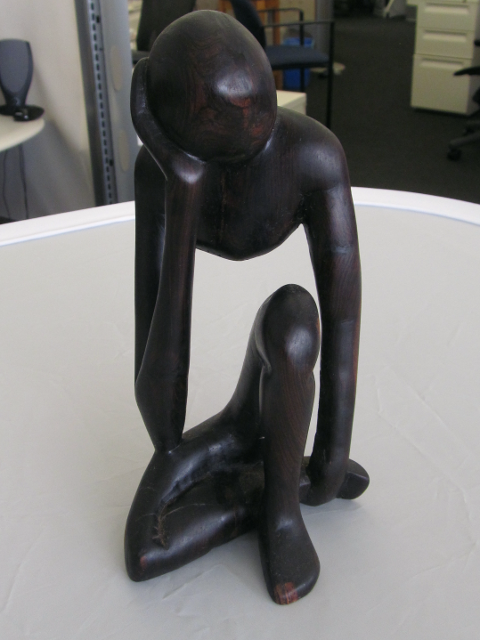}   \\      
      Point cloud & %
      Poisson surface&%
      collapsed Poisson&%
      \cite{Marinov2005} subdiv. surf.&
      our subdiv. surf. &
      recognizable \\
      139553 points & %
      20932 vertices &%
      250 vertices &%
      250 vertices &
      250 vertices &
      picture
    \end{tabular}
        \begin{tabular}{ccccc}        
      \includegraphics[trim=18cm 1cm 18cm 0cm, clip=true, width = 0.15\textwidth]{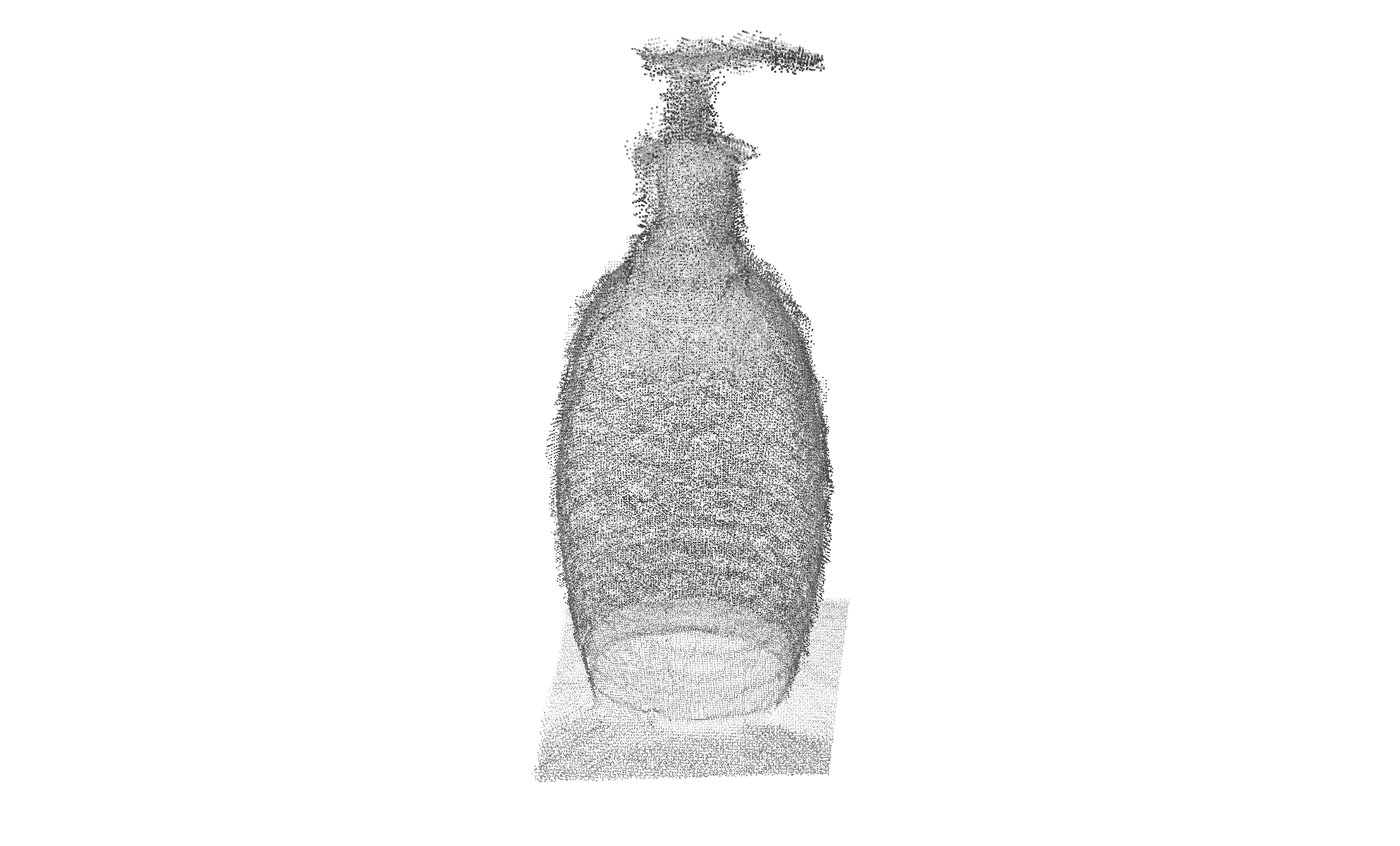}   &     
      \includegraphics[trim=18cm 0cm 18cm 0cm, clip=true, width = 0.15\textwidth]{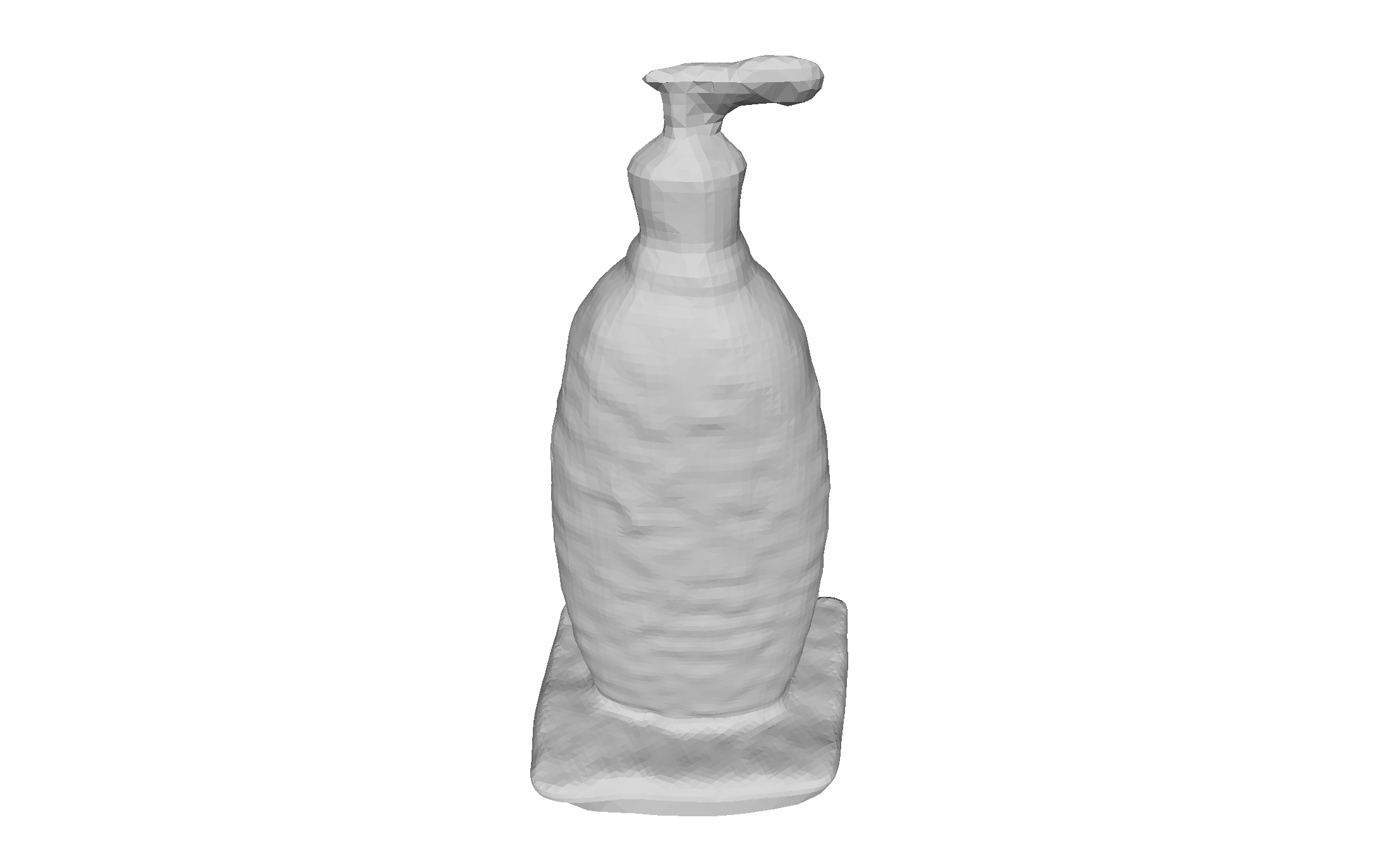}    &
      \includegraphics[trim=18cm 0cm 18cm 0cm, clip=true, width = 0.15\textwidth]{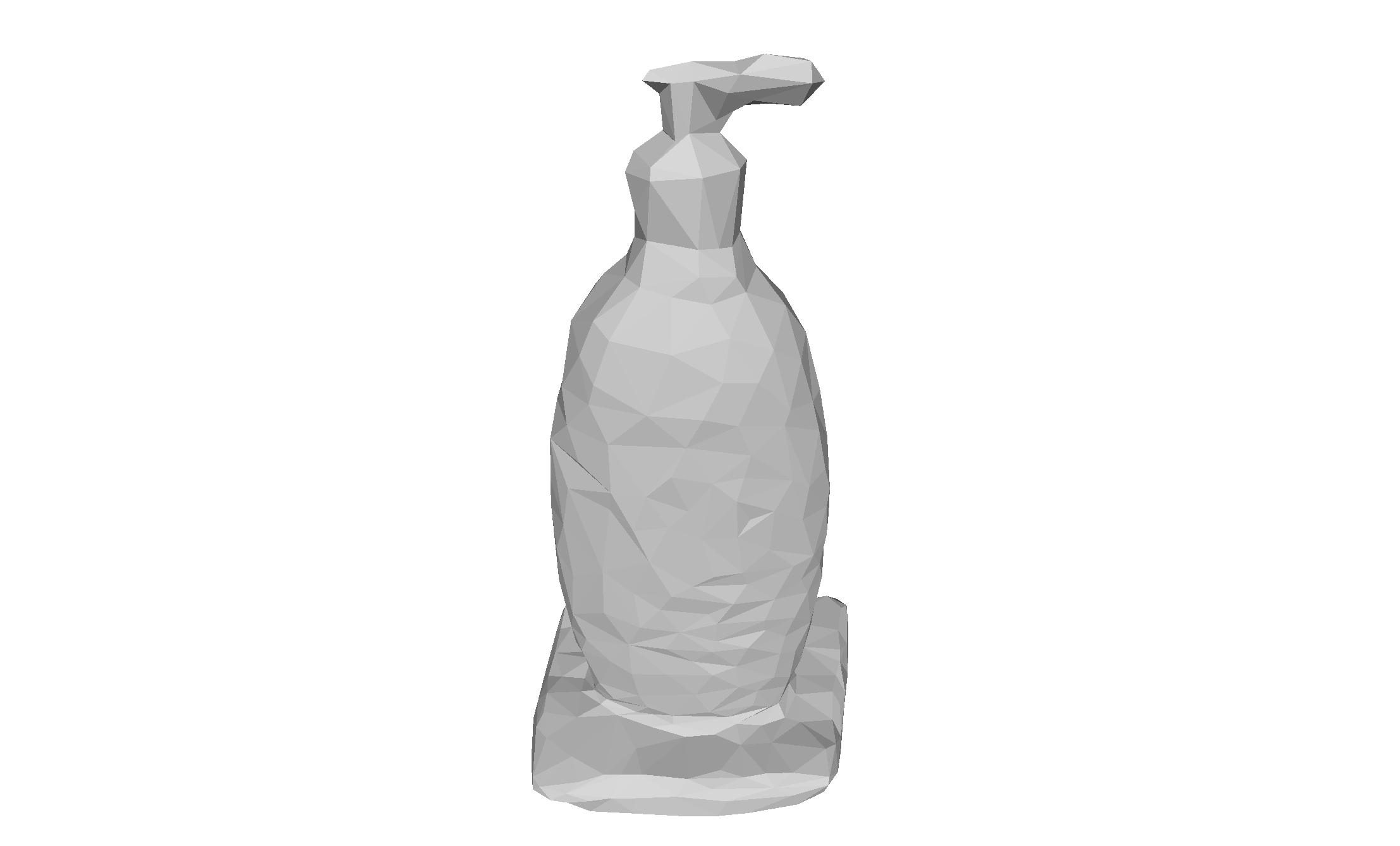}   &
      \includegraphics[trim=18cm 0cm 18cm 0cm, clip=true, width = 0.15\textwidth]{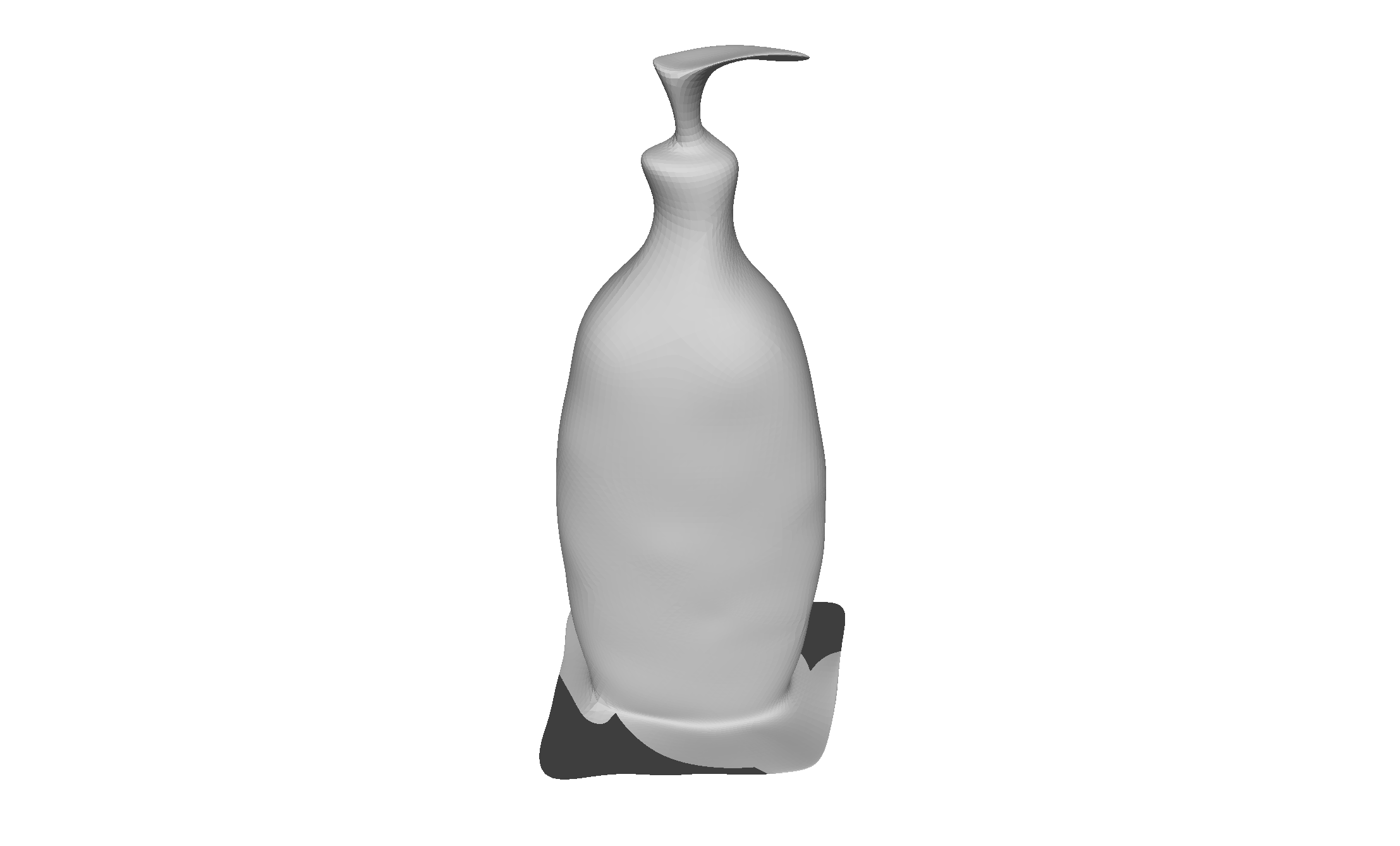} &
      \includegraphics[trim=18cm 0cm 18cm 0cm, clip=true, width = 0.15\textwidth]{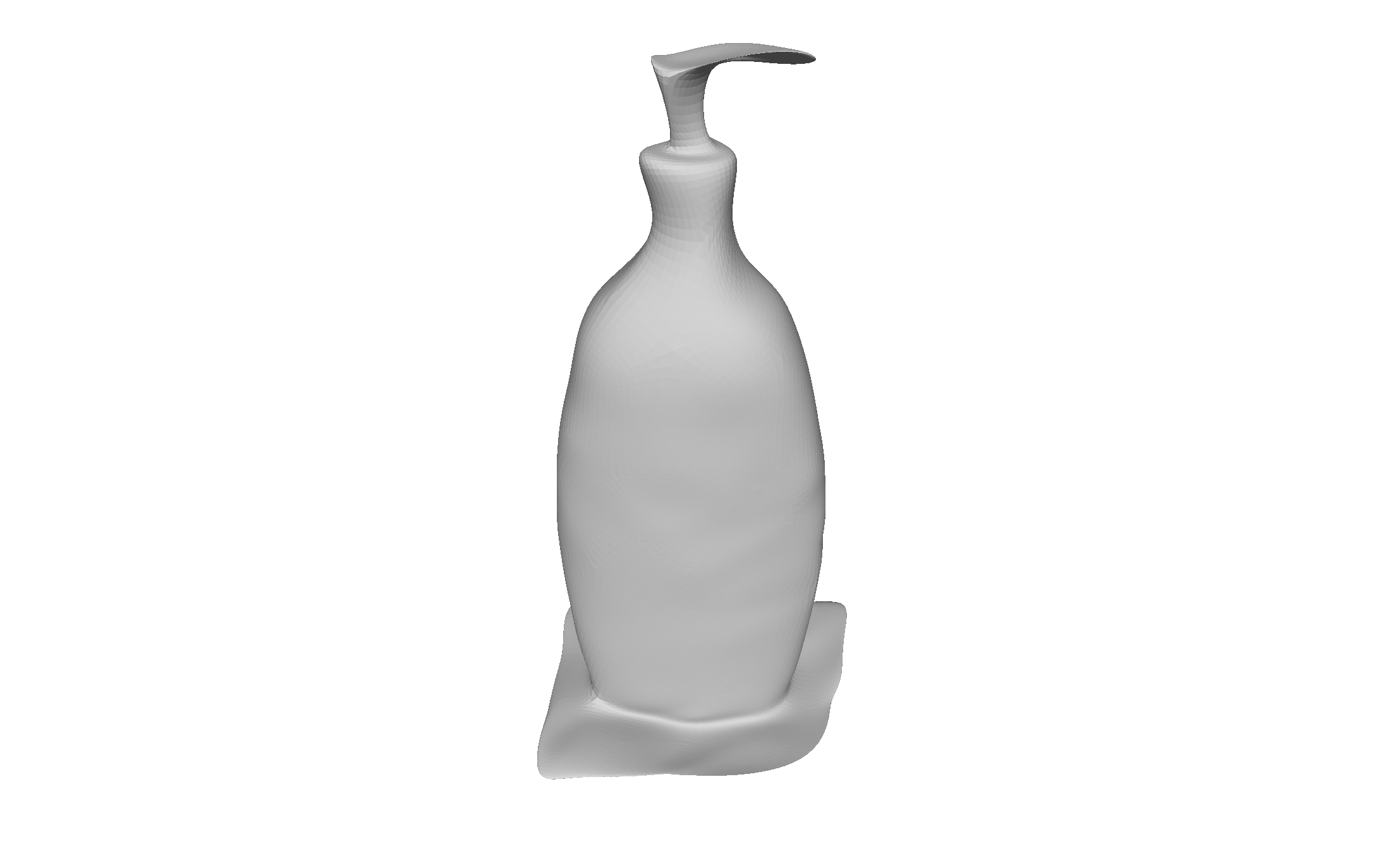}  \\
      Point cloud & %
      Poisson surface&%
      collapsed Poisson&%
      \cite{Marinov2005} subdiv. surf.&
      our subdiv. surf. \\
      107534 points & %
      12928 vertices &%
      250 vertices &%
      250 vertices &
      250 vertices \\
    \end{tabular}
  \caption{Surfaces reconstructed from a Kinect point cloud (columns 1)
    with a state of the art Poisson surface reconstruction method
    \cite{Kazhdan2013} (column 2) and with the proposed
    subdivision surface (columns 5). Our technique reconstructs and
    represents the surface with an accuracy comparable to a
    high-resolution mesh while compressing the representation to less
    than 2\%. We reconstruct the nozzle of the bottle with higher accuracy thanks to our robust emery model. The subdivision surface of \cite{Marinov2005} (columns 4) shrinks the statue's V-shaped torso while it oversmooths its hands and produces artifacts at the base of the bottle where samples are irregularly spaced and overlap.  }
    \label{fig:thinker}
\end{figure*}

Our experiments compare our subdivision representation to triangular meshes with two different settings: reconstructing a surface from noisy Kinect pointclouds, and approximating a fine artifact-free mesh with small-scale structures. For all our experiments we set $q=1.3$ and manually choose $\alpha,\beta$ to fit each shape and model ( either our model or the state-of-the-art methods we compare to ). When the input is a mesh, we set these parameters to minimize the Haussdorff distance between the subdivsision surface and the input mesh.

Figures \ref{fig:geodesics0}, \ref{fig:thinker} show that our subdivision surface reconstructs a surface from a noisy pointcloud as accurately as the mesh obtained by the state-of-the-art reconstruction technique~\cite{Kazhdan2013} but reduces the number of representation variables to $2\%$ of the mesh size. Our subdivision surfaces preserve the smooth regions and sharp corners (torso and hands of Figure \ref{fig:thinker}), and texture details (human hair in Figure \ref{fig:geodesics0}) of the fine mesh that are lost in a triangular mesh of the same size obtained by quadratic edge collapse. 
  
\begin{figure*}
  \centering
    \begin{tabular}{cccc}
      \includegraphics[width = 0.20\textwidth]{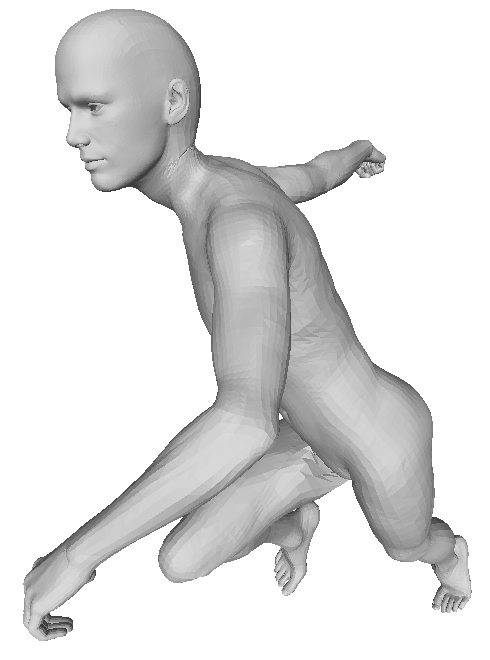}&
      \includegraphics[width = 0.20\textwidth]{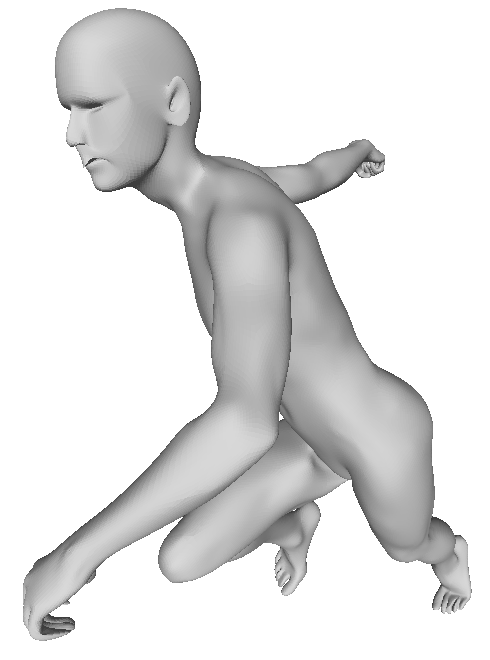}&
       \includegraphics[width = 0.20\textwidth]{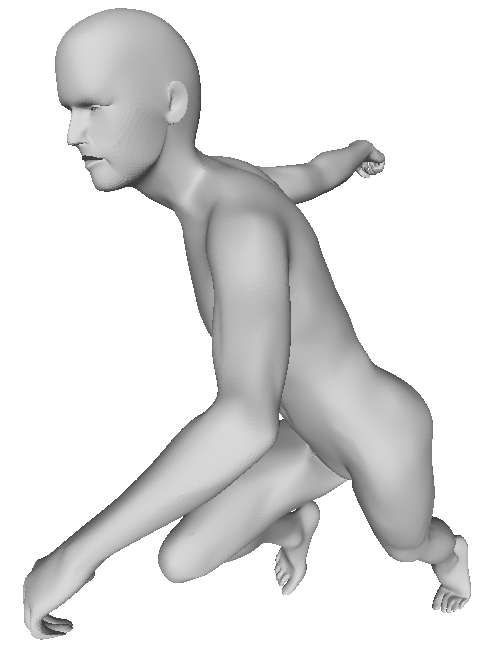} &
      \includegraphics[width = 0.20\textwidth]{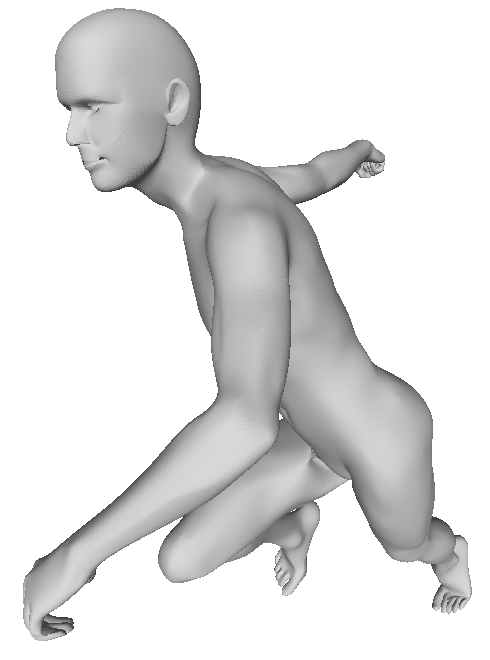}\\
     TOSCA mesh, 52565 vertices & \cite{Marinov2005} subdiv. surface & LM solver & our subdiv. surface\\           
      \includegraphics[trim=0cm 16cm 8cm 0cm, clip=true, width = 0.20\textwidth]{david13_tosca}&
      \includegraphics[trim=0cm 16cm 8cm 0cm, clip=true, width = 0.20\textwidth]{david13_1storder}&
       \includegraphics[trim=0cm 16cm 8cm 0cm, clip=true, width = 0.20\textwidth]{david13_LM} &
      \includegraphics[trim=0cm 16cm 8cm 0cm, clip=true, width = 0.20\textwidth]{david13_our}\\
		zoom on TOSCA mesh & zoom on \cite{Marinov2005} surface & zoom on LM solver & zoom on our surface\\  		
      \includegraphics[width = 0.20\textwidth]{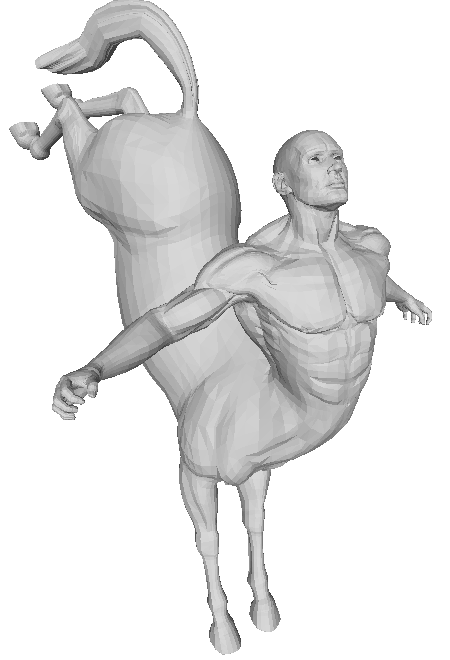}&
      \includegraphics[width = 0.20\textwidth]{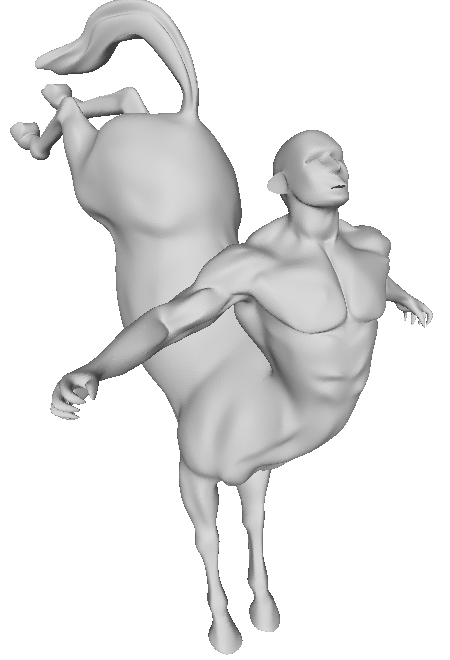}&
      \includegraphics[width = 0.20\textwidth]{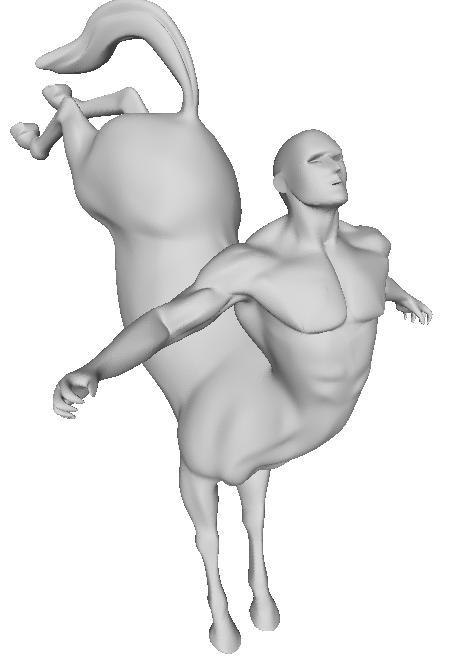}&
      \includegraphics[width = 0.20\textwidth]{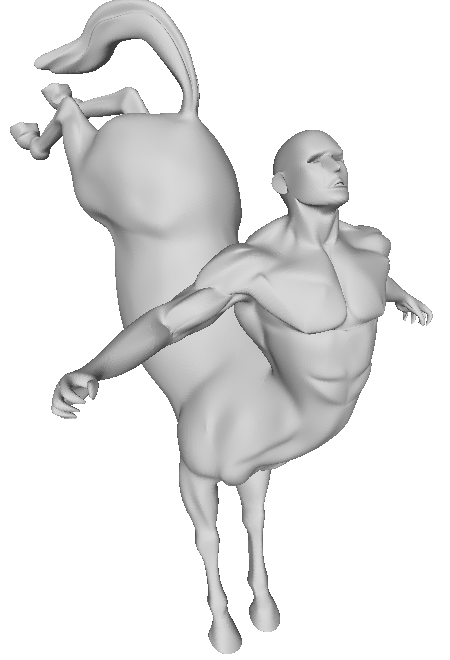}\\
      TOSCA mesh, 15768 vertices& \cite{Marinov2005} subdiv. surface & LM solver & our subdiv. surface\\           
      \includegraphics[trim=9cm 14cm 2cm 4.5cm, clip=true, width = 0.20\textwidth]{centaur3_tosca}&
      \includegraphics[trim=9cm 14cm 2cm 4.5cm, clip=true, width = 0.20\textwidth]{centaur3_1storder}&
      \includegraphics[trim=9cm 14cm 2cm 4.5cm, clip=true, width = 0.20\textwidth]{centaur3_LM}&
      \includegraphics[trim=9cm 14cm 2cm 4.5cm, clip=true, width = 0.20\textwidth]{centaur3_our}\\
      zoom on TOSCA mesh & zoom on \cite{Marinov2005} surface & zoom on LM solver & zoom on our surface\\           
    \end{tabular}
  \caption{Comparison to state-of-the-art: high-resolution input mesh
    (column 1) mesh and subdivision surfaces fitted with \cite{Marinov2005} (column 2) and our proposed model optimized with Levenberg-Marquard (column 3) and our proposed optimization algorithm (column 4). Comparing columns 2 and 4, shows the benefits of using a robust $q$-power distance function and a second-order, instead of a first-order, approximation of the squared distance to the surface to fit thin surface structures like the ears or lips. Comparing columns 3 and 4 show the benefits of quadratic sequential program to avoid small-scale local minima close to the initialization that do not capture the shape details of the nose, lips, or mesh ears. All the subdivision surfaces have 1700 control vertices and are visualized by refining $\Mesh_0$ 3 times.}
  \label{fig:sds-compare-literature0}
\end{figure*}

\begin{figure*}
\centering
\begin{tabular}{cccc}
      \includegraphics[width=0.20\textwidth]{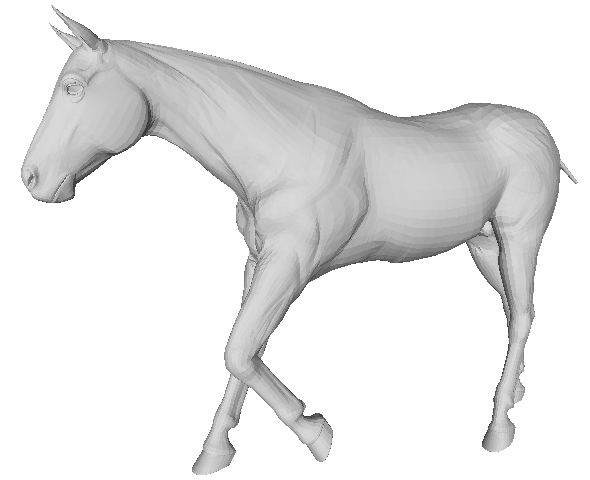}&
      \includegraphics[width=0.20\textwidth]{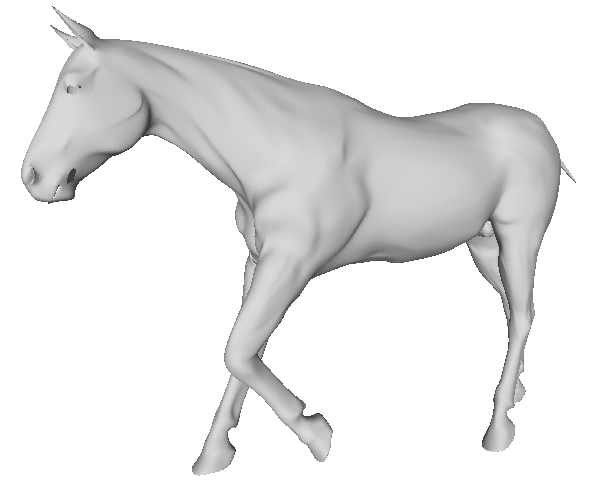}&
      \includegraphics[width=0.20\textwidth]{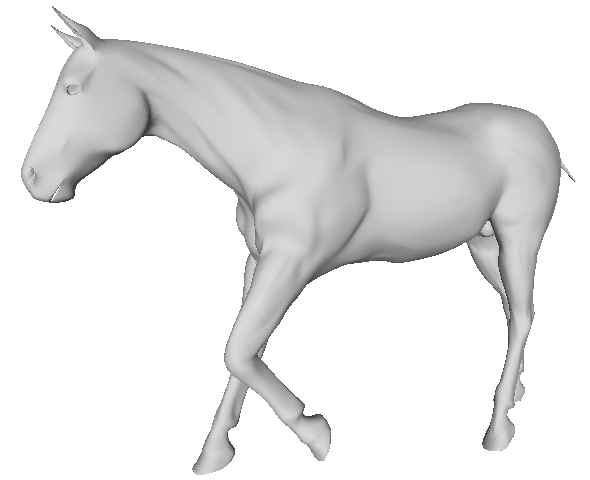}&
      \includegraphics[width=0.20\textwidth]{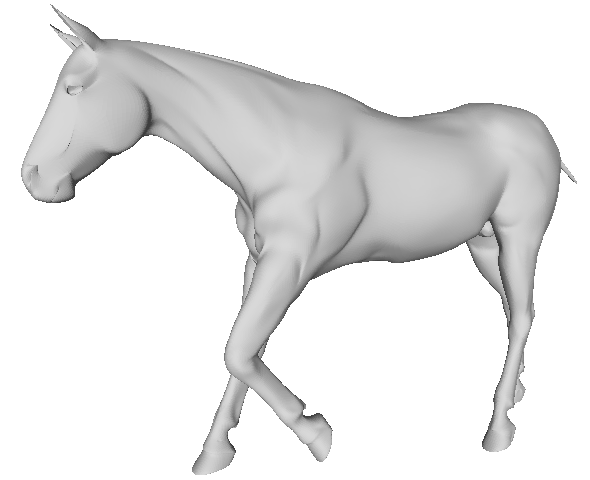}\\
      TOSCA mesh, 19248 vertices & \cite{Marinov2005} subdiv. surface & LM solver  & our subdiv. surface \\       
      \includegraphics[trim=0cm 9.5cm 14cm 0cm, clip=true, width=0.20\textwidth]{horse6_tosca}&
      \includegraphics[trim=0cm 9.5cm 14cm 0cm, clip=true, width=0.20\textwidth]{horse6_1storder}&
      \includegraphics[trim=0cm 9.5cm 14cm 0cm, clip=true, width=0.20\textwidth]{horse6_LM}&
      \includegraphics[trim=0cm 9.5cm 14cm 0cm, clip=true, width=0.20\textwidth]{horse6_our}\\
      zoom on TOSCA mesh & zoom on \cite{Marinov2005} surface & zoom on LM solver  & zoom on our surface\\              
      \includegraphics[trim=0cm 6cm 0cm 0cm, clip=true,height = 0.4\textwidth]{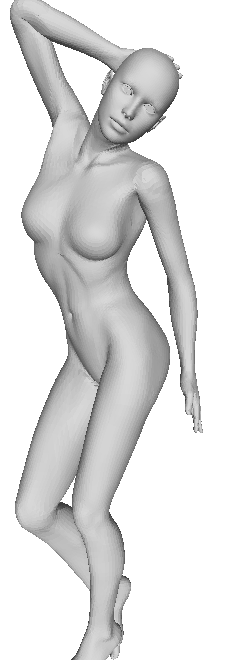}&
      \includegraphics[trim=0cm 6cm 0cm 0cm, clip=true, height = 0.4\textwidth]{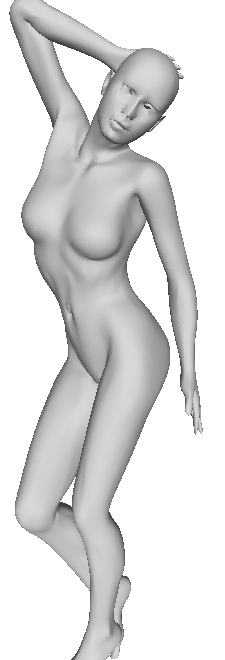}&
       \includegraphics[trim=0cm 6cm 0cm 0cm, clip=true, height = 0.4\textwidth]{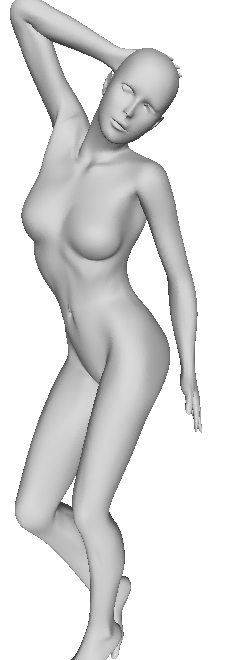} &
      \includegraphics[trim=0cm 6cm 0cm 0cm, clip=true, height = 0.4\textwidth]{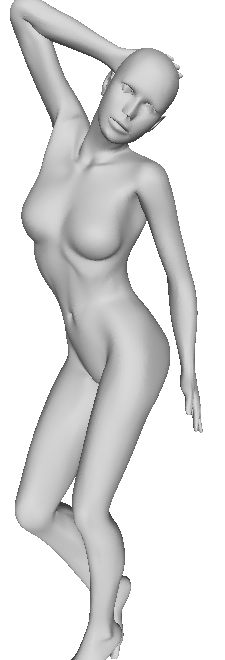}\\
      TOSCA mesh, 45659 vertices& \cite{Marinov2005} subdiv. surface & LM solver  & our subdiv. surface\\
      \includegraphics[trim=2cm 18cm 1cm 0.5cm, clip=true, width=0.20\textwidth]{victoria2_tosca}&
      \includegraphics[trim=2cm 18cm 1cm 0.5cm, clip=true, width=0.20\textwidth]{victoria2_1storder}&
       \includegraphics[trim=2cm 18cm 1cm 0.5cm, clip=true, width=0.20\textwidth]{victoria2_LM} &
      \includegraphics[trim=2cm 18cm 1cm 0.5cm, clip=true, width=0.20\textwidth]{victoria2_our}\\
      zoom on TOSCA mesh & zoom on \cite{Marinov2005} surface & zoom on LM solver  & zoom on our surface\\        
    \end{tabular}
  \caption{Comparison to state-of-the-art: high-resolution input mesh
    (column 1) mesh and subdivision surfaces fitted with \cite{Marinov2005} (column 2) and our proposed model optimized with Levenberg-Marquard (column 3) and our proposed optimization algorithm (column 4). Comparing columns 2 and 4, shows the benefits of using a robust $q$-power distance function to fit thin surface structures like the lips or mesh eyelids without artifacts. Comparing columns 3 and 4 show the benefits of quadratic sequential program to avoid small-scale local minima close to the initialization that do not capture the shape details of the nose, lips, or mesh ears. All the subdivision surfaces have 1700 control vertices and are visualized by refining $\Mesh_0$ 3 times.}
  \label{fig:sds-compare-literature1}
\end{figure*}

\begin{figure*}
  \centering
    \begin{tabular}{cccc}
      \includegraphics[width=0.20\textwidth]{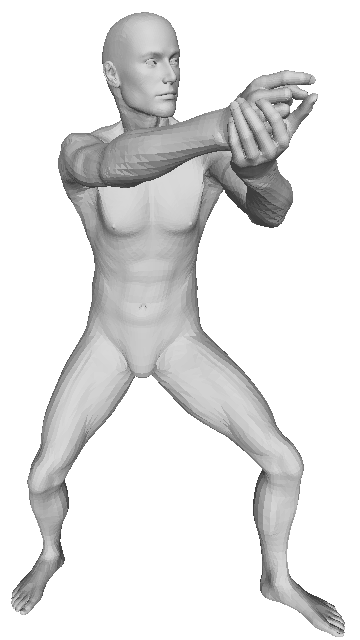}&
      \includegraphics[width=0.20\textwidth]{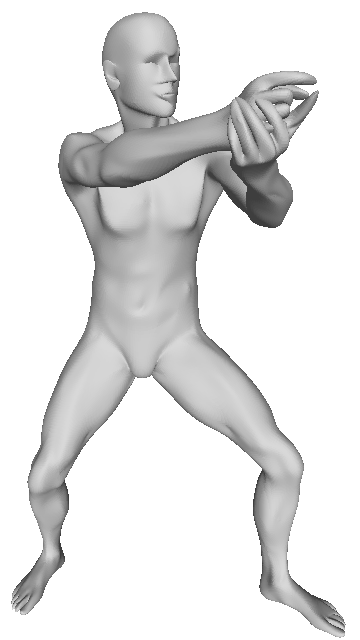}&
      \includegraphics[width=0.20\textwidth]{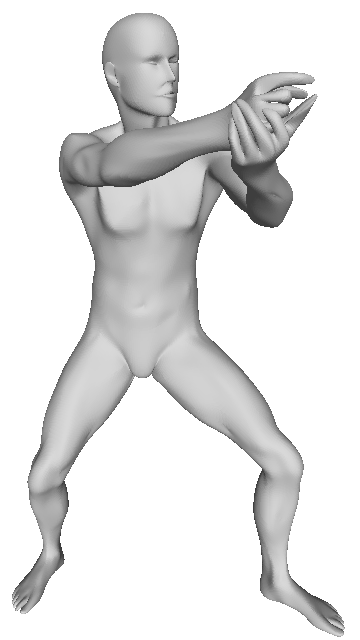}&
      \includegraphics[width=0.20\textwidth]{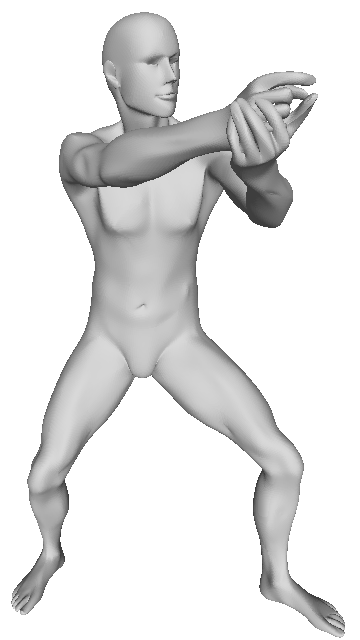}\\
      TOSCA mesh, 52565 vertices& \cite{Marinov2005} subdiv. surface & LM solver  & our subdiv. surface\\  
\includegraphics[trim=2cm 14cm 1cm 0.5cm, clip=true, width=0.20\textwidth]{michael13_tosca}&
      \includegraphics[trim=2cm 14cm 1cm 0.5cm, clip=true, width=0.20\textwidth]{michael13_1storder}&
      \includegraphics[trim=2cm 14cm 1cm 0.5cm, clip=true, width=0.20\textwidth]{michael13_LM}&
      \includegraphics[trim=2cm 14cm 1cm 0.5cm, clip=true, width=0.20\textwidth]{michael13_our}\\
     zoom on TOSCA mesh & zoom on \cite{Marinov2005} surface& zoom on LM solver  & zoom on our surface\\  
          \end{tabular}     
  \caption{Comparison to state-of-the-art: high-resolution input mesh
    (column 1) mesh and subdivision surfaces fitted with \cite{Marinov2005} (column 2) and our proposed model optimized with Levenberg-Marquard (column 3) and our proposed optimization algorithm (column 4). Comparing columns 2 and 4, shows the benefits of using the second-order, instead of a first-order, approximation of the squared distance to the surface to capture the curvature details of small structures like the mesh ears. Comparing columns 3 and 4 show the benefits of quadratic sequential program to avoid small-scale local minima close to the initialization that do not capture the shape details of the nose, lips, or mesh eyes. All the subdivision surfaces have 1700 control vertices and are visualized by refining $\Mesh_0$ 3 times.}
  \label{fig:sds-compare-literature2}
\end{figure*}

For all our experiments with meshes, we use the high-resolution
versions of the TOSCA dataset and fitted a subdivision surface with 1700 and 1000 vertices. Tables \ref{tab:distance}--\ref{tab:time} report the execution time and the Hausdorff distance between the input mesh and our compressed subdivision representation for 72 high-resolution TOSCA meshes. Figures \ref{fig:sds-compare-literature0}--\ref{fig:sds-compare-literature2} show the subdivision surfaces fitted to the input meshes with the different models for a qualitative comparison.

\textbf{Comparison to State-of-the-Art Models}
When the input is a noisy pointcloud, our surfaces avoid the artifacts of \cite{Marinov2005} because our energy model is robust to outliers (Figure \ref{fig:thinker}). Robustness also guards us against erroneous correspondences in thin structures of clean meshes. We also fit better small curved structures because we use a second-order distance approximation, instead of the first-order of \cite{Marinov2005}, that takes into account curvature information. This is highlighted in the representation of mouths, eyes, and ears of Figures \ref{fig:sds-compare-literature0}--\ref{fig:sds-compare-literature2}.

\textbf{Comparison to State-of-the-Art Optimization} 
We compare the subdivision surfaces obtained by optimizing our energy model with the proposed algorithm and by employing Levenberg-Marquardt to minimize the MM majorizer with respect to the mesh vertices and the correspondence parameters. Our optimization is an order of magnitude faster and less sensitive to local minima than the LM approach proposed by \cite{Taylor2016a,Cashman2013,Ili2006,Jaimez2017}. As a result, our surfaces evolve further away from their initialization to reproduce the small-scale details of the input meshes (ears and mouths of Figure \ref{fig:sds-compare-literature0}).

Our fitting technique is moderately slower than the quadratic model of \cite{Marinov2005} as a result of our robust energy, as expected. This speed loss is compensated by our improved surface quality, see Table \ref{tab:distance}.

\textbf{Effects of the different terms in our energy model:} Our model always benefits from the robust term while the tangent-fit is slightly detrimental in rare cases, see columns 1--3 in Table \ref{tab:distance}.


\begin{table*}
\vspace{-3cm}
\thisfloatpagestyle{empty}
\setlength\extrarowheight{-2pt}
\centering
\caption{Hausdorff distance (as a \textperthousand  of the bounding box diagonal) between the input mesh and a subdivision surface with 1000 and 1700 control vertices. Columns 1--3 and 6--8 show the importance of each term in the energy model, while columns 4--5 and 9--10 compare our algorithm to state-of-the-art model of \cite{Marinov2005} and to the Levenberg-Marquardt (LM) optimization advocated by \cite{Taylor2016a,Cashman2013,Jaimez2017}. }
\footnotesize
\begin{tabular}{| l | c c c c c | c c c c c |}
\hline
& \multicolumn{5}{|c|}{1000 control vertices} & \multicolumn{5}{|c|}{1700 control vertices } \\
 mesh       & {$q=2$} & {$\alpha=0$} & {full model}  &  {\cite{Marinov2005}} & {LM solver} & {$q=2$} & {$\alpha=0$} & {full model}  &  {\cite{Marinov2005}}  & {LM solver}
   \\
\hline
cat0 & 0.180 & 0.176 & \textbf{0.158} & 0.209 & 0.237 & 0.113 & \textbf{0.103} & 0.103 & 0.123 & 0.141 \\
cat1 & 0.251 & 0.237 & \textbf{0.218} & 0.295 & 0.327 & 0.160 & 0.155 & \textbf{0.148} & 0.192 & 0.192 \\
cat10 & 0.213 & 0.201 & \textbf{0.019} & 0.246 & 0.251 & 0.132 & 0.123 & \textbf{0.121} & 0.155 & 0.184 \\
cat2 & 0.436 & 0.390 & \textbf{0.352} & 0.475 & 0.572 & 0.272 & 0.256 & \textbf{0.240} & 0.300 & 0.366 \\
cat3 & 0.210 & 0.200 & \textbf{0.183} & 0.259 & 0.263 & 0.128 & 0.127 & \textbf{0.126} & 0.155 & 0.165 \\
cat4 & 0.254 & 0.247 & \textbf{0.232} & 0.303 & 0.345 & 0.168 & 0.166 & \textbf{0.154} & 0.195 & 0.209 \\
cat5 & 0.219 & 0.208 & \textbf{0.201} & 0.254 & 0.286 & 0.139 & 0.131 & \textbf{0.129} & 0.166 & 0.165 \\
cat6 & 0.390 & 0.390 & \textbf{0.359} & 0.454 & 0.443 & 0.260 & 0.257 & \textbf{0.240} & 0.313 & 0.313 \\
cat7 & 0.437 & 0.391 & \textbf{0.373} & 0.468 & 0.483 & 0.268 & 0.253 & \textbf{0.246} & 0.304 & 0.444 \\
cat8 & 0.233 & 0.209 & \textbf{0.197} & 0.251 & 0.261 & 0.156 & 0.145 & \textbf{0.139} & 0.168 & 0.189 \\
cat9 & 0.283 & 0.263 & \textbf{0.248} & 0.336 & 0.356 & 0.179 & 0.171 & \textbf{0.166} & 0.204 & 0.238 \\
centaur0 & 0.448 & 0.432 & \textbf{0.400} & 0.490 & 0.589 & 0.305 & 0.303 & \textbf{0.285} & 0.347 & 0.357 \\
centaur1 & 0.508 & 0.449 & \textbf{0.424} & 0.507 & 0.526 & 0.323 & 0.321 & \textbf{0.298} & 0.361 & 0.374 \\
centaur2 & 0.632 & 0.578 & \textbf{0.553} & 0.653 & 0.733 & 0.402 & 0.403 & \textbf{0.381} & 0.459 & 0.571 \\
centaur3 & 0.443 & 0.446 & \textbf{0.417} & 0.489 & 0.597 & 0.306 & 0.310 & \textbf{0.289} & 0.352 & 0.361 \\
centaur4 & 0.545 & 0.488 & \textbf{0.478} & 0.550 & 0.651 & 0.336 & 0.346 & \textbf{0.320} & 0.393 & 0.394 \\
centaur5 & 0.474 & 0.466 & \textbf{0.441} & 0.521 & 0.563 & 0.319 & 0.325 & \textbf{0.304} & 0.371 & 0.422 \\
david0 & 0.274 & 0.250 & \textbf{0.232} & 0.273 & 0.402 & 0.174 & 0.171 & \textbf{0.163} & 0.187 & 0.203 \\
david1 & 0.403 & 0.377 & \textbf{0.370} & 0.445 & 0.645 & 0.275 & 0.262 & \textbf{0.255} & 0.296 & 0.502 \\
david10 & 0.396 & 0.329 & \textbf{0.324} & 0.378 & 0.453 & 0.218 & 0.216 & \textbf{0.199} & 0.251 & 0.283 \\
david11 & 0.477 & 0.434 & \textbf{0.421} & 0.479 & 1.004 & 0.313 & 0.293 & \textbf{0.284} & 0.351 & 0.410 \\
david12 & 0.409 & 0.363 & \textbf{0.360} & 0.391 & 0.465 & 0.257 & 0.233 & \textbf{0.219} & 0.272 & 0.381 \\
david13 & 0.349 & 0.341 & \textbf{0.321} & 0.381 & 0.647 & 0.238 & 0.227 & \textbf{0.216} & 0.253 & 0.304 \\
david6 & 0.297 & 0.295 & \textbf{0.272} & 0.333 & 0.343 & 0.185 & 0.186 & \textbf{0.177} & 0.211 & 0.288 \\
dog0 & 0.302 & 0.291 & \textbf{0.272} & 0.349 & 0.424 & 0.195 & 0.184 & \textbf{0.180} & 0.225 & 0.286 \\
dog1 & 0.433 & 0.407 & \textbf{0.398} & 0.470 & 0.624 & 0.269 & 0.264 & \textbf{0.250} & 0.323 & 0.381 \\
dog10 & 0.440 & 0.378 & \textbf{0.369} & 0.469 & 0.539 & 0.244 & 0.246 & \textbf{0.232} & 0.289 & 0.295 \\
dog2 & 0.360 & 0.333 & \textbf{0.311} & 0.425 & 0.421 & 0.214 & 0.207 & \textbf{0.202} & 0.249 & 0.303 \\
dog3 & 0.384 & 0.377 & \textbf{0.362} & 0.453 & 0.463 & 0.273 & 0.253 & \textbf{0.242} & 0.297 & 0.402 \\
dog5 & 0.336 & 0.327 & \textbf{0.309} & 0.391 & 0.449 & 0.209 & 0.205 & \textbf{0.193} & 0.244 & 0.266 \\
dog6 & 0.365 & 0.335 & \textbf{0.328} & 0.408 & 0.455 & 0.221 & \textbf{0.206} & 0.206 & 0.247 & 0.391 \\
dog7 & 0.444 & 0.429 & \textbf{0.413} & 0.496 & 0.628 & 0.288 & 0.280 & \textbf{0.268} & 0.331 & 0.389 \\
dog8 & 0.458 & 0.435 & \textbf{0.408} & 0.523 & 0.598 & 0.288 & 0.287 & \textbf{0.279} & 0.340 & 0.362 \\
gorilla1 & 0.497 & \textbf{0.379} & 0.398 & 0.457 & 0.753 & 0.340 & 0.274 & \textbf{0.274} & 0.312 & 0.449 \\
gorilla14 & 0.854 & \textbf{0.688} & 0.719 & 0.885 & 1.348 & 0.599 & \textbf{0.486} & 0.489 & 0.572 & 0.776 \\
gorilla5 & 0.711 & 0.568 & \textbf{0.546} & 0.689 & 0.880 & 0.446 & \textbf{0.380} & 0.383 & 0.444 & 0.706 \\
gorilla8 & 0.793 & 0.650 & \textbf{0.621} & 0.775 & 1.080 & 0.530 & \textbf{0.449} & 0.451 & 0.538 & 0.863 \\
horse0 & 0.372 & 0.366 & \textbf{0.355} & 0.431 & 1.252 & 0.250 & 0.246 & \textbf{0.233} & 0.290 & 0.317 \\
horse10 & 0.351 & 0.323 & \textbf{0.315} & 0.375 & 1.223 & 0.226 & 0.225 & \textbf{0.216} & 0.274 & 0.294 \\
horse14 & 0.370 & 0.348 & \textbf{0.343} & 0.422 & 0.457 & 0.246 & 0.248 & \textbf{0.240} & 0.298 & 0.346 \\
horse15 & 0.352 & \textbf{0.331} & 0.336 & 0.399 & 0.478 & 0.230 & 0.230 & \textbf{0.212} & 0.273 & 0.281 \\
horse17 & 0.379 & 0.356 & \textbf{0.352} & 0.414 & 0.448 & 0.257 & 0.250 & \textbf{0.239} & 0.297 & 0.342 \\
horse5 & 0.416 & 0.374 & \textbf{0.364} & 0.430 & 0.484 & 0.260 & 0.264 & \textbf{0.250} & 0.314 & 0.370 \\
horse6 & 0.353 & 0.350 & \textbf{0.340} & 0.411 & 0.424 & 0.233 & 0.237 & \textbf{0.222} & 0.283 & 0.309 \\
horse7 & 0.351 & 0.333 & \textbf{0.327} & 0.398 & 0.435 & 0.238 & 0.233 & \textbf{0.224} & 0.284 & 0.284 \\
michael0 & 0.309 & 0.276 & \textbf{0.266} & 0.320 & 0.392 & 0.185 & 0.189 & \textbf{0.176} & 0.210 & 0.257 \\
michael1 & 0.376 & 0.340 & \textbf{0.333} & 0.396 & 0.480 & 0.246 & 0.226 & \textbf{0.214} & 0.256 & 0.322 \\
michael10 & 0.417 & 0.373 & \textbf{0.346} & 0.418 & 0.618 & 0.267 & 0.256 & \textbf{0.240} & 0.290 & 0.384 \\
michael11 & 0.665 & 0.576 & \textbf{0.547} & 0.681 & 0.879 & 0.443 & \textbf{0.409} & 0.412 & 0.475 & 0.904 \\
michael12 & 0.436 & 0.365 & \textbf{0.344} & 0.431 & 0.507 & 0.268 & 0.262 & \textbf{0.255} & 0.297 & 0.394 \\
michael13 & 0.438 & 0.371 & \textbf{0.364} & 0.422 & 0.698 & 0.283 & 0.267 & \textbf{0.255} & 0.296 & 0.382 \\
michael15 & 0.440 & 0.387 & \textbf{0.369} & 0.449 & 0.639 & 0.283 & 0.263 & \textbf{0.258} & 0.294 & 0.427 \\
michael16 & 0.403 & 0.366 & \textbf{0.349} & 0.416 & 0.552 & 0.263 & 0.259 & \textbf{0.249} & 0.296 & 0.459 \\
michael18 & 0.431 & 0.371 & \textbf{0.346} & 0.422 & 0.594 & 0.266 & 0.255 & \textbf{0.245} & 0.285 & 0.449 \\
michael19 & 0.417 & 0.381 & \textbf{0.367} & 0.433 & 0.517 & 0.259 & 0.254 & \textbf{0.243} & 0.294 & 0.363 \\
michael2 & 0.578 & 0.476 & \textbf{0.470} & 0.537 & 0.647 & 0.358 & 0.325 & \textbf{0.318} & 0.365 & 0.527 \\
michael3 & 0.426 & 0.356 & \textbf{0.346} & 0.413 & 0.564 & 0.262 & 0.249 & \textbf{0.233} & 0.279 & 0.278 \\
michael4 & 0.414 & 0.364 & \textbf{0.350} & 0.415 & 0.600 & 0.264 & 0.249 & \textbf{0.239} & 0.282 & 0.386 \\
michael5 & 0.366 & 0.326 & \textbf{0.313} & 0.384 & 0.454 & 0.222 & 0.220 & \textbf{0.206} & 0.251 & 0.342 \\
michael6 & 0.384 & 0.363 & \textbf{0.349} & 0.411 & 0.453 & 0.248 & 0.234 & \textbf{0.227} & 0.265 & 0.325 \\
michael7 & 0.474 & 0.394 & \textbf{0.365} & 0.450 & 0.645 & 0.287 & 0.273 & \textbf{0.258} & 0.311 & 0.387 \\
victoria1 & 0.321 & \textbf{0.267} & 0.281 & 0.330 & 0.365 & 0.194 & 0.176 & \textbf{0.166} & 0.205 & 0.298 \\
victoria10 & 0.278 & 0.275 & \textbf{0.252} & 0.314 & 0.567 & 0.165 & 0.166 & \textbf{0.156} & 0.202 & 0.239 \\
victoria17 & 0.370 & \textbf{0.305} & 0.309 & 0.357 & 0.436 & 0.203 & 0.190 & \textbf{0.174} & 0.222 & 0.310 \\
victoria2 & 0.302 & 0.267 & \textbf{0.243} & 0.322 & 0.363 & 0.164 & 0.157 & \textbf{0.150} & 0.189 & 0.240 \\
victoria21 & 0.537 & \textbf{0.449} & 0.457 & 0.532 & 1.168 & 0.322 & 0.293 & \textbf{0.281} & 0.337 & 0.449 \\
victoria24 & 0.311 & 0.293 & \textbf{0.277} & 0.354 & 0.416 & 0.188 & 0.173 & \textbf{0.169} & 0.205 & 0.329 \\
victoria25 & 0.283 & 0.258 & \textbf{0.233} & 0.312 & 0.377 & 0.152 & 0.144 & \textbf{0.137} & 0.174 & 0.236 \\
victoria4 & 0.296 & 0.280 & \textbf{0.267} & 0.317 & 0.390 & 0.174 & 0.167 & \textbf{0.160} & 0.197 & 0.243 \\
wolf0 & 0.429 & 0.427 & \textbf{0.414} & 0.477 & 0.510 & 0.331 & 0.334 & \textbf{0.320} & 0.355 & 0.431 \\
wolf1 & 0.410 & 0.419 & \textbf{0.389} & 0.455 & 0.517 & 0.298 & 0.315 & \textbf{0.293} & 0.335 & 0.367 \\
wolf2 & 0.443 & 0.463 & \textbf{0.427} & 0.501 & 0.570 & 0.330 & 0.345 & \textbf{0.327} & 0.363 & 0.471 \\
\hline
\end{tabular}
\label{tab:distance}
\end{table*}

\begin{table*}
\vspace{-3cm}
\setlength\extrarowheight{-2pt}
\centering
\caption{Execution time (in seconds) to fit a subdivision surface with 1000 and 1700 control vertices.  Columns 1--3 and 6--8 show how each additional term in the energy model slows the optimization, while columns 4--5 and 9--10 compare our technique to state-of-the-art model of \cite{Marinov2005} and to the Levenberg-Marquardt (LM) optimization algorithm advocated by \cite{Taylor2016a,Cashman2013,Jaimez2017}. }
\footnotesize
\begin{tabular}{| l | c c c c c | c c c c c |}
\hline
& \multicolumn{5}{|c|}{1000 control vertices} & \multicolumn{5}{|c|}{1700 control vertices } \\
\hline
 mesh       & {$q=2$} & {$\alpha=0$} & {full model}  &  {\cite{Marinov2005}} & {LM solver} & {$q=2$} & {$\alpha=0$} & {full model}  &  {\cite{Marinov2005}}  & {LM solver}
   \\
\hline
cat0 & 32.7 & 34.7 & 56.0 & \textbf{20.8} & 408.0 & 24.6 & 44.5 & 36.0 & \textbf{18.6} & 346.5 \\
cat1 & 37.1 & 41.2 & 44.0 & \textbf{12.9} & 345.2 & 34.5 & 43.3 & 52.6 & \textbf{15.1} & 367.5 \\
cat10 & 24.7 & 36.5 & 34.9 & \textbf{18.2} & 403.3 & \textbf{30.3} & 50.0 & 38.7 & 37.4 & 544.6 \\
cat2 & 23.5 & 33.9 & 43.6 & \textbf{21.3} & 450.0 & 31.3 & 44.3 & 42.3 & \textbf{21.0} & 399.9 \\
cat3 & 28.1 & 36.6 & 47.4 & \textbf{16.4} & 333.7 & 41.5 & 51.8 & 32.4 & \textbf{18.6} & 382.5 \\
cat4 & 30.6 & 43.4 & 46.7 & \textbf{16.5} & 344.1 & 28.6 & 34.8 & 54.7 & \textbf{20.6} & 384.0 \\
cat5 & 18.7 & 38.8 & 41.6 & \textbf{17.4} & 547.1 & 29.6 & 47.3 & 33.4 & \textbf{17.1} & 372.9 \\
cat6 & 24.9 & 43.9 & 40.7 & \textbf{18.0} & 365.1 & 28.5 & 48.3 & 41.1 & \textbf{16.1} & 332.3 \\
cat7 & 30.1 & 23.9 & 40.9 & \textbf{14.3} & 402.6 & 26.6 & 50.9 & 51.4 & \textbf{20.3} & 468.0 \\
cat8 & 34.7 & 42.1 & 35.9 & \textbf{17.4} & 267.5 & 27.4 & 48.7 & 44.1 & \textbf{20.8} & 378.9 \\
cat9 & 22.6 & 35.7 & 56.6 & \textbf{19.1} & 361.4 & 35.5 & 29.7 & 47.2 & \textbf{17.0} & 468.4 \\
centaur0 & 20.7 & 16.5 & 21.5 & \textbf{8.2} & 165.4 & 16.2 & 22.1 & 24.0 & \textbf{15.8} & 185.1 \\
centaur1 & 12.0 & 21.9 & 16.5 & \textbf{8.9} & 279.3 & 23.8 & 22.0 & 33.7 & \textbf{11.7} & 374.1 \\
centaur2 & 16.0 & 15.9 & 26.0 & \textbf{9.9} & 209.2 & 19.2 & 24.9 & 24.1 & \textbf{10.4} & 171.3 \\
centaur3 & 18.3 & 14.1 & 25.9 & \textbf{8.5} & 192.4 & 18.8 & 21.9 & 26.4 & \textbf{10.2} & 258.7 \\
centaur4 & 16.9 & 24.0 & 26.9 & \textbf{8.8} & 267.1 & 24.2 & 28.3 & 27.9 & \textbf{11.3} & 307.5 \\
centaur5 & 14.3 & 17.0 & 28.1 & \textbf{8.8} & 254.2 & 18.9 & 25.0 & 25.9 & \textbf{9.8} & 197.1 \\
david0 & 59.5 & 59.4 & 86.9 & \textbf{39.3} & 602.6 & 64.8 & 56.1 & 63.4 & \textbf{40.2} & 473.5 \\
david1 & 58.8 & 56.4 & 70.1 & \textbf{34.9} & 447.5 & 60.1 & 55.2 & 63.0 & \textbf{36.0} & 799.9 \\
david10 & 44.7 & 47.1 & 72.1 & \textbf{37.6} & 589.8 & 59.0 & 59.8 & 79.9 & \textbf{40.2} & 611.4 \\
david11 & 68.1 & 58.2 & 128.5 & \textbf{37.3} & 683.6 & 59.2 & 76.5 & 100.4 & \textbf{33.1} & 713.5 \\
david12 & 63.4 & 52.0 & 61.7 & \textbf{38.8} & 528.8 & 63.8 & 53.2 & 74.9 & \textbf{45.6} & 622.2 \\
david13 & 61.0 & 45.7 & 86.8 & \textbf{37.4} & 807.0 & 58.8 & 58.4 & 108.8 & \textbf{43.0} & 832.3 \\
david6 & 50.9 & 42.6 & 71.3 & \textbf{30.7} & 592.7 & 54.2 & 60.7 & 63.2 & \textbf{33.7} & 551.9 \\
dog0 & 20.7 & 25.4 & 48.8 & \textbf{14.0} & 325.0 & 23.2 & 31.1 & 39.6 & \textbf{14.6} & 321.8 \\
dog1 & 22.9 & 27.4 & 30.1 & \textbf{15.7} & 281.2 & 39.2 & 39.6 & 41.8 & \textbf{16.8} & 368.4 \\
dog10 & 21.4 & 30.5 & 39.6 & \textbf{12.8} & 283.4 & 31.0 & 31.4 & 32.8 & \textbf{15.0} & 348.5 \\
dog2 & 28.1 & 27.2 & 38.9 & \textbf{14.3} & 251.6 & 29.3 & 45.7 & 48.0 & \textbf{18.8} & 323.8 \\
dog3 & 21.1 & 34.0 & 32.3 & \textbf{16.1} & 303.0 & 28.3 & 27.1 & 34.1 & \textbf{15.4} & 311.2 \\
dog5 & 31.7 & 28.2 & 44.8 & \textbf{20.4} & 271.4 & 34.3 & 48.8 & 34.8 & \textbf{17.5} & 207.0 \\
dog6 & 35.4 & 31.3 & 53.0 & \textbf{18.8} & 293.2 & 24.8 & 45.5 & 49.4 & \textbf{18.5} & 378.7 \\
dog7 & 21.0 & 28.3 & 38.6 & \textbf{15.2} & 427.1 & 44.0 & 47.4 & 45.0 & \textbf{22.2} & 355.7 \\
dog8 & 20.4 & 27.4 & 41.8 & \textbf{14.8} & 304.5 & 28.5 & 45.4 & 37.6 & \textbf{15.7} & 366.5 \\
gorilla1 & 36.8 & 33.0 & 61.2 & \textbf{23.6} & 531.9 & 45.8 & 48.3 & 64.4 & \textbf{34.1} & 416.0 \\
gorilla14 & 32.8 & 41.3 & 52.9 & \textbf{21.0} & 519.9 & 41.6 & 72.1 & 65.6 & \textbf{25.3} & 620.7 \\
gorilla5 & \textbf{37.9} & 47.2 & 55.1 & 38.3 & 565.8 & 28.3 & 55.3 & 63.8 & \textbf{24.3} & 837.3 \\
gorilla8 & 46.5 & 44.5 & 59.6 & \textbf{19.9} & 436.8 & 37.8 & 64.6 & 69.5 & \textbf{26.9} & 695.7 \\
horse0 & 29.6 & 22.5 & 46.4 & \textbf{12.9} & 98.6 & 22.0 & 42.0 & 34.0 & \textbf{13.0} & 234.2 \\
horse10 & 23.1 & 26.7 & 47.0 & \textbf{12.6} & 120.0 & 34.9 & 36.7 & 32.1 & \textbf{11.8} & 302.6 \\
horse14 & 17.5 & 26.2 & 31.8 & \textbf{13.4} & 261.4 & 30.9 & 45.8 & 45.9 & \textbf{13.2} & 406.2 \\
horse15 & 17.5 & 21.6 & 34.1 & \textbf{12.0} & 489.1 & 20.1 & 24.3 & 40.4 & \textbf{16.5} & 252.8 \\
horse17 & 16.9 & 22.6 & 21.4 & \textbf{13.9} & 233.1 & 24.9 & 35.5 & 36.8 & \textbf{13.1} & 316.4 \\
horse5 & 17.2 & 28.9 & 30.5 & \textbf{13.2} & 322.4 & 29.3 & 35.0 & 37.5 & \textbf{16.5} & 346.2 \\
horse6 & 23.2 & 34.8 & 25.4 & \textbf{11.2} & 250.3 & 30.2 & 44.5 & 38.8 & \textbf{13.1} & 254.3 \\
horse7 & 28.0 & 24.0 & 38.6 & \textbf{11.1} & 194.6 & 19.7 & 29.8 & 28.0 & \textbf{12.7} & 290.6 \\
michael0 & 63.1 & 54.9 & 74.1 & \textbf{37.4} & 632.1 & 60.2 & 54.9 & 76.2 & \textbf{40.9} & 599.8 \\
michael1 & 64.5 & 55.0 & 86.9 & \textbf{47.2} & 588.5 & 56.3 & 56.5 & 63.7 & \textbf{44.5} & 909.7 \\
michael10 & 54.4 & 63.1 & 108.5 & \textbf{38.4} & 476.4 & 69.4 & 51.8 & 76.5 & \textbf{29.3} & 517.2 \\
michael11 & 44.1 & 80.9 & 115.8 & \textbf{29.3} & 583.0 & 54.3 & 56.9 & 84.3 & \textbf{31.3} & 762.0 \\
michael12 & 79.7 & 75.2 & 100.4 & \textbf{37.2} & 607.9 & 77.8 & 61.6 & 118.6 & \textbf{40.5} & 573.2 \\
michael13 & 53.7 & 70.3 & 103.5 & \textbf{39.0} & 553.5 & 52.2 & 51.5 & 71.6 & \textbf{39.4} & 546.4 \\
michael15 & 66.4 & 61.5 & 121.4 & \textbf{34.5} & 690.1 & 62.5 & 53.3 & 68.3 & \textbf{38.2} & 443.0 \\
michael16 & \textbf{46.8} & 50.9 & 65.4 & 47.8 & 495.6 & 80.9 & 69.7 & 72.0 & \textbf{56.1} & 571.5 \\
michael18 & 59.0 & 60.1 & 98.9 & \textbf{32.9} & 311.1 & 55.9 & 69.6 & 78.6 & \textbf{34.9} & 472.5 \\
michael19 & 53.6 & 52.5 & 58.6 & \textbf{34.6} & 473.3 & 71.0 & 65.9 & 60.5 & \textbf{38.0} & 535.3 \\
michael2 & 55.3 & 54.4 & 55.5 & \textbf{35.4} & 764.9 & 61.2 & 47.2 & 68.9 & \textbf{35.7} & 476.2 \\
michael3 & 51.8 & 57.4 & 95.9 & \textbf{36.2} & 452.0 & 52.8 & 49.6 & 60.7 & \textbf{38.3} & 658.7 \\
michael4 & 48.6 & 47.5 & 76.3 & \textbf{33.7} & 569.7 & 76.4 & 52.2 & 70.8 & \textbf{36.3} & 576.4 \\
michael5 & 77.6 & 59.3 & 106.4 & \textbf{27.4} & 468.1 & 57.1 & 53.4 & 57.7 & \textbf{38.9} & 1117.8 \\
michael6 & 63.6 & 39.5 & 68.5 & \textbf{26.5} & 389.0 & 68.5 & 67.0 & 65.8 & \textbf{37.6} & 439.6 \\
michael7 & 55.4 & 65.5 & 96.4 & \textbf{35.0} & 508.0 & 62.0 & 57.8 & 62.4 & \textbf{40.5} & 489.5 \\
victoria1 & 49.7 & 49.4 & 52.5 & \textbf{29.3} & 439.4 & 30.3 & 38.9 & 52.5 & \textbf{27.5} & 381.9 \\
victoria10 & 41.8 & 36.8 & 40.3 & \textbf{24.4} & 527.8 & 59.3 & 60.4 & 57.3 & \textbf{30.3} & 405.8 \\
victoria17 & 39.2 & 64.7 & 55.6 & \textbf{27.7} & 582.8 & 45.0 & 43.9 & 72.0 & \textbf{30.6} & 534.7 \\
victoria2 & 53.5 & 38.0 & 45.8 & \textbf{27.6} & 448.4 & 48.0 & 57.0 & 67.9 & \textbf{27.6} & 584.8 \\
victoria21 & 35.1 & 69.0 & 80.2 & \textbf{27.0} & 321.3 & 37.5 & 42.9 & 68.9 & \textbf{24.5} & 444.0 \\
victoria24 & 40.7 & 46.2 & 84.5 & \textbf{22.7} & 366.5 & 27.6 & 47.2 & 76.1 & \textbf{26.8} & 549.3 \\
victoria25 & 49.9 & 52.9 & 57.7 & \textbf{23.6} & 308.4 & 47.3 & 79.1 & 65.5 & \textbf{28.5} & 412.3 \\
victoria4 & 46.5 & 43.7 & 60.4 & \textbf{27.5} & 393.3 & 41.6 & 46.4 & 58.7 & \textbf{23.7} & 478.4 \\
wolf0 & 7.4 & 12.1 & 11.5 & \textbf{6.8} & 64.2 & 10.6 & 16.7 & 9.1 & \textbf{5.7} & 114.0 \\
wolf1 & \textbf{6.5} & 8.5 & 9.1 & 8.2 & 78.5 & 17.0 & 22.8 & \textbf{15.8} & 16.6 & 137.9 \\
wolf2 & 7.0 & 8.0 & 9.7 & \textbf{5.7} & 52.9 & \textbf{8.0} & 11.6 & 16.7 & 8.1 & 106.7 \\
\hline
\end{tabular}
\label{tab:time}
\end{table*}

\subsection*{Experiments on Shape Analsysis}\label{sec:ExpWKS}
Figure~\ref{fig:eigen} compares the wave kernel signature ~\cite{aubry-et-al-11} at a surface point $w_x$ with three different discretizations: a fine triangular mesh, a coarse triangular mesh, and our subdivision surface. This signature is a shape descriptor that assigns to each point $x\in S$ a function $w_x$ that depends on the value of the $\Delta$-eigenfunctions and their eigenvalues at $x$. The point signature of coarse mesh is different from the signature of fine mesh, which is well matched by the subdivision surface.

\begin{figure*}
  \centering
    \begin{tabular}{ccccc}
      \includegraphics[height = 0.17\textwidth]{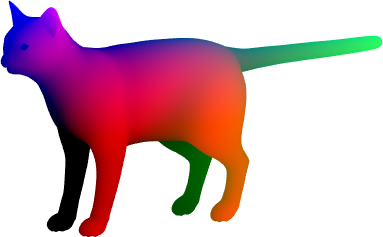}
     & \includegraphics[height = 0.17\textwidth]{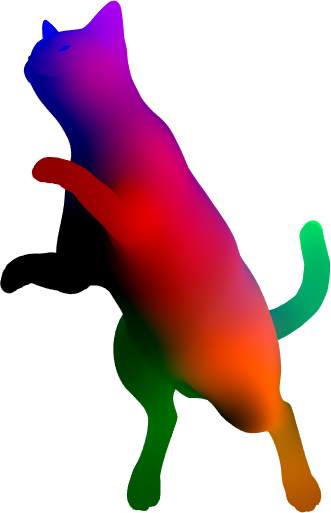}
     & \includegraphics[height = 0.17\textwidth]{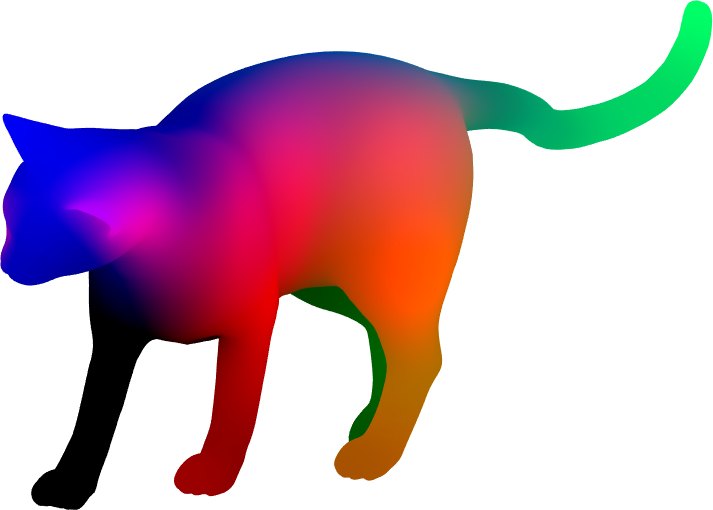}
      &      \includegraphics[height = 0.17\textwidth]{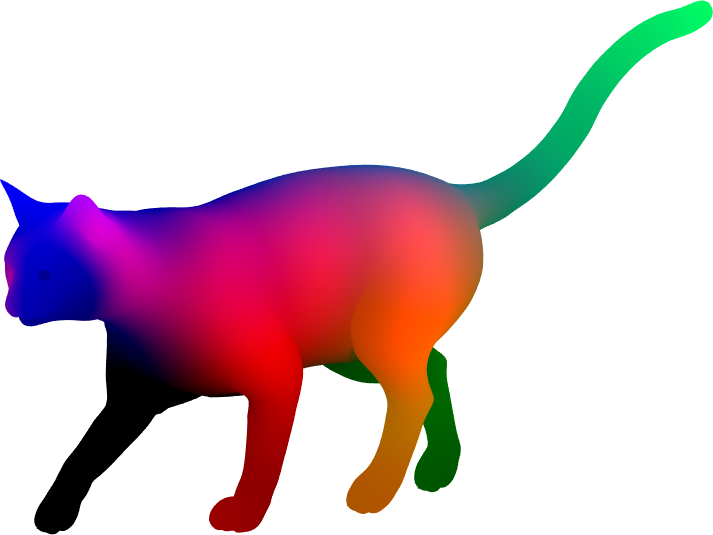} \\
      Fine mesh & Matched shape & Matched shape  & Matched shape \\          
      \\
      \includegraphics[height = 0.17\textwidth]{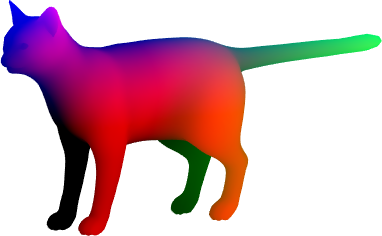}
     & \includegraphics[height = 0.17\textwidth]{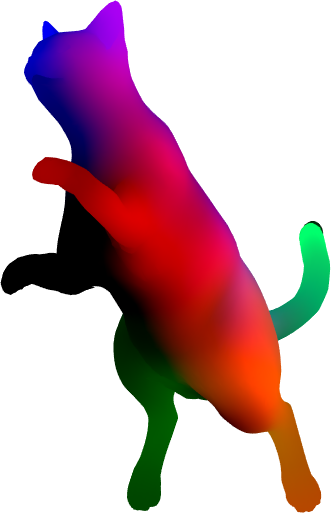}
     & \includegraphics[height = 0.17\textwidth]{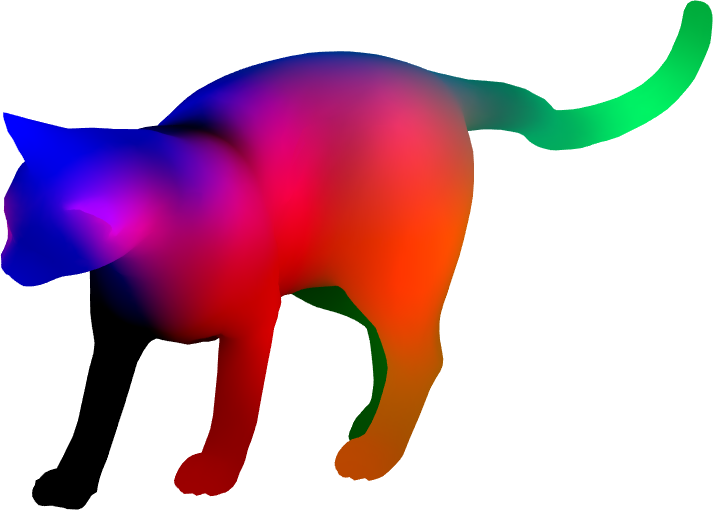}
      &\includegraphics[height = 0.17\textwidth]{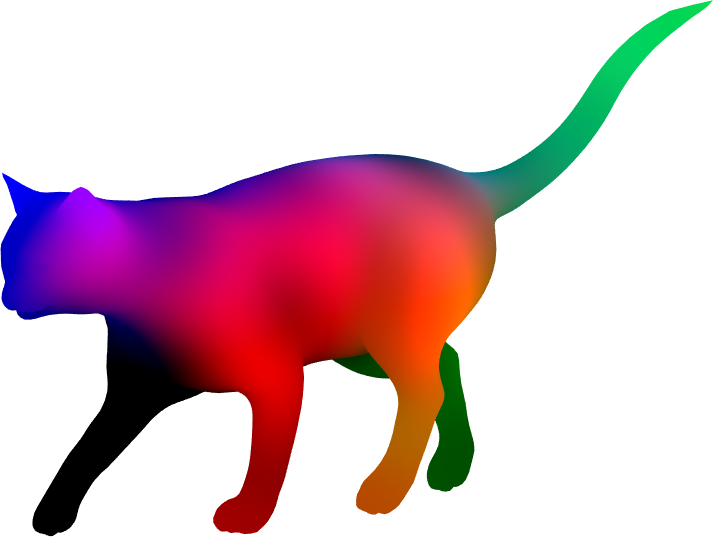}\\           
      Subdivision surface  & Matched shape & Matched shape  & Matched shape \\ 
    \end{tabular}
  \caption{Shape matching results of the functional map
    framework~\cite{Ovsjanikov2012}. After the matching is determined, the color information of the target shape is transferred to the matched shape with the mapping estimated by functional maps\cite{Ovsjanikov2012}.}
  \label{fig:matching}
\end{figure*}

The functional map framework~\cite{Ovsjanikov2012} formulates the problem of matching points of two shapes as matching smooth functions over them. The problem is invariant to isometric transformations by representing smooth function with $\Delta$-eigenfunctions and by finding a linear map between these eigenfunctions. Figure~\ref{fig:matching} shows the computed matching with respect to two different shape representations: a fine mesh with 27894 vertices and a subdivision surface with 1700 control vertices. Our experiments show comparable matching results with a slight loss of accuracy at the tails of some cats, but this is compensated by a surface representation and matching problem of 10\% of the original size. To visualize the results, we show a template shape with a color map that interprets the three spatial coordinates of every point as RGB color information and transfer this color map to each target shape with the matching estimated with functional maps~\cite{Ovsjanikov2012}. In this experiment, using a coarse triangular mesh to do the shape matching simply fails. 

\begin{figure*}
  \centering
    \begin{tabular}{cccc}
      \includegraphics[width=0.20\textwidth]{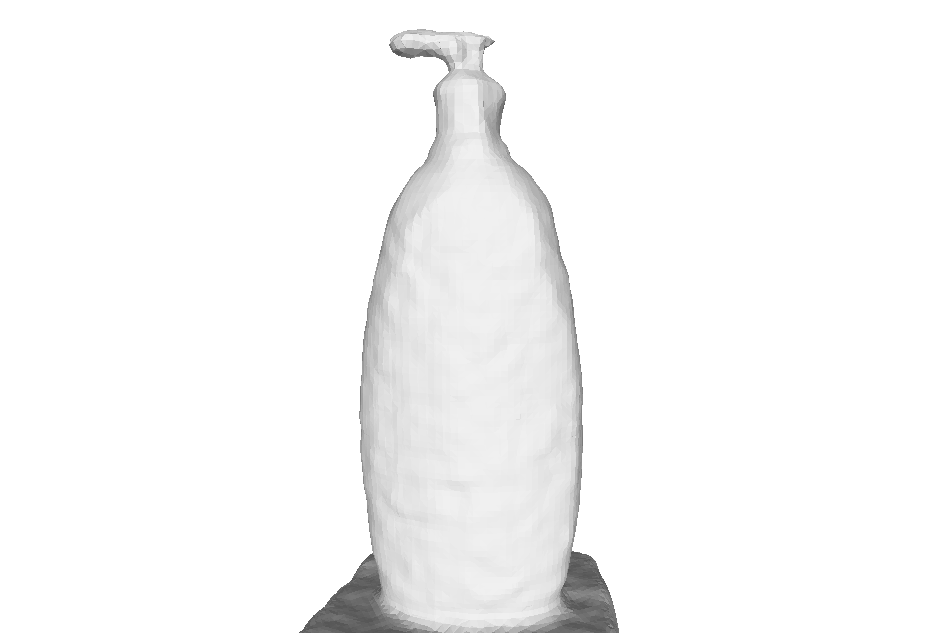}&
      \includegraphics[width=0.20\textwidth]{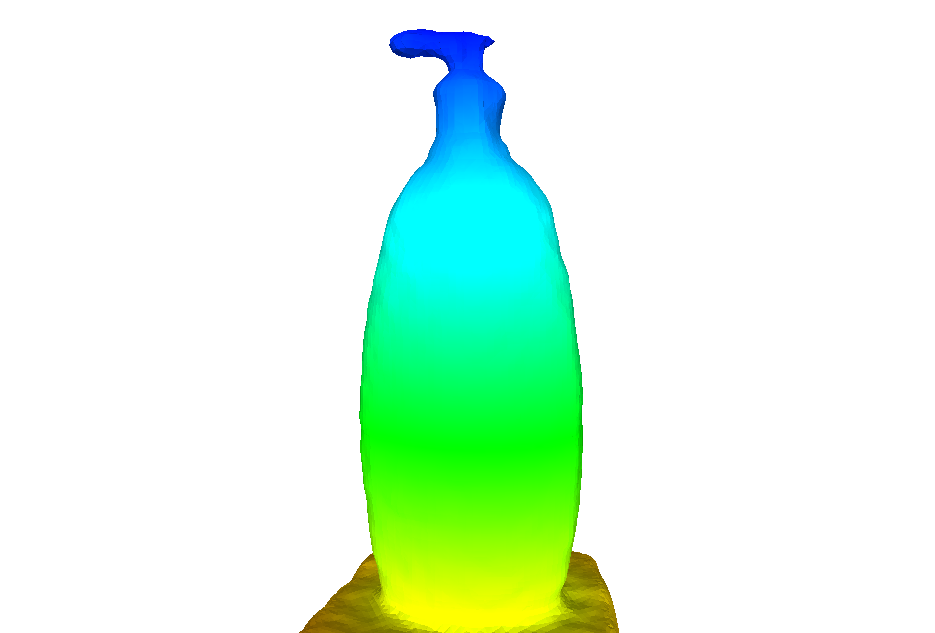} &     
            \includegraphics[width=0.20\textwidth]{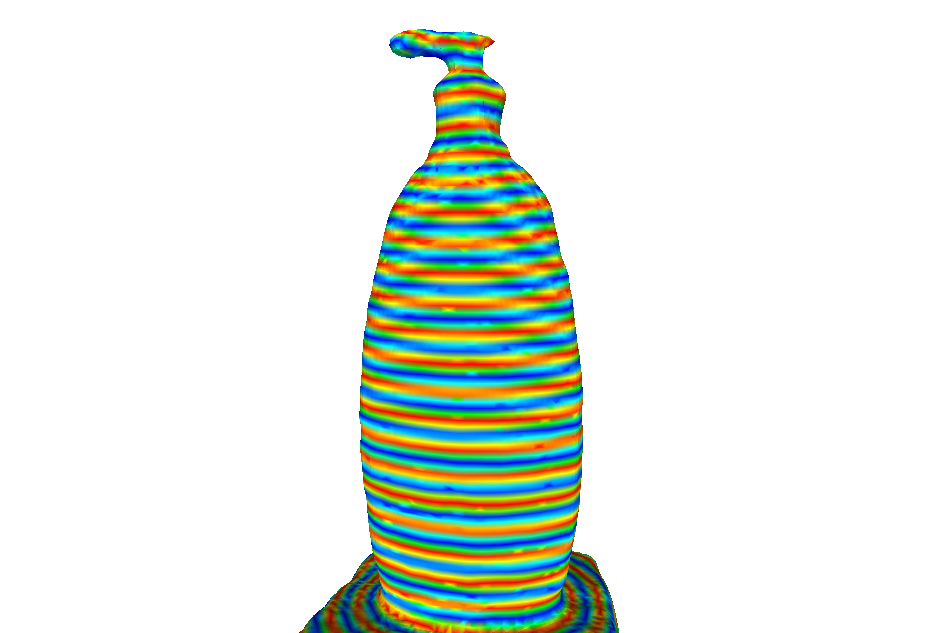}& 
      \includegraphics[width=0.20\textwidth]{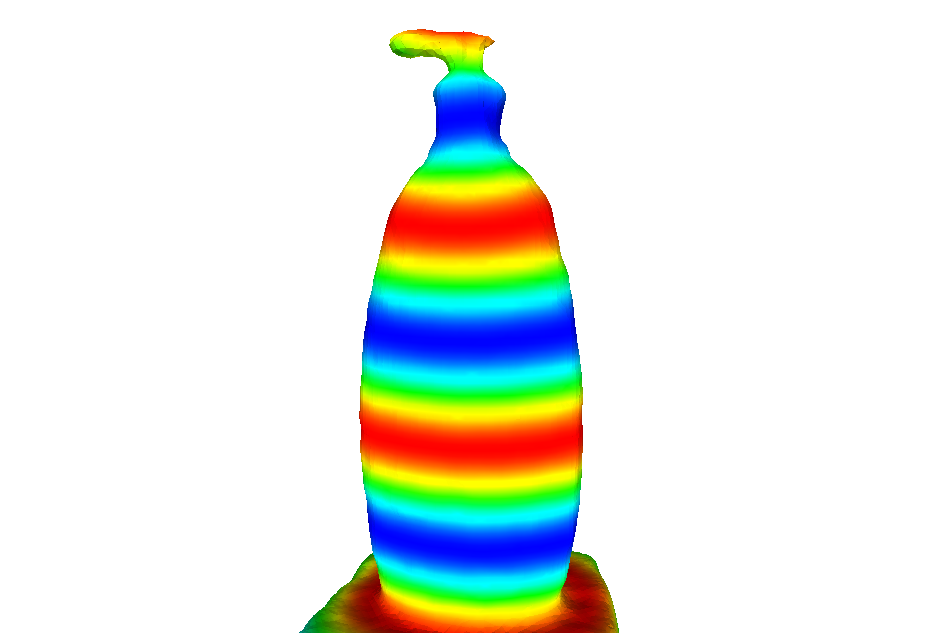} \\
      triangular mesh from \cite{Kazhdan2013} & approximate geodesic  & 30 geodesic level lines & 5 geodesic 5 level lines \\
      12928 vertices & $g$ & $\cos( 30 \pi g )$ & $\cos( 5 \pi g )$ \\
      \includegraphics[width=0.20\textwidth]{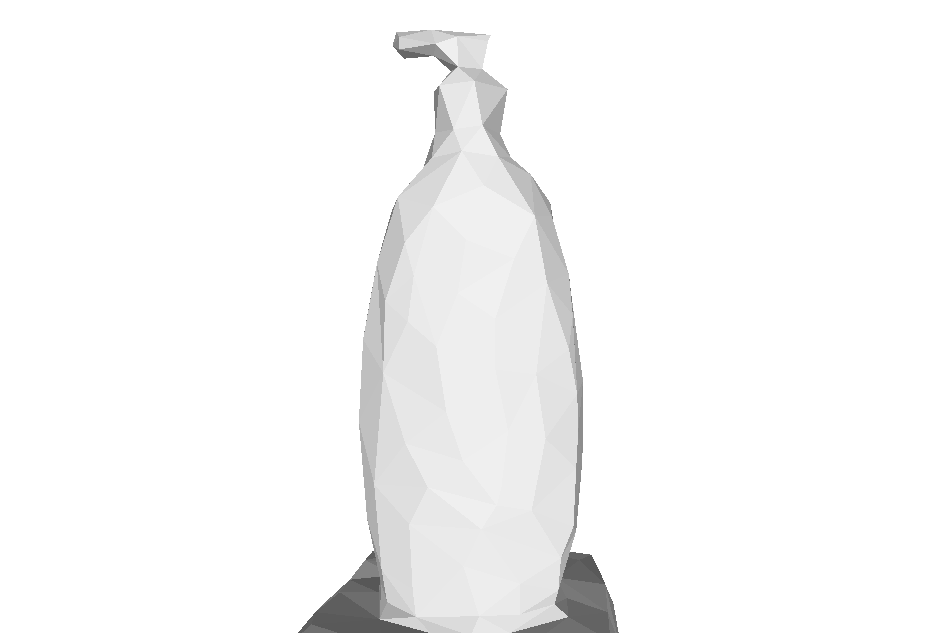}&
      	\includegraphics[width=0.20\textwidth]{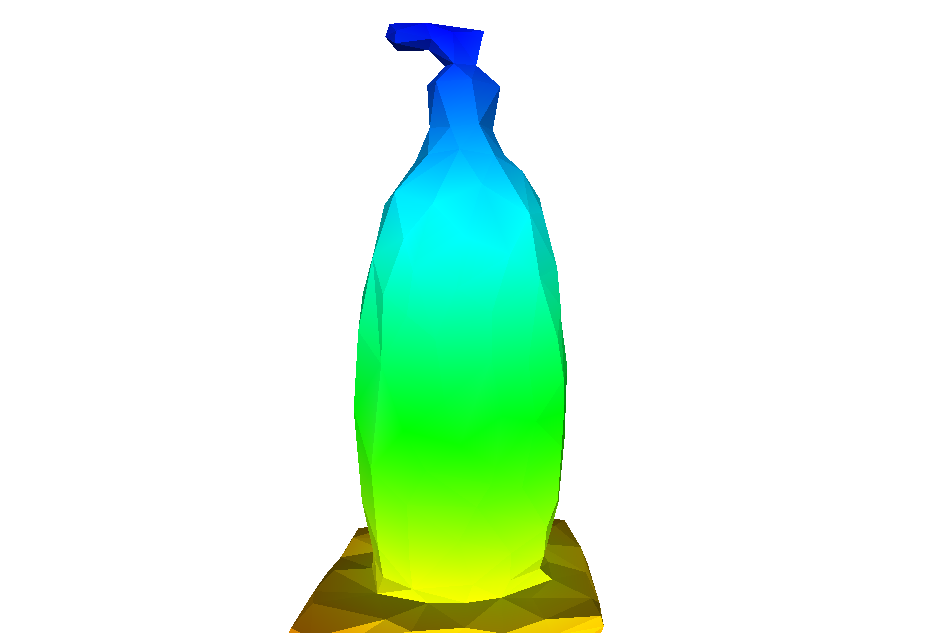}&   
      \includegraphics[width=0.20\textwidth]{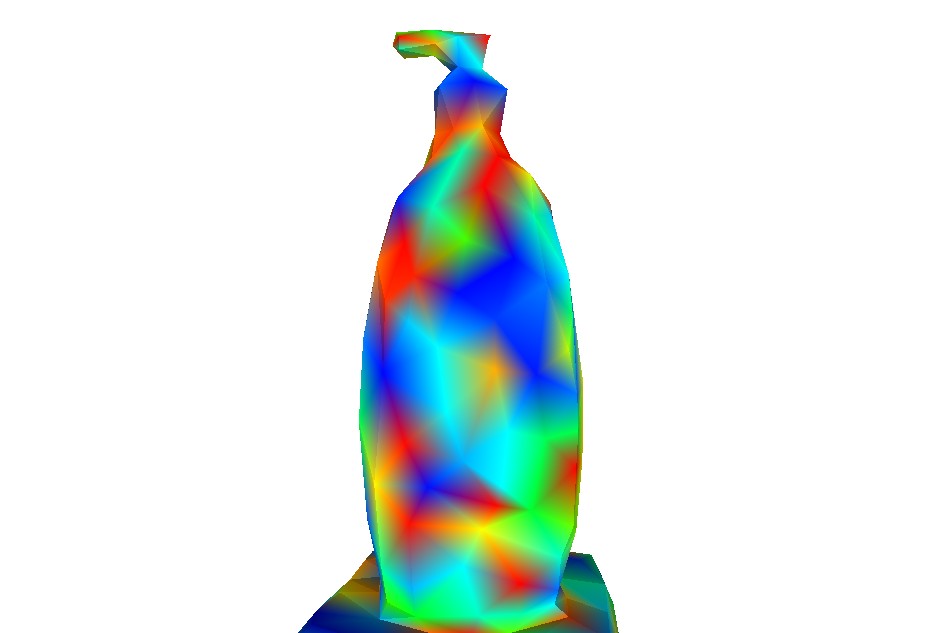} &
      \includegraphics[width=0.20\textwidth]{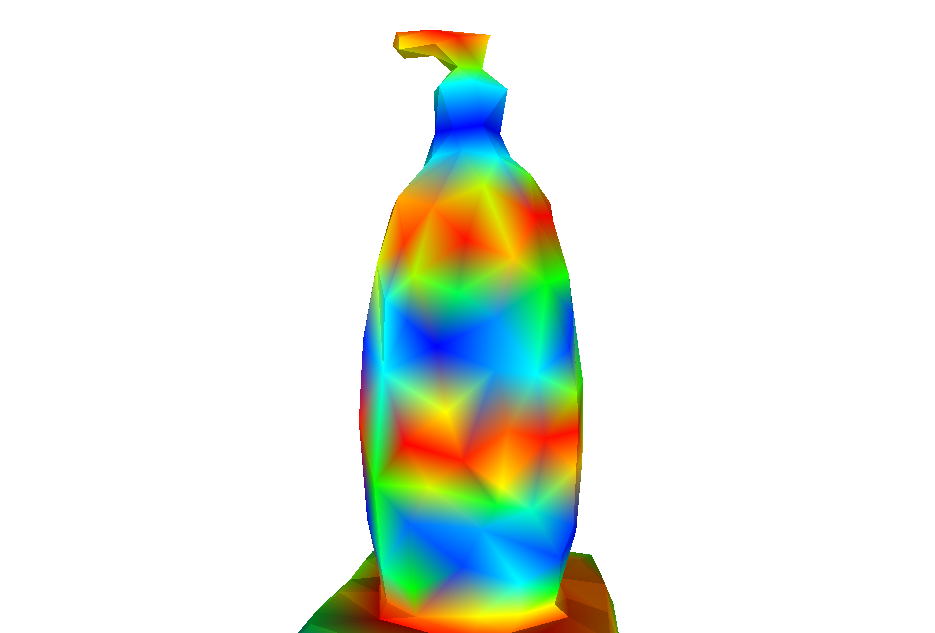} \\
      edge-collapsed mesh \cite{Kazhdan2013}& approximate geodesic  & 30 geodesic level lines & 5 geodesic 5 level lines \\ 
      250 vertices & $g$ & $\cos( 30 \pi g )$ & $\cos( 5 \pi g )$ \\     
      \includegraphics[width=0.20\textwidth]{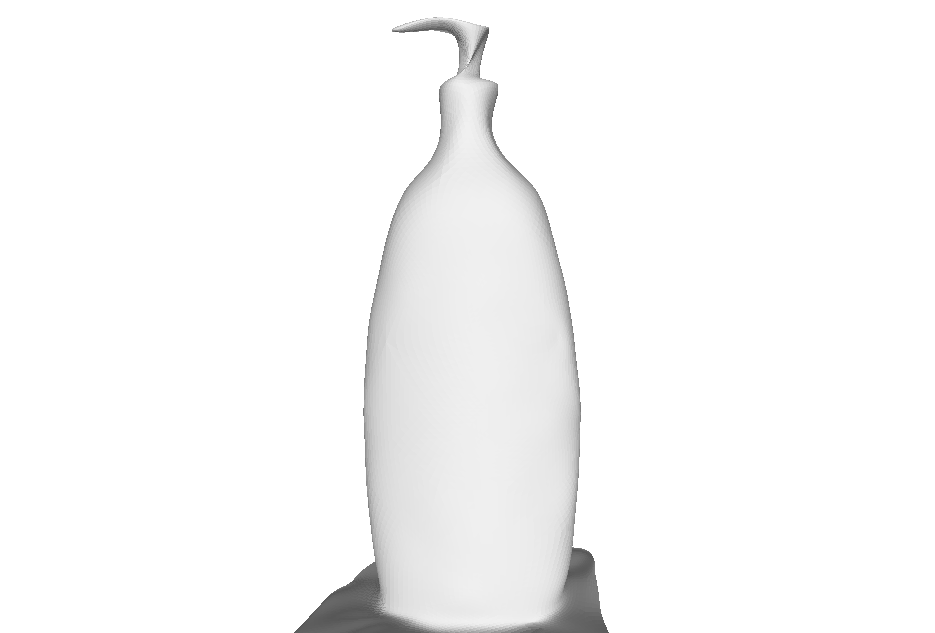} & 
	  \includegraphics[width=0.20\textwidth]{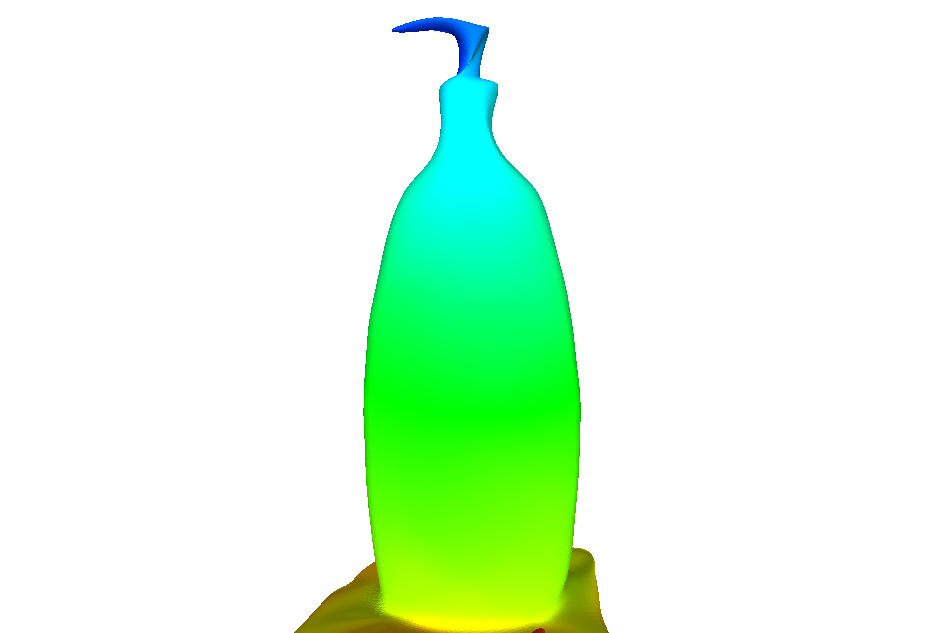}&	 
 	  \includegraphics[width=0.20\textwidth]{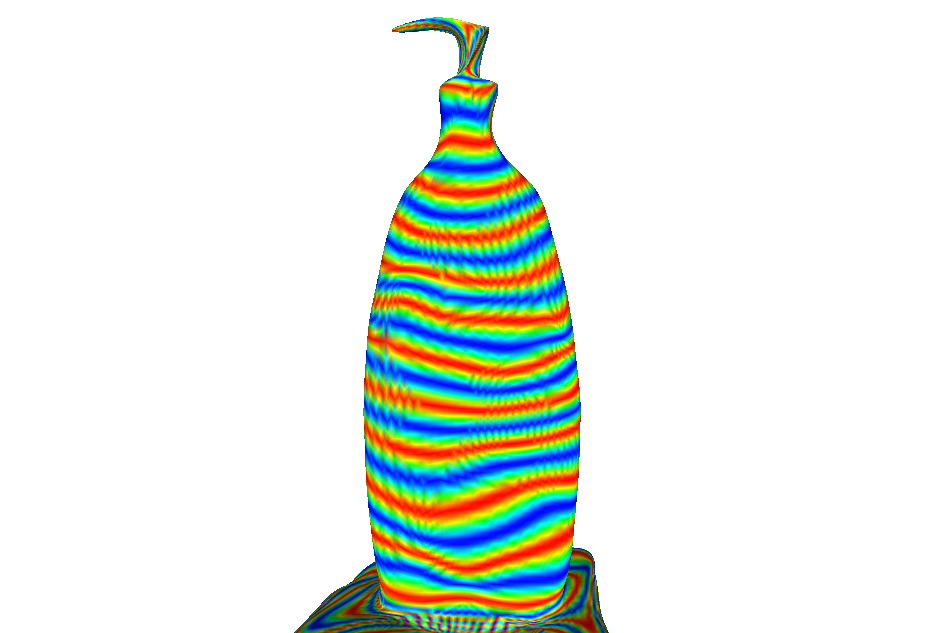} &
 	  \includegraphics[width=0.20\textwidth]{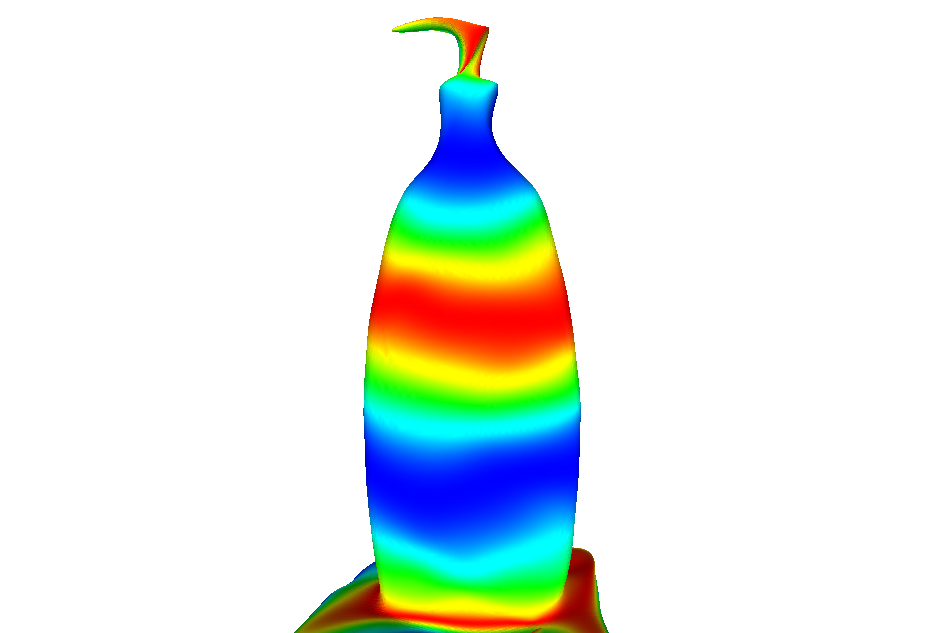}\\ 
      subdivision surface & approximate geodesic  & 30 geodesic level lines & 5 geodesic 5 level lines \\
      250 control vertices & $g$ & $\cos( 30 \pi g )$ & $\cos( 5 \pi g )$ \\
          \end{tabular}     
  \caption{Computation of approximate geodesics by the Heat method \cite{Crane2013} with different surface representations fitted to a Kinect pointcloud. Column 1: representation of the surface with a high-resolution mesh obtained with Poisson reconstruction~\cite{Kazhdan2013} (row 1), a low-resolution triangular mesh obtained by edge collapse of the high-resolution mesh (row 2), and with our subdivision surface (row 3). Column 2: approximate geodesic $g$ estimated with the Heat method \cite{Crane2013} on each surface representation. Columns 3 and 4: level lines of the geodesics visualized as $\cos( \varpi g)$, with $\varpi = 30\pi$ in column 3 and $\varpi = 5\pi$ in column 4. Our subdivision surface is comparable to the surface obtained with the Poisson reconstruction method \cite{Kazhdan2013} and can faithfully represent geodesics at both low and high resolution even though it has 2\% of its vertices. In contrast, compressing the state-of-the-art mesh with the quadratic edge collapse method of \cite{Garland1997} looses all the small scale details of the surface and its geodesics. This is highlighted by representing the level lines of the geodesic on the mesh at high resolution (column 3) where we show how the discretization associated with a coarse triangular mesh representation cannot estimate the high-frequency information of the geodesic accurately.}
  \label{fig:geodesics1}
\end{figure*}

\begin{figure*}
  \centering
    \begin{tabular}{cccc}
      \includegraphics[width = 0.23\textwidth]{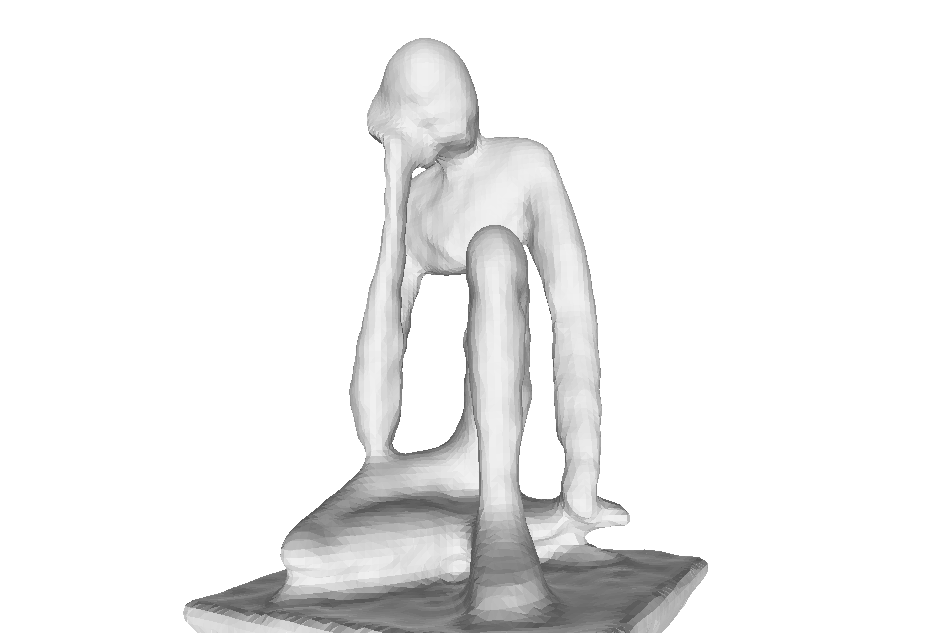}&
      \includegraphics[width = 0.23\textwidth]{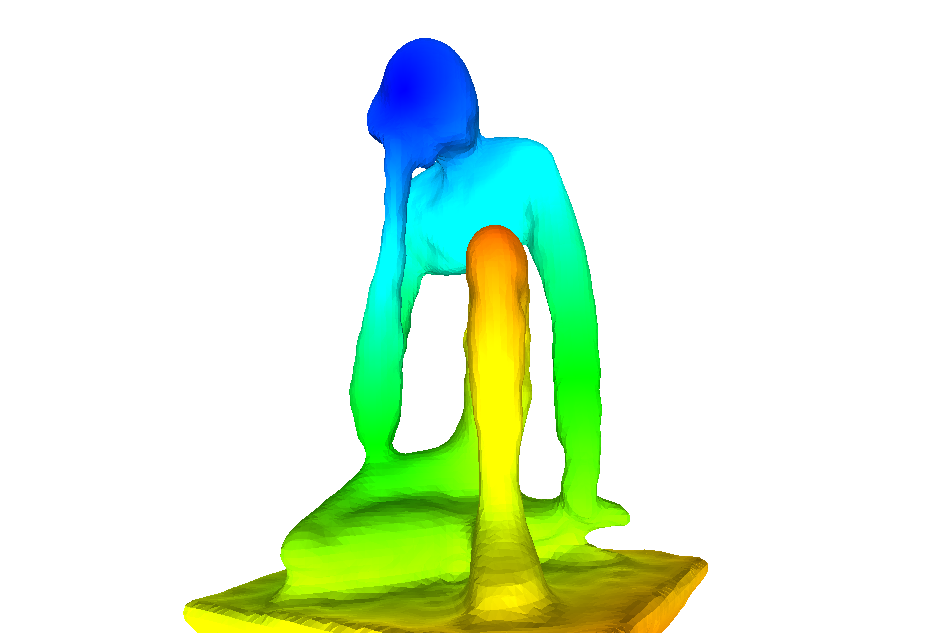} &     
            \includegraphics[width = 0.23\textwidth]{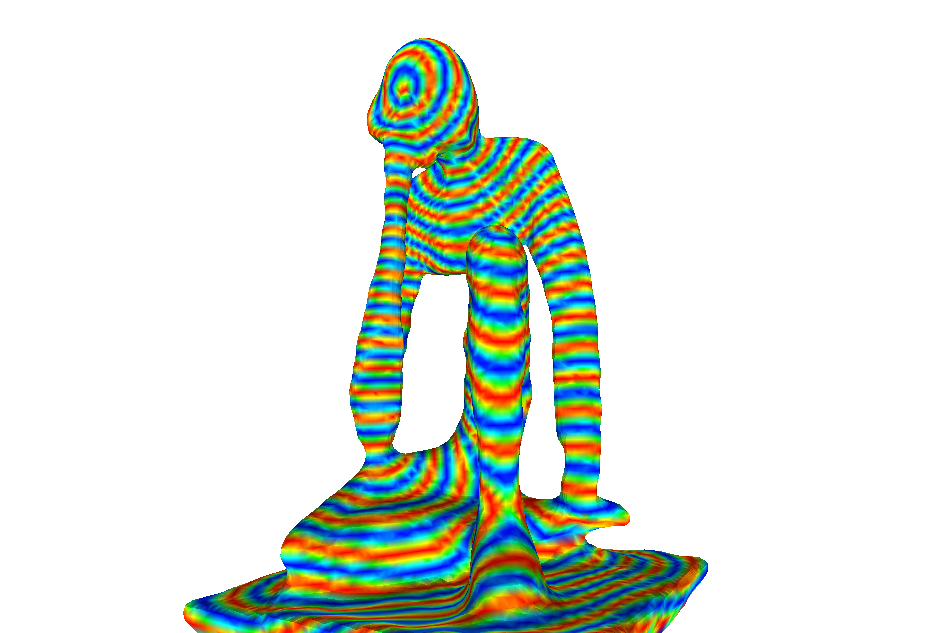}& 
      \includegraphics[width = 0.23\textwidth]{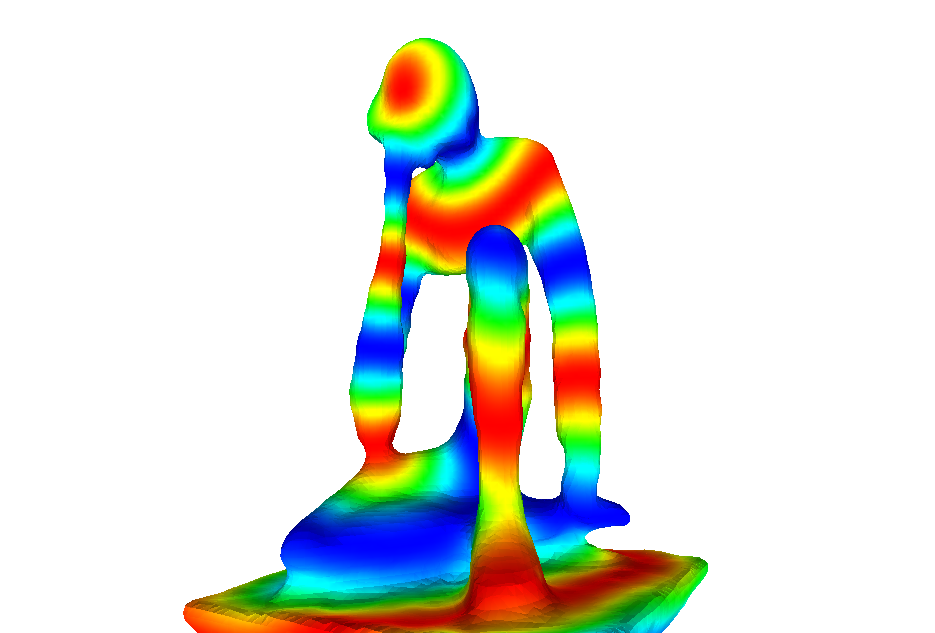} \\
      triangular mesh from  \cite{Kazhdan2013} & approximate geodesic  & 30 geodesic level lines & 5 geodesic 5 level lines \\
       20932 vertices & $g$ & $\cos( 30 \pi g )$ & $\cos( 5 \pi g )$ \\
      \includegraphics[width = 0.23\textwidth]{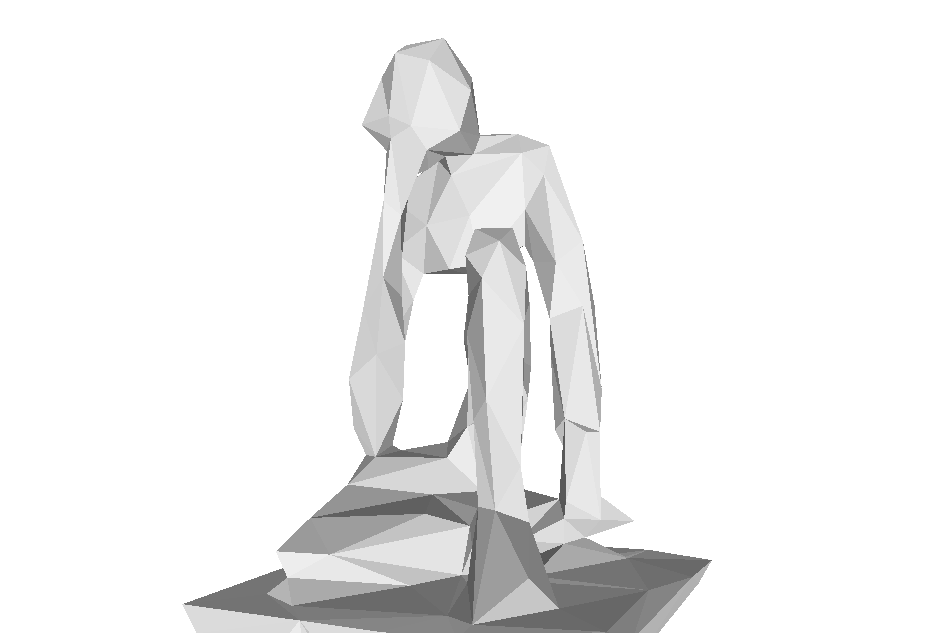}&
      	\includegraphics[width = 0.23\textwidth]{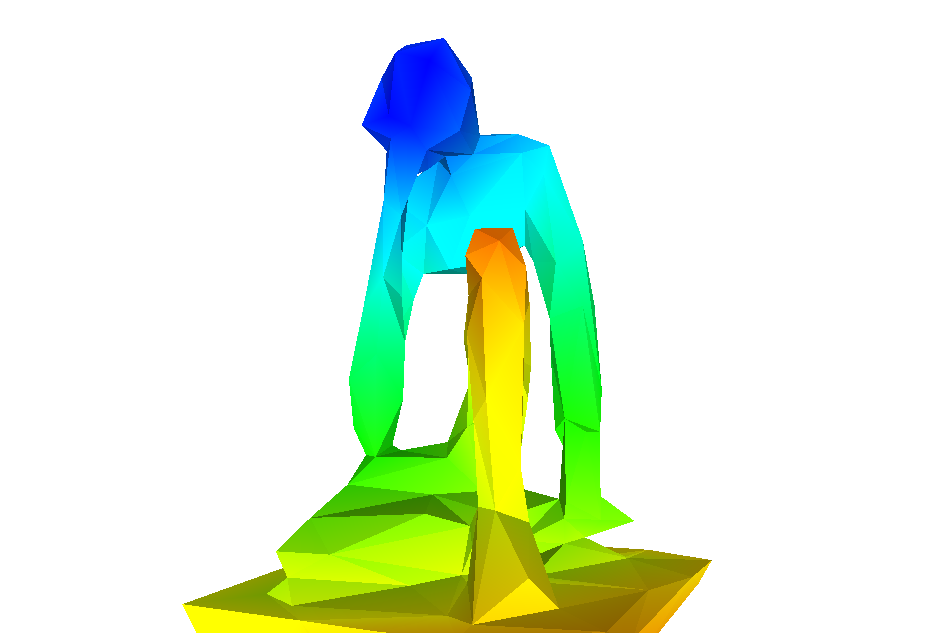}&   
      \includegraphics[width = 0.23\textwidth]{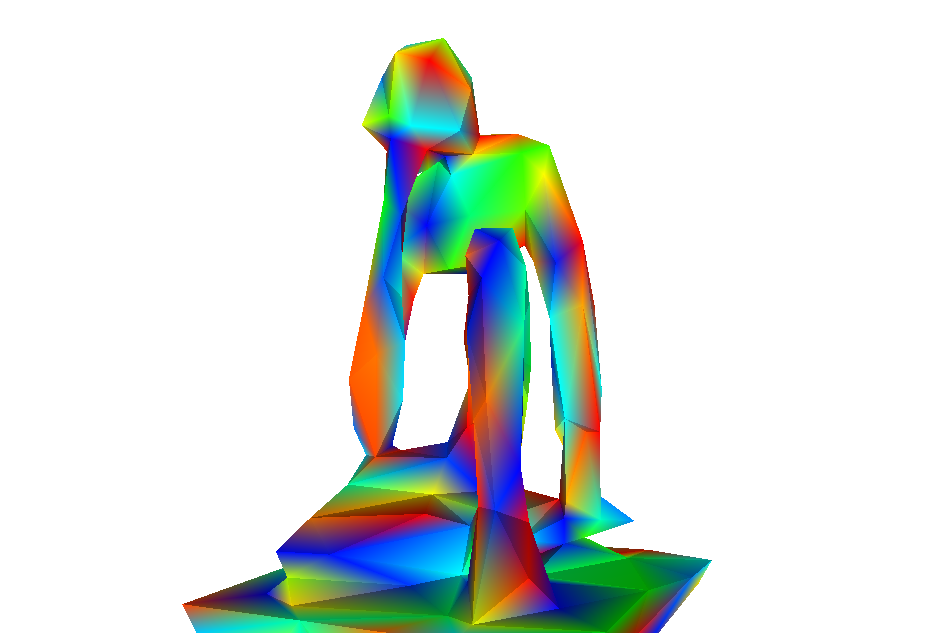} &
      \includegraphics[width = 0.23\textwidth]{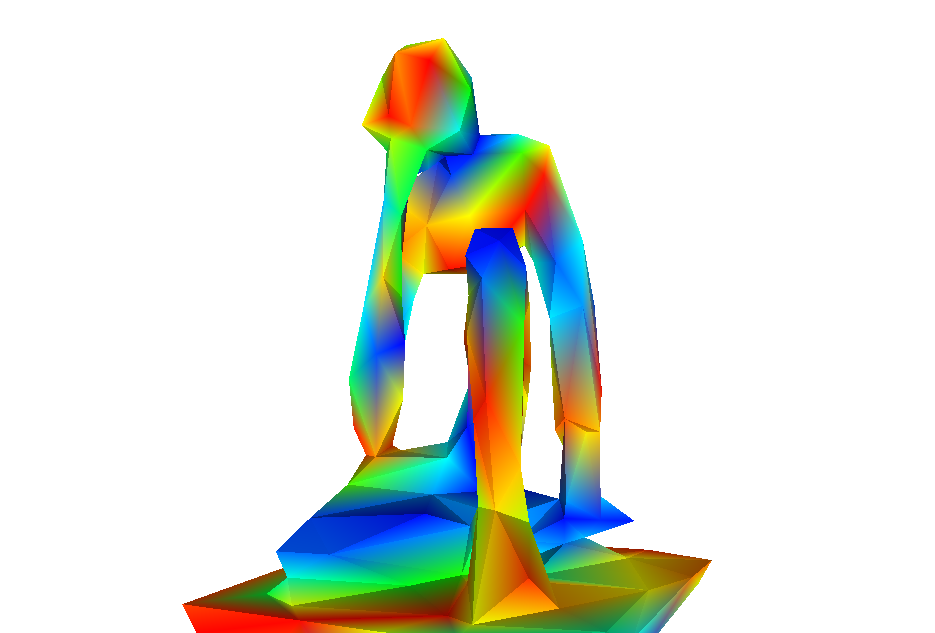} \\
      edge-collapsed mesh \cite{Kazhdan2013} & approximate geodesic  & 30 geodesic level lines & 5 geodesic 5 level lines \\  
      250 vertices & $g$ & $\cos( 30 \pi g )$ & $\cos( 5 \pi g )$ \\    
      \includegraphics[width = 0.23\textwidth]{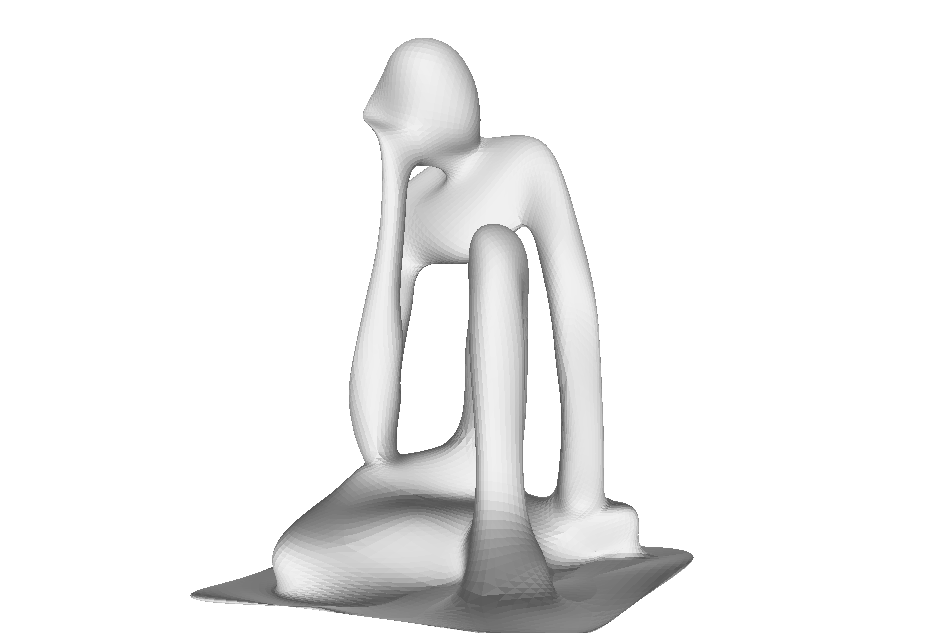} & 
	  \includegraphics[width = 0.23\textwidth]{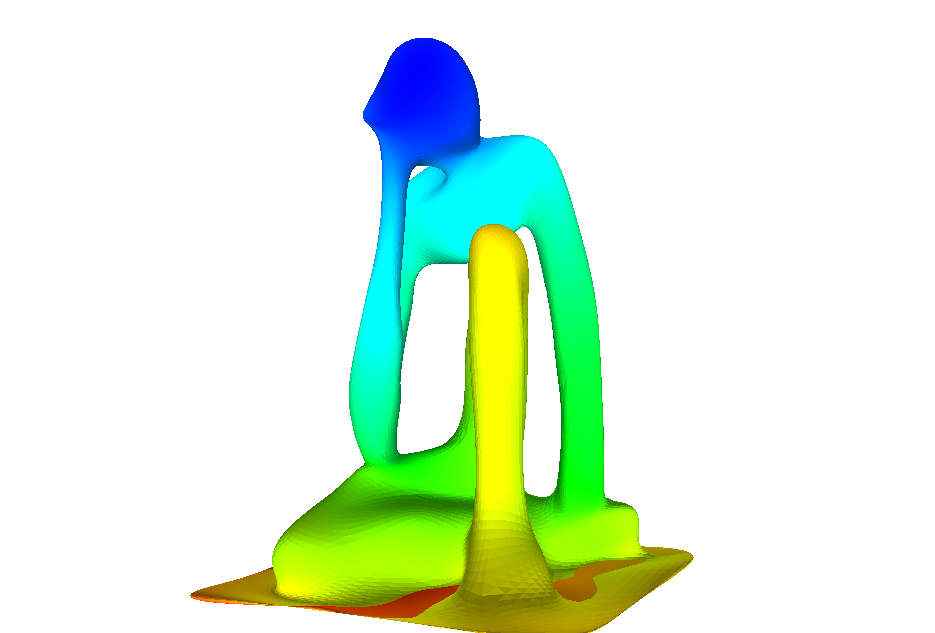}&	 
 	  \includegraphics[width = 0.23\textwidth]{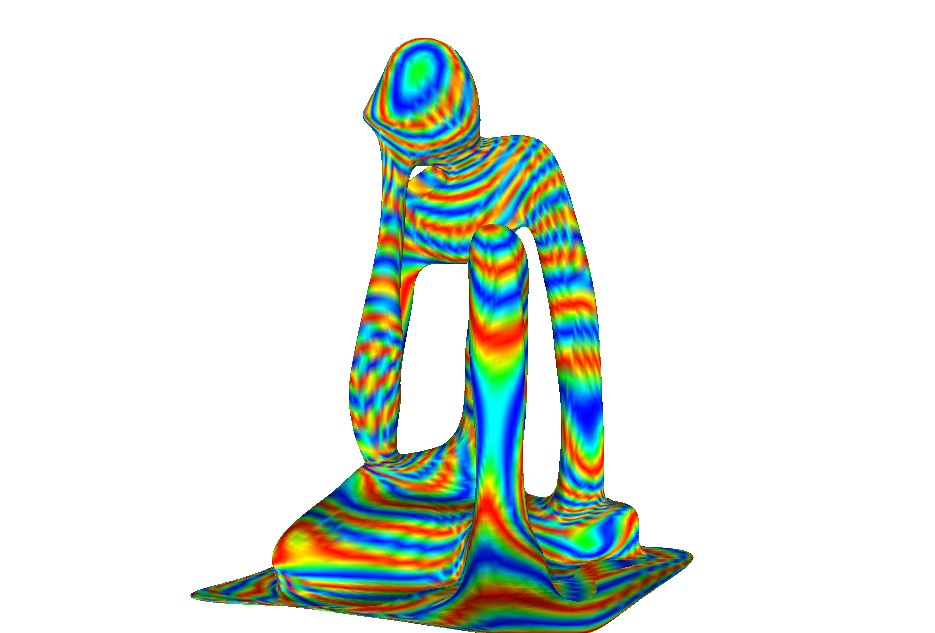} &
 	  \includegraphics[width = 0.23\textwidth]{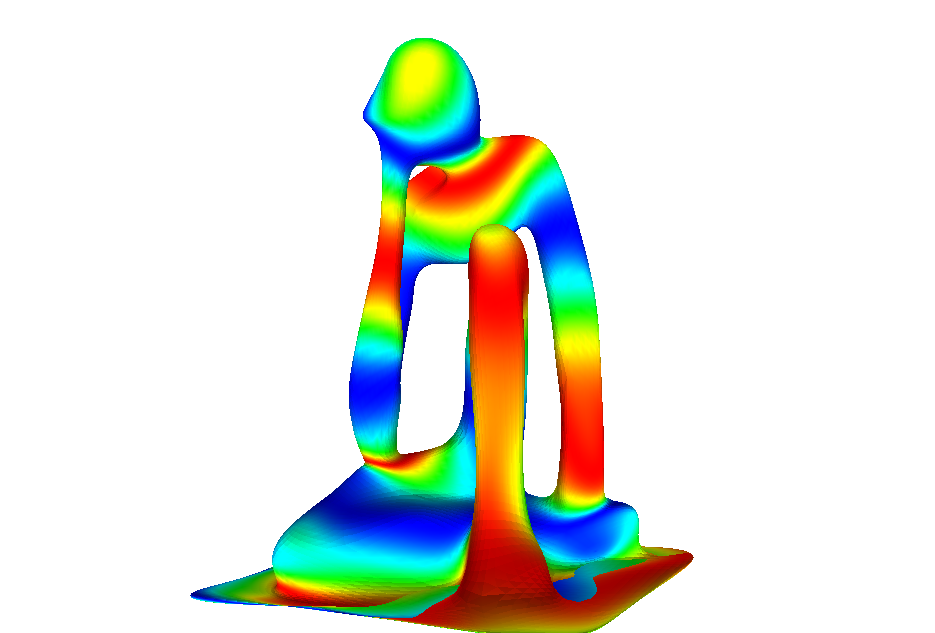}\\ 
      subdivision surface & approximate geodesic  & 30 geodesic level lines & 5 geodesic 5 level lines \\
      250 control vertices & $g$ & $\cos( 30 \pi g )$ & $\cos( 5 \pi g )$ \\
          \end{tabular}     
  \caption{Computation of approximate geodesics by the Heat method \cite{Crane2013} with different surface representations fitted to a Kinect pointcloud. Column 1: representation of the surface with a high-resolution mesh obtained with Poisson reconstruction~\cite{Kazhdan2013} (row 1), a low-resolution triangular mesh obtained by edge collapse of the high-resolution mesh (row 2), and with our subdivision surface (row 3). Column 2: approximate geodesic $g$ estimated with the Heat method \cite{Crane2013} on each surface representation. Columns 3 and 4: level lines of the geodesics visualized as $\cos( \varpi g)$, with $\varpi = 30\pi$ in column 3 and $\varpi = 5\pi$ in column 4. Our subdivision surface is comparable to the surface obtained with the Poisson reconstruction method \cite{Kazhdan2013} and can faithfully represent geodesics at both low and high resolution even though it has 2\% of its vertices. In contrast, compressing the state-of-the-art mesh with the quadratic edge collapse method of \cite{Garland1997} looses all the small scale details of the surface and its geodesics. This is highlighted by representing the level lines of the geodesic on the mesh at high resolution (column 3) where we show how the discretization associated with a coarse triangular mesh representation cannot estimate the high-frequency information of the geodesic accurately.}
  \label{fig:geodesics2}
\end{figure*}

Figures \ref{fig:geodesics0}--\ref{fig:geodesics2} compare geodesics computed by the Heat method \cite{Crane2013} with different surface representations fitted to the Kinect pointclouds of Figures \ref{fig:geodesics0} and \ref{fig:thinker}. As a proxy for ground-truth, we use a high-resolution triangular mesh obtained by the Poisson reconstruction method~\cite{Kazhdan2013} and compare it to two compressed surface representations of the same size: a low-resolution triangular mesh obtained by quadratic edge collapse~\cite{Garland1997} of the high-resolution mesh and our subdivision representation. Our subdivision surface is comparable to the high-resolution mesh and can be used to compressed both the shape and its geodesics functions. In contrast, the low-resolution triangular mesh provides a discretization too coarse to represent the small scale details of the surface and its geodesic. To highlight the loss of small-scale structure on the geodesic, we represent the level lines of the geodesic at high and low resolution and demonstrate how the geodesic estimated from the coarse triangular mesh is unable to capture the high-frequency information of the geodesic (as estimated from the high-resolution mesh) while our surface representation captures both low and high frequency information accurately.

\section{Conclusions}\label{sec:conclusion}

We have presented a method to compress surfaces with a representation
well suited for shape analysis. To this purpose, we fit a subdivision surface to low-level surface representations and develop standard shape analysis operators with it.

Our surface fitting model is robust to noise and outliers and can handle both clean input meshes and noisy point cloud obtained from real data. The resulting subdivision surfaces let us represent smooth shapes at high accuracy with a fraction of the variables used in triangular meshes and use the same subdivision schemes to parametrize functions over the surface and compute tangent spaces or curvatures analytically. Our experiments show how the smoothness of the representation reduces the size of the Laplace-Beltrami operator and its eigenfunctions without loss of accuracy and allows us to compute wave-kernel signatures, approximate geodesics, and shape matches in a compact representation. 

\appendix
\section{Appendix: Quadratic Solver}\label{sec:optimization}
 We optimize the objective energy 
\begin{align}\label{eq:optimization}
  \min_{S} \ E(S) =& \underbrace{\dist(S,\Pointcloud)}_\text{Point Fit} + %
  \underbrace{\alpha \NormalPenalty(S, \mathbf{T})}_\text{Tangent
    Fit}+%
  \underbrace{\beta R(S)}_\text{Regularization},
\end{align}
 that fits a subdivision surface $S$ to a set of point $\Pointcloud$ and normal data $\mathbf{T}$ by solving a sequence of convex problems
  \begin{align}
    V^{m+1} \leftarrow \min_V \, V^\top\left(Q^m V - b^m\right)
  \end{align} %
over the vertices $V = [v_1,\ldots,v_n]$ of the control mesh $\Mesh$. In this section, we show how to obtain the quadratic objective function $V^\top\left(Q^m V - b^m\right)$ that approximates the original energy at iteration $m$.

Following a majorize-minimize (MM) principle, at iteration $m$ we substitute our robust energy by the weighted least-squares problem
  \begin{align}
    \min_{U,V} & \ \sum_{j=1}^{N} w^m_j \norm{p_j - \Phi(u_j)}^2 + %
    \sum_{i=1}^2 \alpha^{m}_{ij}\vecprod{t_j}{\partial_i\Phi(u_j)}^2 \nonumber \\
   &+ \beta \norm{R V}^2 \label{eq:RWLS}
  \end{align} %
  over the mesh vertices $V$ and the correspondence parameters $u_1,\ldots,u_N \in \cT_\Mesh$. The parametrization of the subdivision surface 
  \begin{align*}\Phi(u)=\sum_{i=1}^n v_i \Phi_i(u),
  \end{align*}
  depends linearly on the mesh vertices and can be described compactly with an $N \times n$ sparse matrix $L$ with $L_{ji}=\Phi_i(u_j)$ that samples the surface as
  \begin{align*}
   x_j=\sum_{i=1}^n v_i \Phi_i(u_j)~~j=1,\ldots,N && X = L\cdot V.
  \end{align*}   
Analogously, we define $L_1$ and $L_2$ for the partial derivatives of the basis functions%
  \begin{align}
    \partial_i\Phi(u_j) =& \sum_{k=1}^n L_{i,jk}v_k
  \end{align}
and introduce the auxiliary variables $x'_{ij}=\partial_i\Phi(u_j)$ that sample the tangent plane of the surface by
  \begin{align*}
    X'_i =& L_i\cdot V \in\IR^{N\times 3}.
  \end{align*} %

We remove the variables $u_j$ from the optimization by approximating the squared distance with the quadratic expression in $V$
  \begin{align*}
  \small
    \sum_{j=1}^{n} w^m_j \norm{p_j - \Phi(u_j)}^2 &\approx 
    \sum_{j=1}^{n} w^m_j \norm{p_j - (L\cdot V)_j}_{D_j^m}^2,\\
    &\approx \sum_{j=1}^{n} w^m_j (p_j - x_j)^{\top} D^m_j (p_j - x_j)
  \end{align*}
where the symmetric positive definite matrix 
\begin{align}
D_j^m = \frac{d}{d+\rho_1} \tau_1\tau_1^\top + \frac{d}{d+\rho_2} \tau_2\tau_2^\top + \nu\nu^\top
\end{align}
is determined by the approximation to the squared distance to the surface $S^m$ at $p_j$. In particular, $x_j$ is a surface point close to $p_j$ (but not necessarily its projection onto $S^m$), $d$ is the distance from $p_j$ to this point, and $\rho_1, \rho_2, \tau_1, \tau_2,  \nu$ are the absolute values of the principal curvature radii, the principal curvature directions, and the normal to the surface $S^m$ at $x_j$. Considering $D^m_j$ independent of $x_j$ at iteration $m$, the data fit term defines a quadratic problem.
  
 The tangent fit energy 
  \begin{align*}
  \small
    \sum_{j=1}^N \sum_{i=1}^2 \alpha^m_{ij}
    \vecprod{t_j}{\partial_i\Phi^m(u_j)}^2 = & %
    \sum_{j=1}^N \sum_{i=1}^2 \alpha^m_{ij}
    \vecprod{t_j}{\left(L^m_i\cdot V\right)_j}^2. 
  \end{align*}
is also quadratic in $V$. Due to non-negative
  $\alpha_{ij}^m$ this expression is positive semi-definite. As the regularization $\norm{R V}^2 = V^\top R^\top R V$ is also quadratic and positive semi-definite, we can efficiently solve the linear system that characterizes its unique minimum with the conjugate gradient method. This method is well suited to the problem at hand, because the matrices that define our quadratic objective are sparse and we can use the MM iterations to warm start the optimization at iteration $m$ with the solution at iteration $m-1$. %

After optimizing $V$ at iteration $m$, we update the parameter values $u_1, \ldots, u_n$ where we approximate the squared distance to the surface with the matrices $D^m_1,\ldots, D^m_n$. In particular, we sample the surface $S^m=\Phi^m(\cT_\Mesh)$ uniformly and create a kd-tree from these samples to find quickly the parameters $u_j\in\cT_\Mesh$ of the surface sample closest to each input point $p_j\in\Pointcloud$. Although this update strategy might seem reminiscent of coordinate descent, our combination of coordinate updates with the quadratic approximation corresponds to applying a quasi-Newton algorithm~\cite{Wang2006,Liu2008} to the majorizer to avoid the slow convergence rates of coordinate descent. This removes the correspondence parameters from the optimization but can affect the majorizing property of \eqref{eq:RWLS} and puts our algorithm into the category of quadratic sequential programs instead of majorize-minimize algorithms. For this reason, we stop the process as soon as the original energy $E(\cdot)$ does not decrease any more. 
Our experiments indicate that this method provides better results than state-of-the-art methods.

{\small
\bibliographystyle{ieee}

}

\end{document}